\documentclass[sigconf]{acmart}

\AtBeginDocument{%
  \providecommand\BibTeX{{%
    \normalfont B\kern-0.5em{\scshape i\kern-0.25em b}\kern-0.8em\TeX}}}

\copyrightyear{2024}
\acmYear{2024}
\setcopyright{acmlicensed}\acmConference[KDD '24]{Proceedings of the 30th ACM SIGKDD Conference on Knowledge Discovery and Data Mining}{August 25--29, 2024}{Barcelona, Spain}
\acmBooktitle{Proceedings of the 30th ACM SIGKDD Conference on Knowledge Discovery and Data Mining (KDD '24), August 25--29, 2024, Barcelona, Spain}
\acmDOI{10.1145/3637528.3671954}
\acmISBN{979-8-4007-0490-1/24/08}

\acmConference[KDD '24]{Make sure to enter the correct
  conference title from your rights confirmation emai}{August 25–29, 2024}{Barcelona, Spain}

\usepackage{subfigure}
\usepackage{multirow}
\usepackage{caption}
\usepackage{graphicx}
\usepackage{float}
\usepackage{balance}
\begin{document}
\title{TSC: A Simple Two-Sided Constraint against Over-Smoothing}


%
%
%
%
%
%

\author{Furong Peng}
\email{pengfr@sxu.edu.cn}
\author{Kang Liu}
\email{not\_have@163.com}
\affiliation{%
  \institution{Institute of Big Data Science and Industry/School of Computer and Information Technology, Shanxi University}
  \streetaddress{92 Wucheng Roud, Xiao Dian}
  \city{Taiyuan}
  \state{Shanxi}
  \country{China}
  \postcode{030006}
}

\author{Xuan Lu}
\authornote{Corresponding author.}
\email{luxuan@sxu.edu.cn}
\affiliation{%
	\institution{College of Physics and Electronic Engineering, Shanxi University}
	\streetaddress{92 Wucheng Roud, Xiao Dian}
	\city{Taiyuan}
	\state{Shanxi}
	\country{China}
	\postcode{030006}
}

\author{Yuhua Qian}
\email{jinchengqyh@sxu.edu.cn}
\author{Hongren Yan}
\email{hyan\_sx@hotmail.com}
\affiliation{%
	\institution{Institute of Big Data Science and Industry/School of Computer and Information Technology, Shanxi University}
	\streetaddress{92 Wucheng Roud, Xiao Dian}
	\city{Taiyuan}
	\state{Shanxi}
	\country{China}
	\postcode{030006}
}

\author{Chao Ma}
\affiliation{%
  \institution{HOPERUN Information Technology}
  \streetaddress{168 Software Avenue, Yuhuatai}
  \city{Nanjing}\state{Jiangsu}
  \country{China}\postcode{210012}}
\email{AISuperMa@outlook.com}


\renewcommand{\shortauthors}{Furong Peng et al.}
\newcommand{\alert}[1]{\textcolor{red}{#1}}

\begin{abstract}
Graph Convolutional Neural Network (GCN), a widely adopted method for analyzing relational data, enhances node discriminability through the aggregation of neighboring information. Usually, stacking multiple layers can improve the performance of GCN by leveraging information from high-order neighbors. However, the increase of the network depth will induce the over-smoothing problem, which can be attributed to the quality and quantity of neighbors changing: (a) neighbor quality, node's neighbors become overlapping in high order, leading to aggregated information becoming indistinguishable, (b) neighbor quantity,  the exponentially growing aggregated neighbors submerges the node's initial feature by recursively aggregating operations. Current solutions mainly focus on one of the above causes and seldom consider both at once.

Aiming at tackling both causes of over-smoothing in one shot, we introduce a simple Two-Sided Constraint (TSC) for GCNs, comprising two straightforward yet potent techniques: random masking and contrastive constraint. The random masking acts on the representation matrix's columns to regulate the degree of information aggregation from neighbors, thus preventing the convergence of node representations. Meanwhile, the contrastive constraint, applied to the representation matrix's rows, enhances the discriminability of the nodes. Designed as a plug-in module, TSC can be easily coupled with GCN or SGC architectures. Experimental analyses on diverse real-world graph datasets verify that our approach markedly reduces the convergence of node's representation and the performance degradation in deeper GCN.
\end{abstract}

\ccsdesc[500]{Information systems~Data mining}
\ccsdesc[500]{Computing methodologies~Learning latent representations}

\keywords{GCN, Over-smoothing, Random masking, Contrastive learning}


\maketitle

\section{Introduction}\label{sc:intro}

Graph Convolutional Neural Networks (GCNs) \cite{Defferrard2016NIPS,kipf2017semisupervisedICLR,Gilmer2017ICML,feng2020graph} has gained significant attention for their promising results in processing relational data. Gradually emerging as a mainstream technology for various applications, including natural language processing \cite{Wu2023Suervey}, E-commerce \cite{Xv2023KDD}, materials chemistry \cite{Reiser2022ComMat}, biomedical \cite{Hao2020KDD}, and trajectory prediction \cite{Xu2022CVPR}, GCNs excel at leveraging topological structures to capture relational characteristics. Despite their effectiveness, stacking multiple GCN layers for high-order neighbors introduces challenges such as over-smoothing \cite{Li2018AAAI}, gradients vanishing \cite{Wu2021NIPS}, and overfitting \cite{Bo2022AAAI}. Notably, over-smoothing plays a crucial role in affecting performance in deep GCNs.

\begin{figure}[t]
	\centering
	\subfigure[GCN's node visualization at 1st, 8th, and 32nd layers]{
		\centering
		\includegraphics[width=0.33\columnwidth]{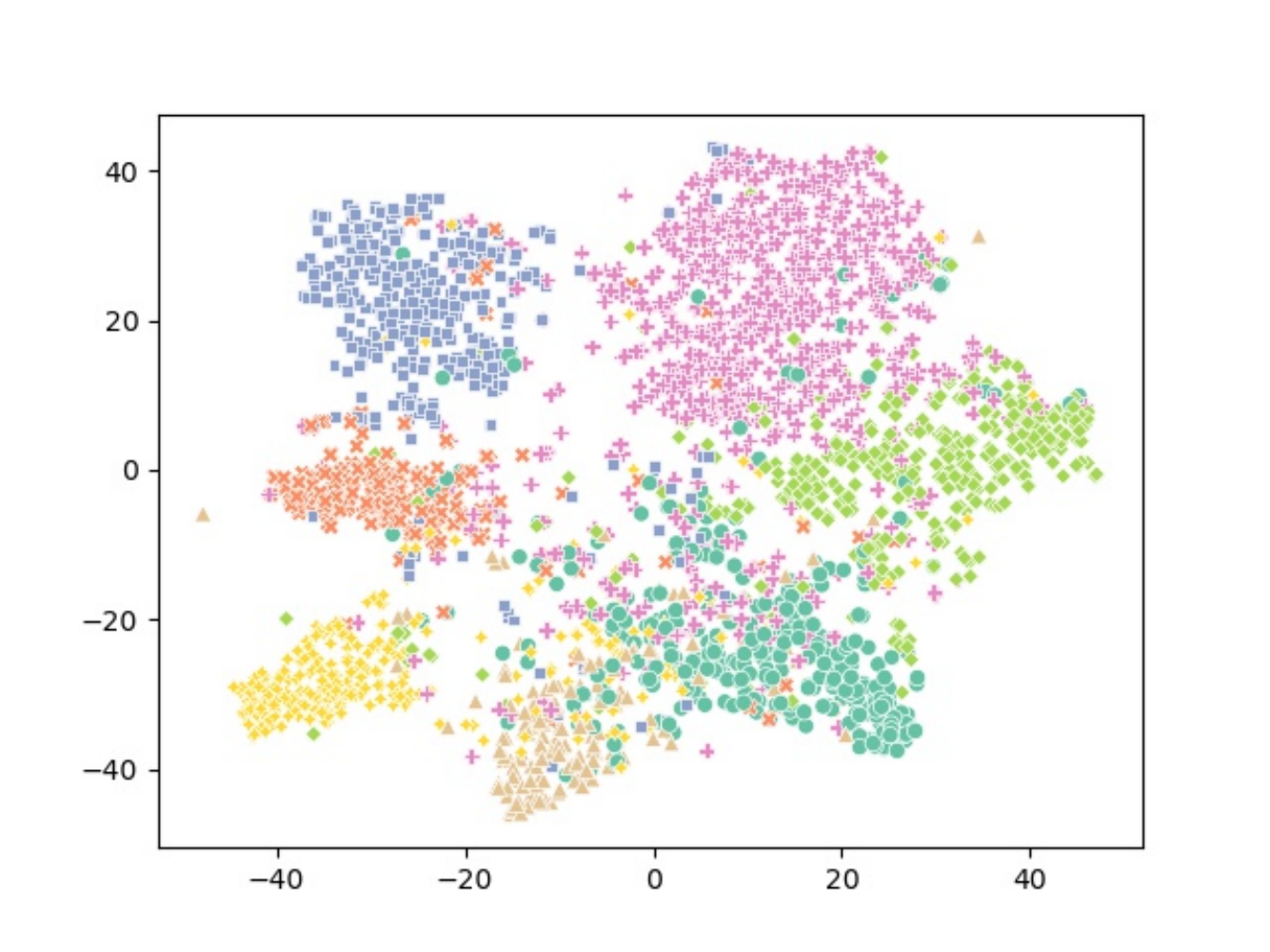}
		\includegraphics[width=0.33\columnwidth]{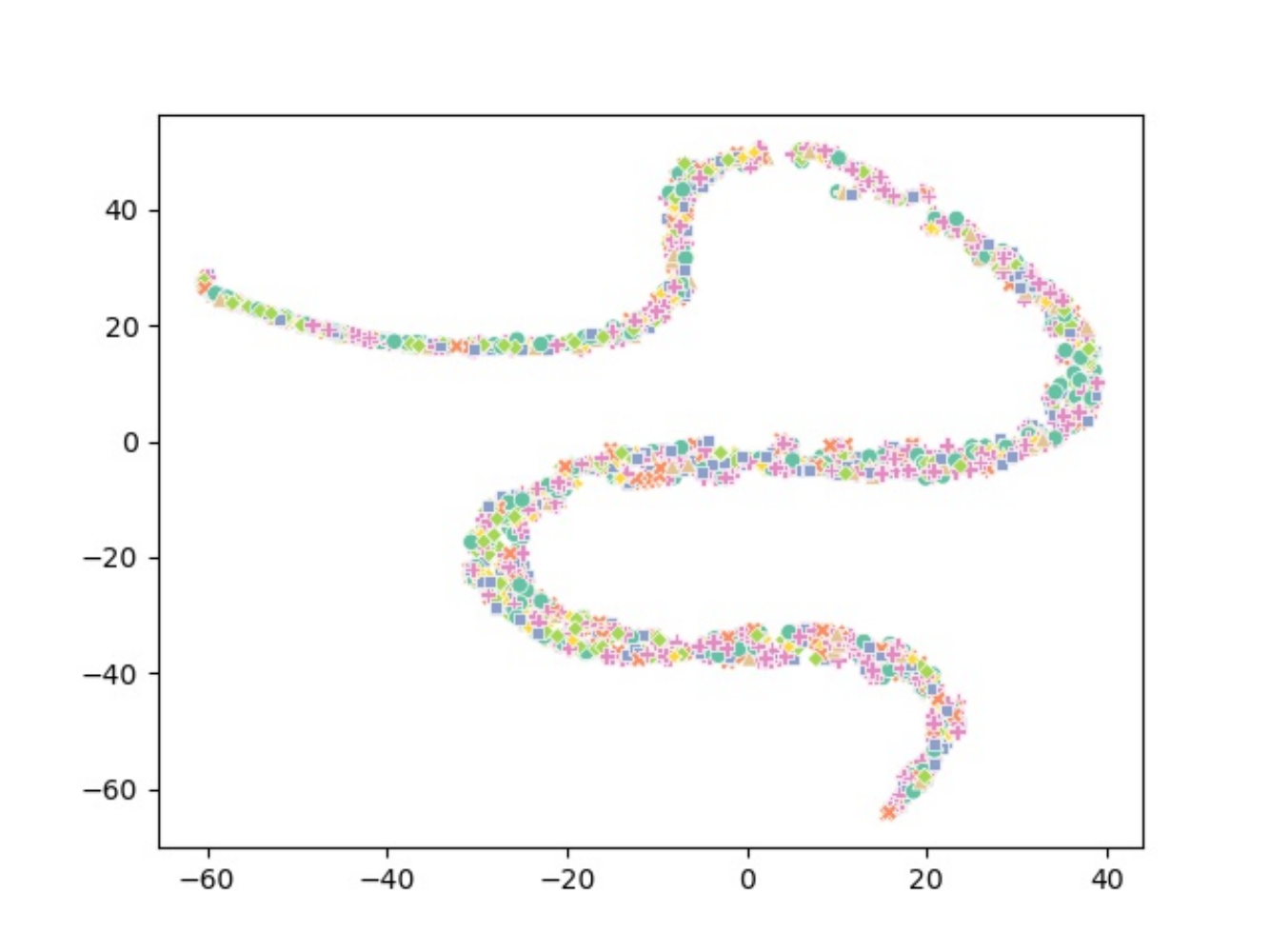}
		\includegraphics[width=0.33\columnwidth]{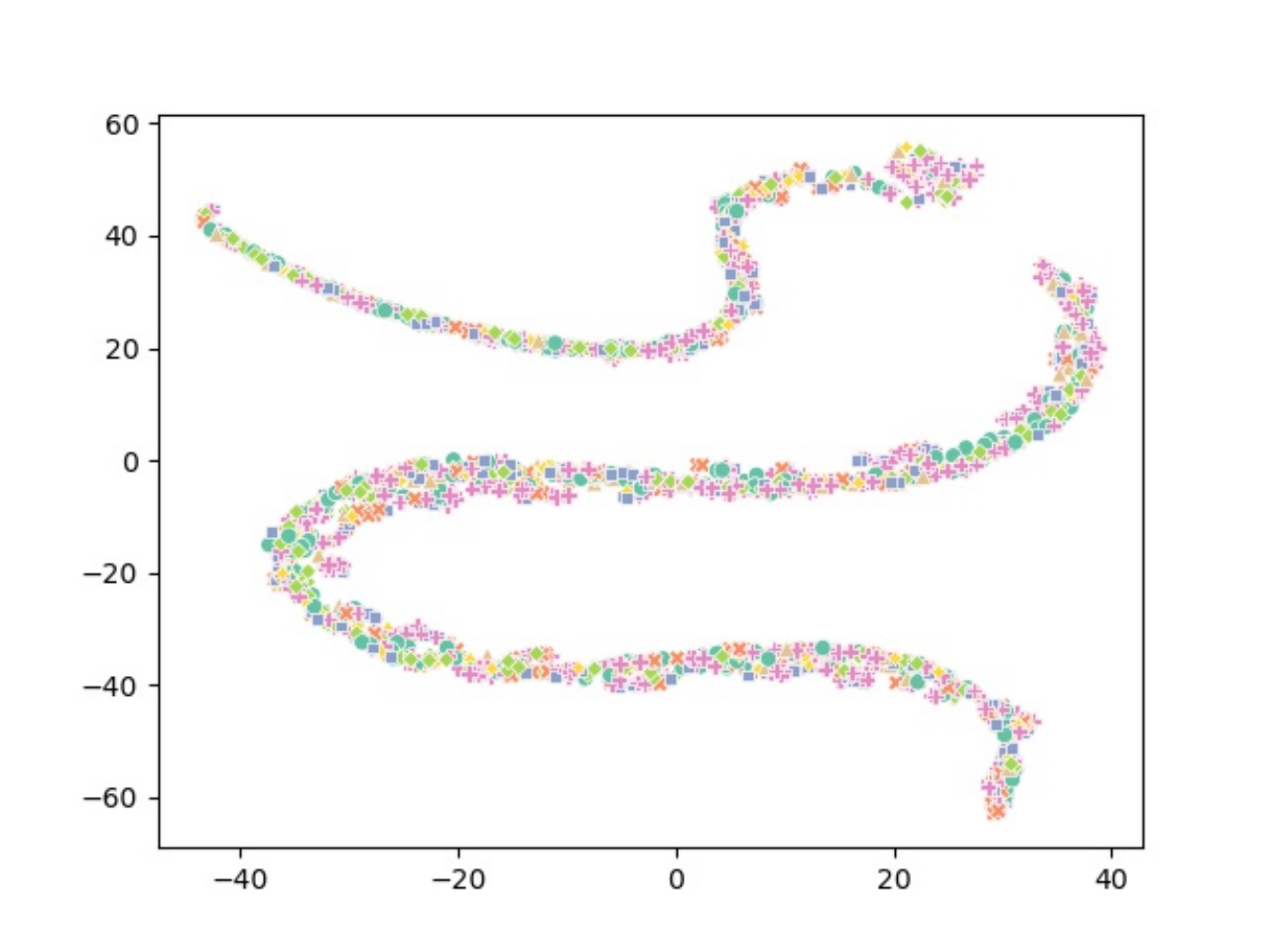}
		
	}%
	\hspace{0.1\linewidth}
	\subfigure[GCN+DropMessage's node visualization at 1st, 8th, and 32nd layers]{
		\centering
		\includegraphics[width=0.33\columnwidth]{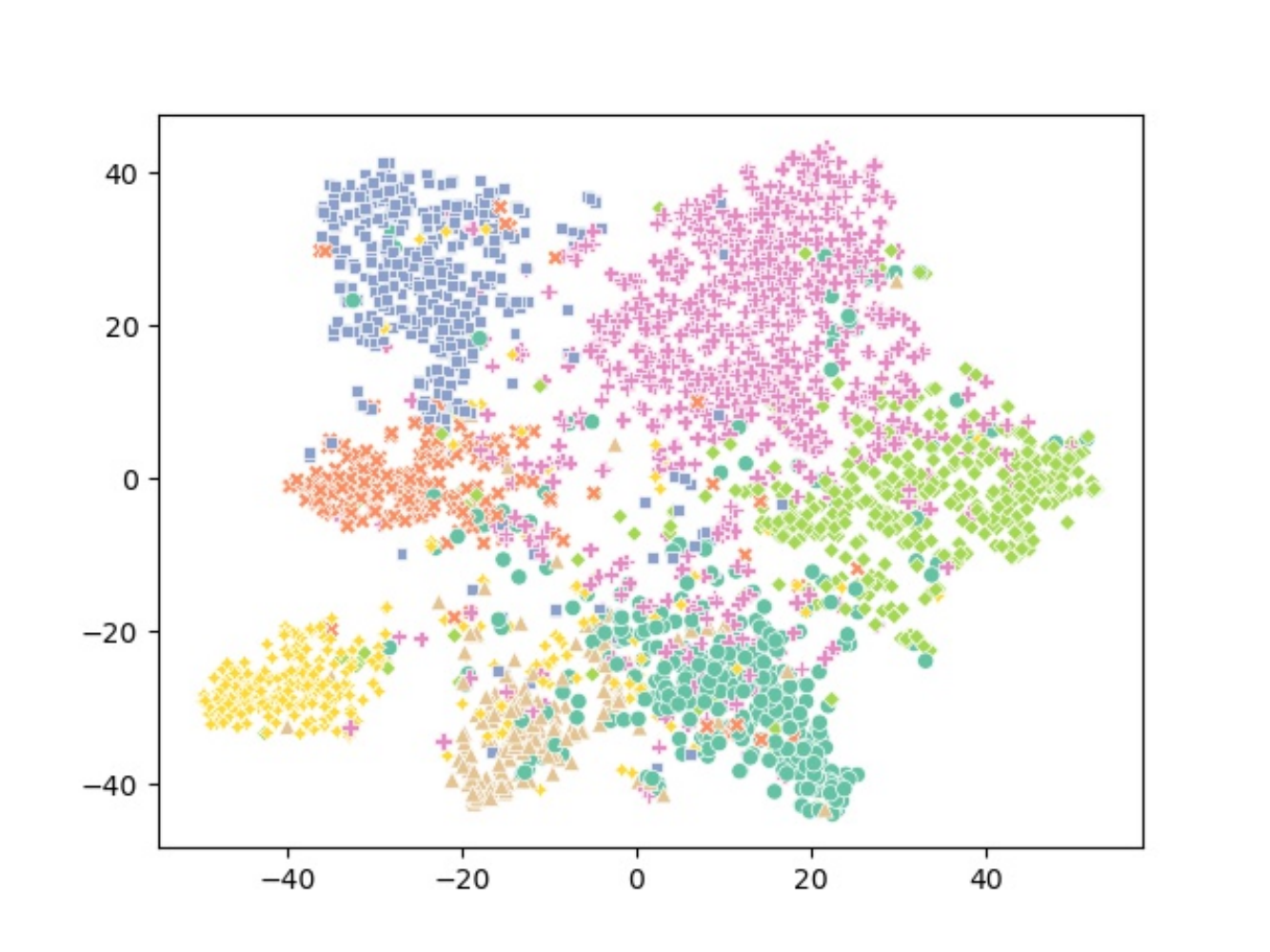}
		\includegraphics[width=0.33\columnwidth]{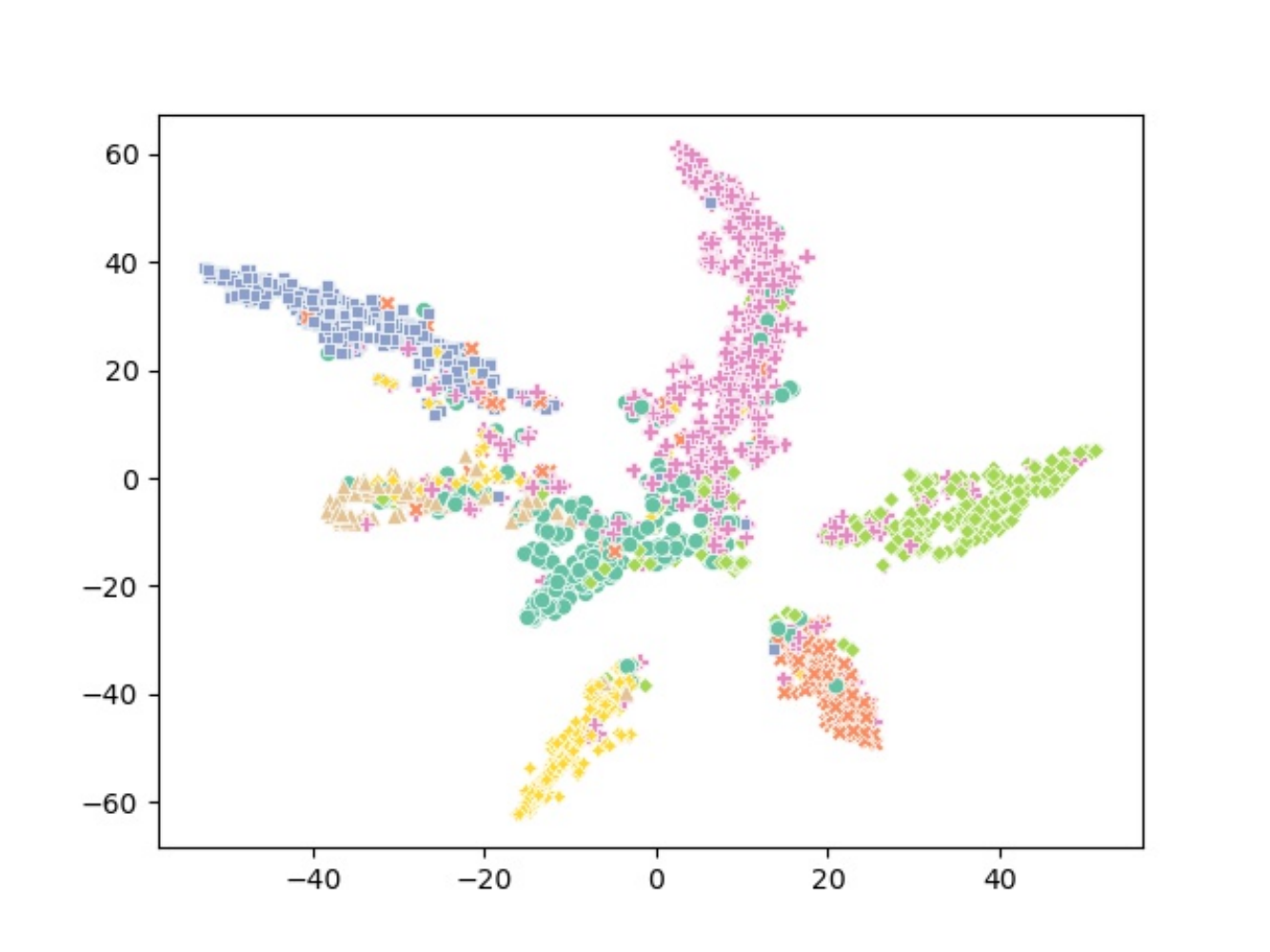}
		\includegraphics[width=0.33\columnwidth]{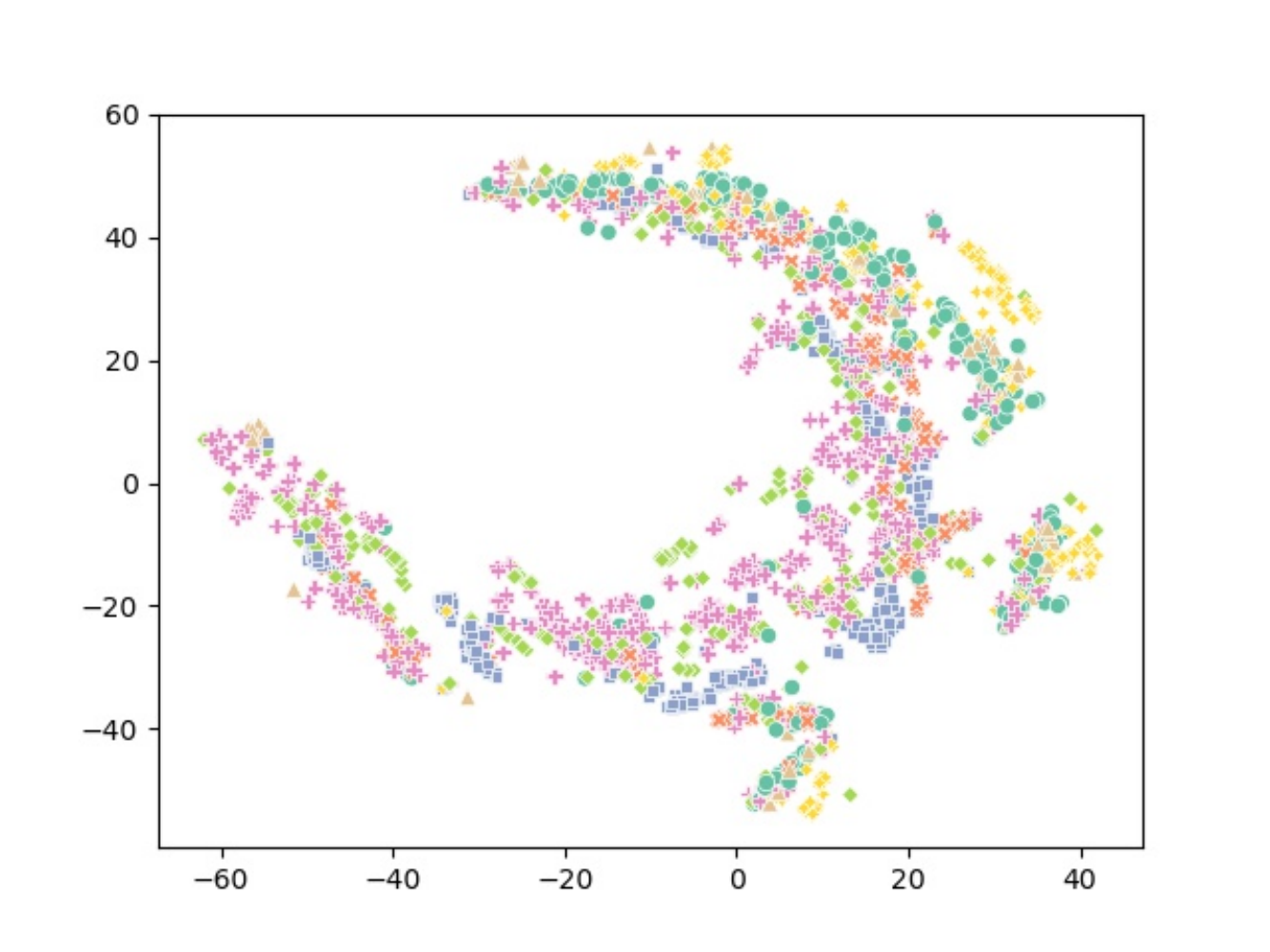}
		\label{fg:node-representation-dropmessage}
	
	}%
	\hspace{0.1\linewidth}
	\subfigure[ GCN+ContraNorm's node visualization at 1st, 8th, and 32nd layers]{
		\centering
		\includegraphics[width=0.33\linewidth]{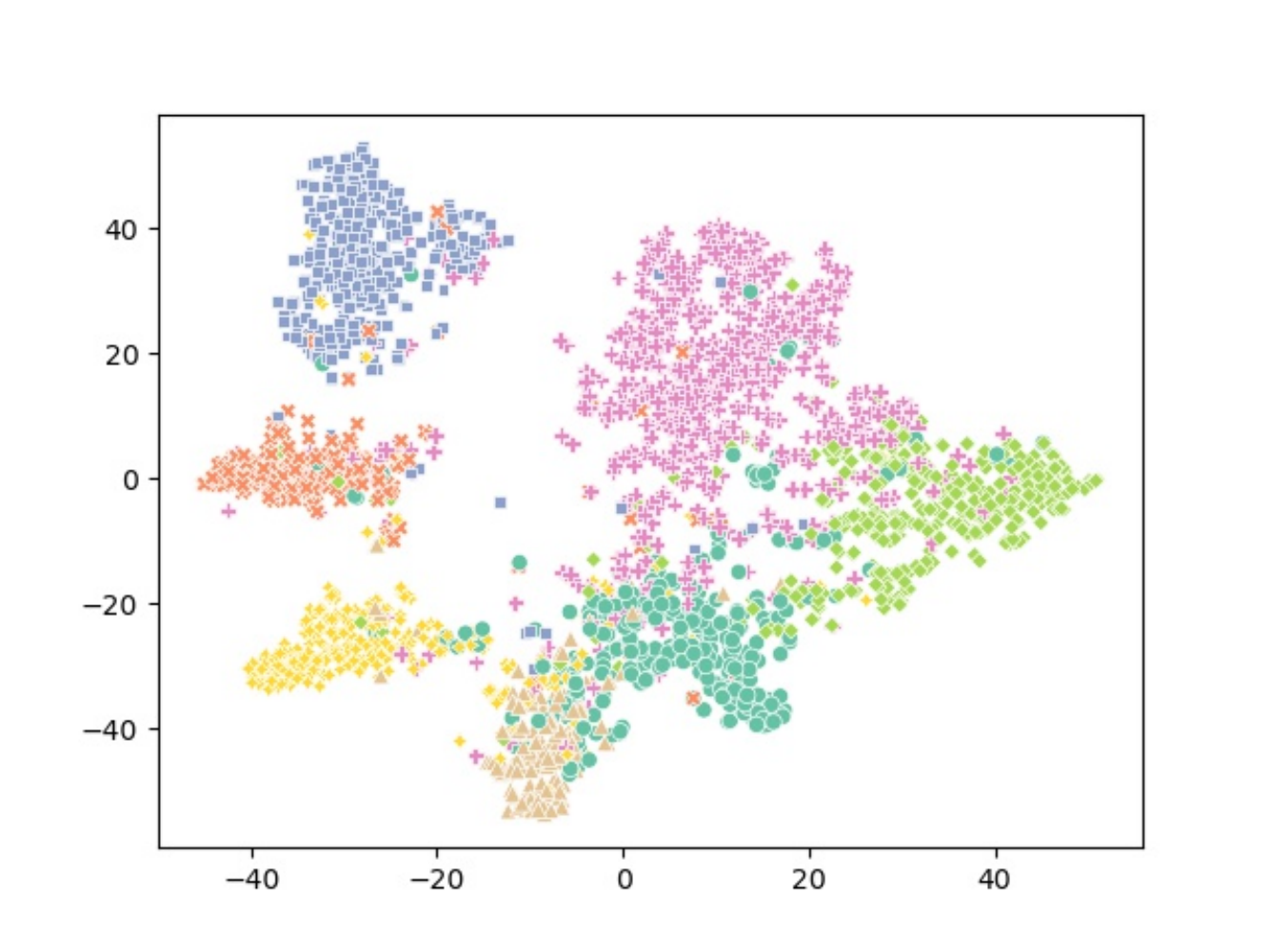}
		\includegraphics[width=0.33\linewidth]{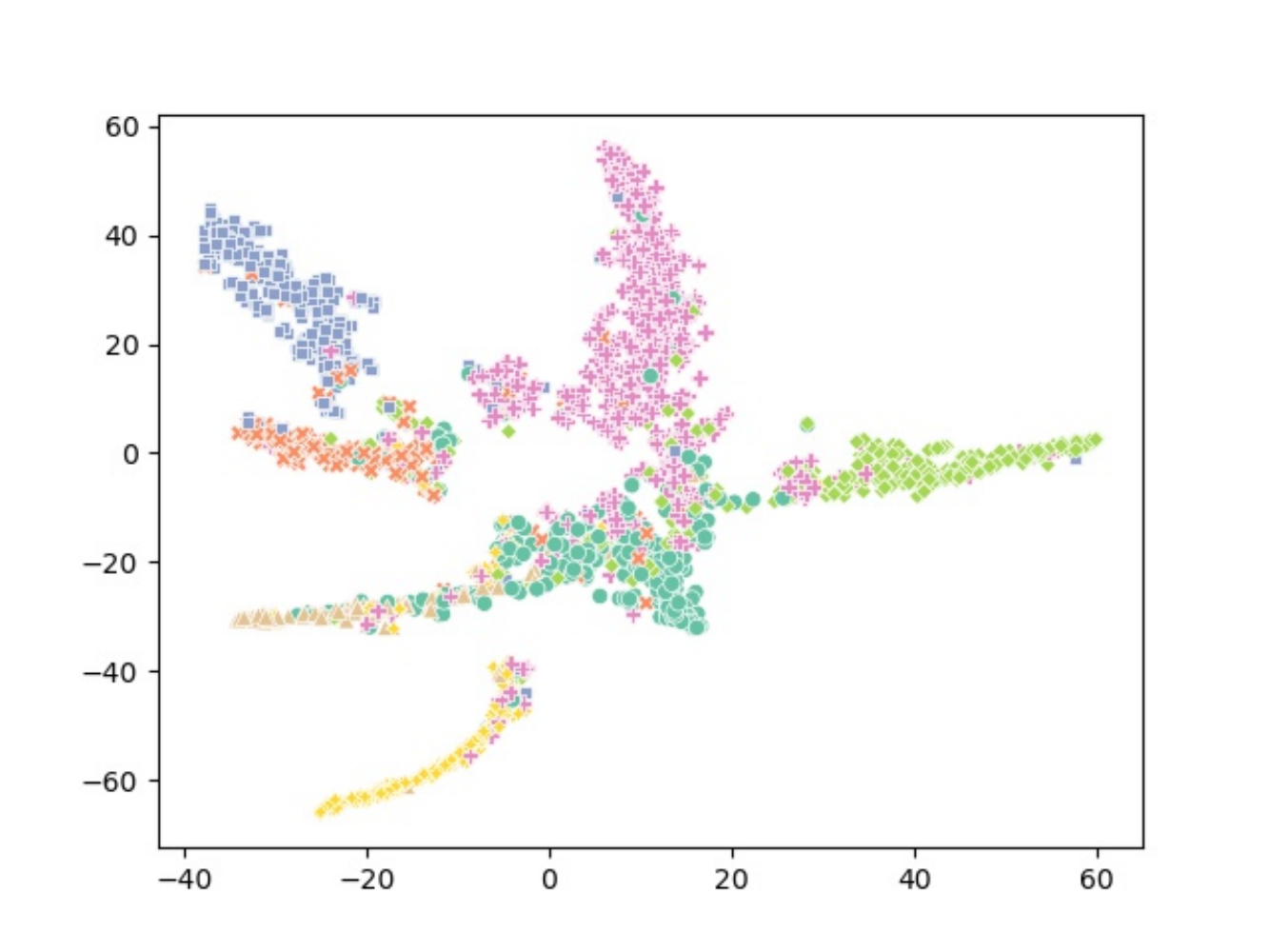}
		\includegraphics[width=0.33\linewidth]{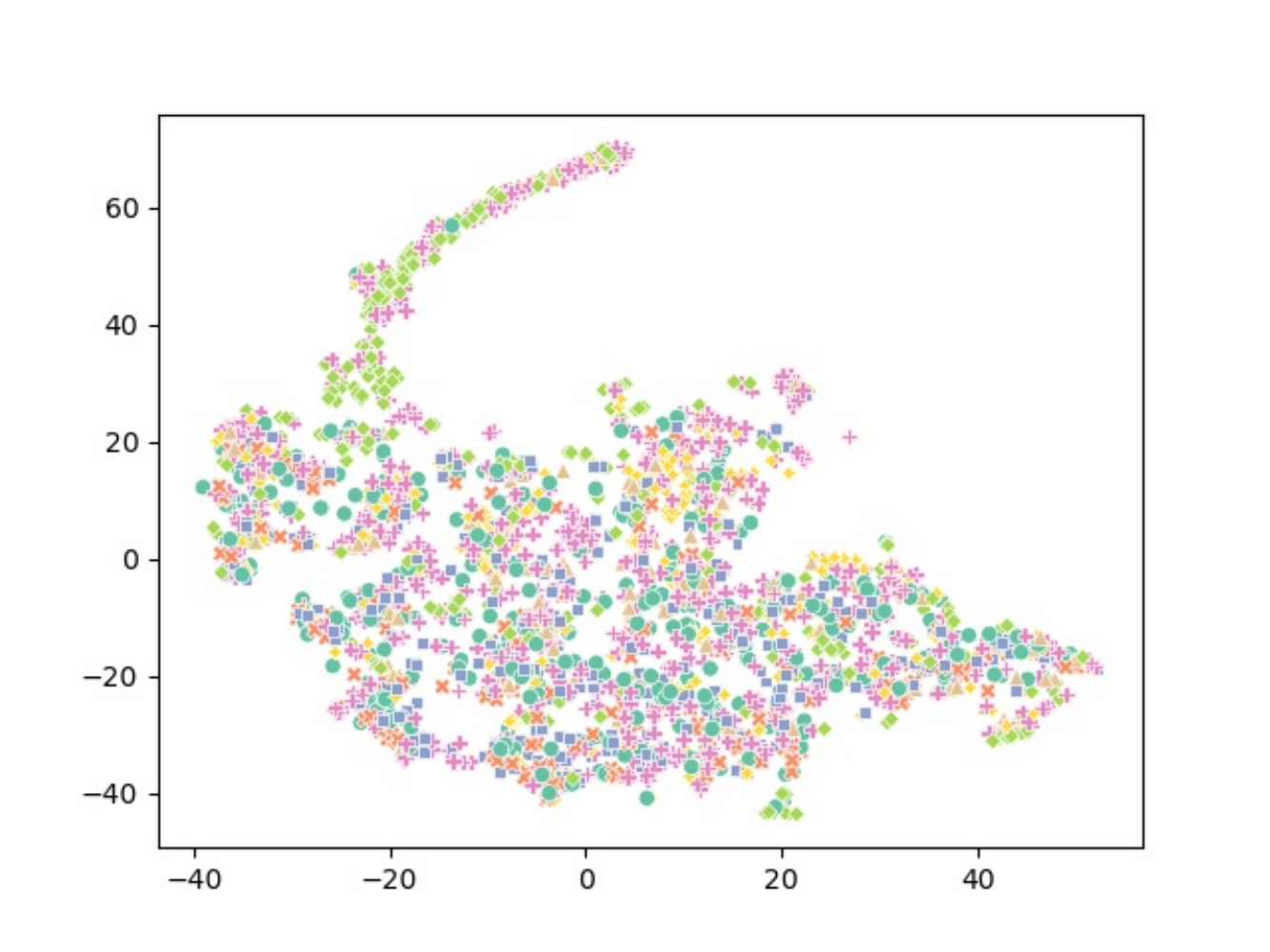}
		\label{fg:node-representation-contranorm}	
		
	}%
	\hspace{0.1\linewidth}
		\subfigure[Our model's node visualization at 1st, 8th, and 32nd layers]{
			\centering
			\includegraphics[width=0.33\linewidth]{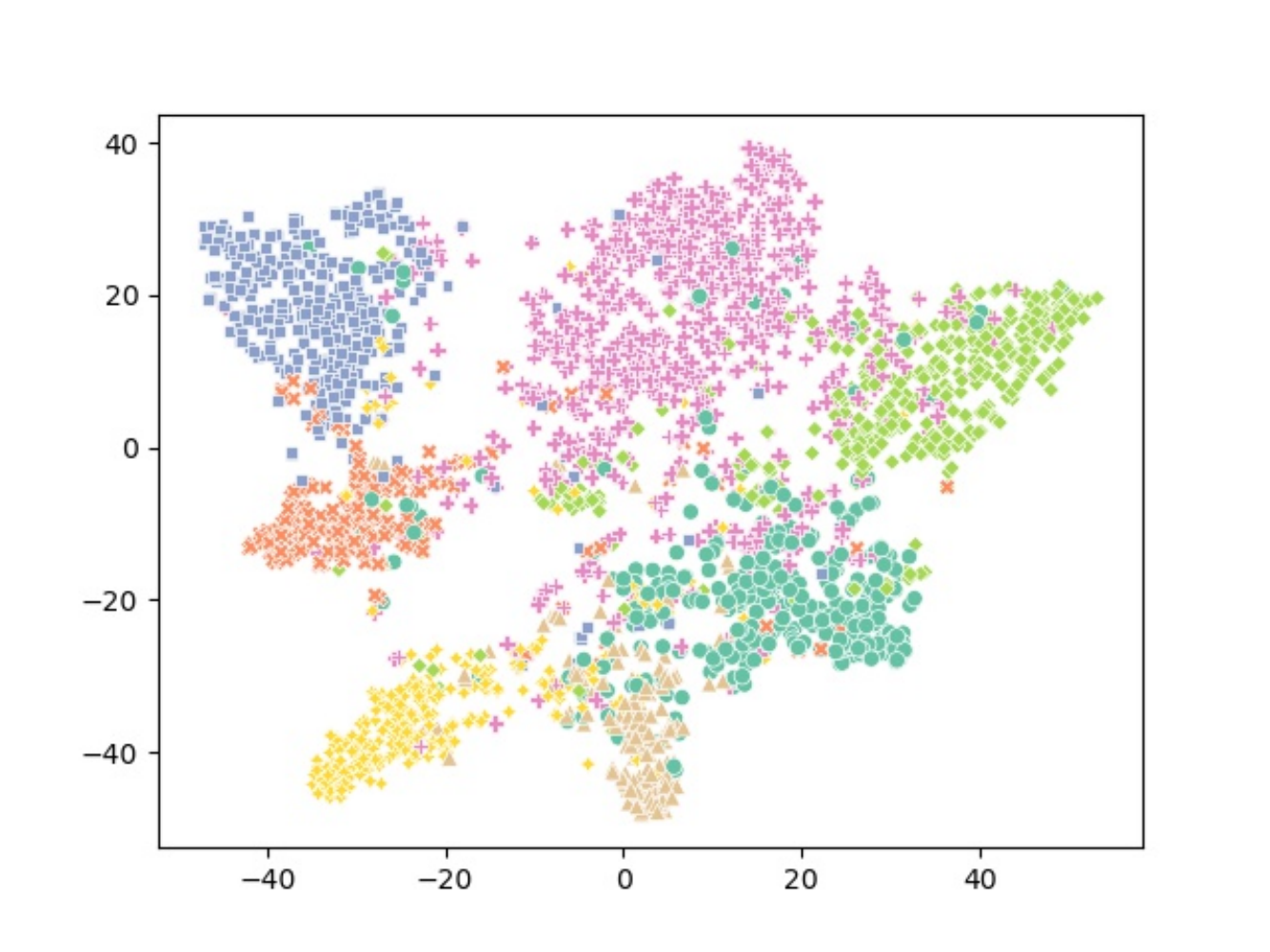}
			\includegraphics[width=0.33\linewidth]{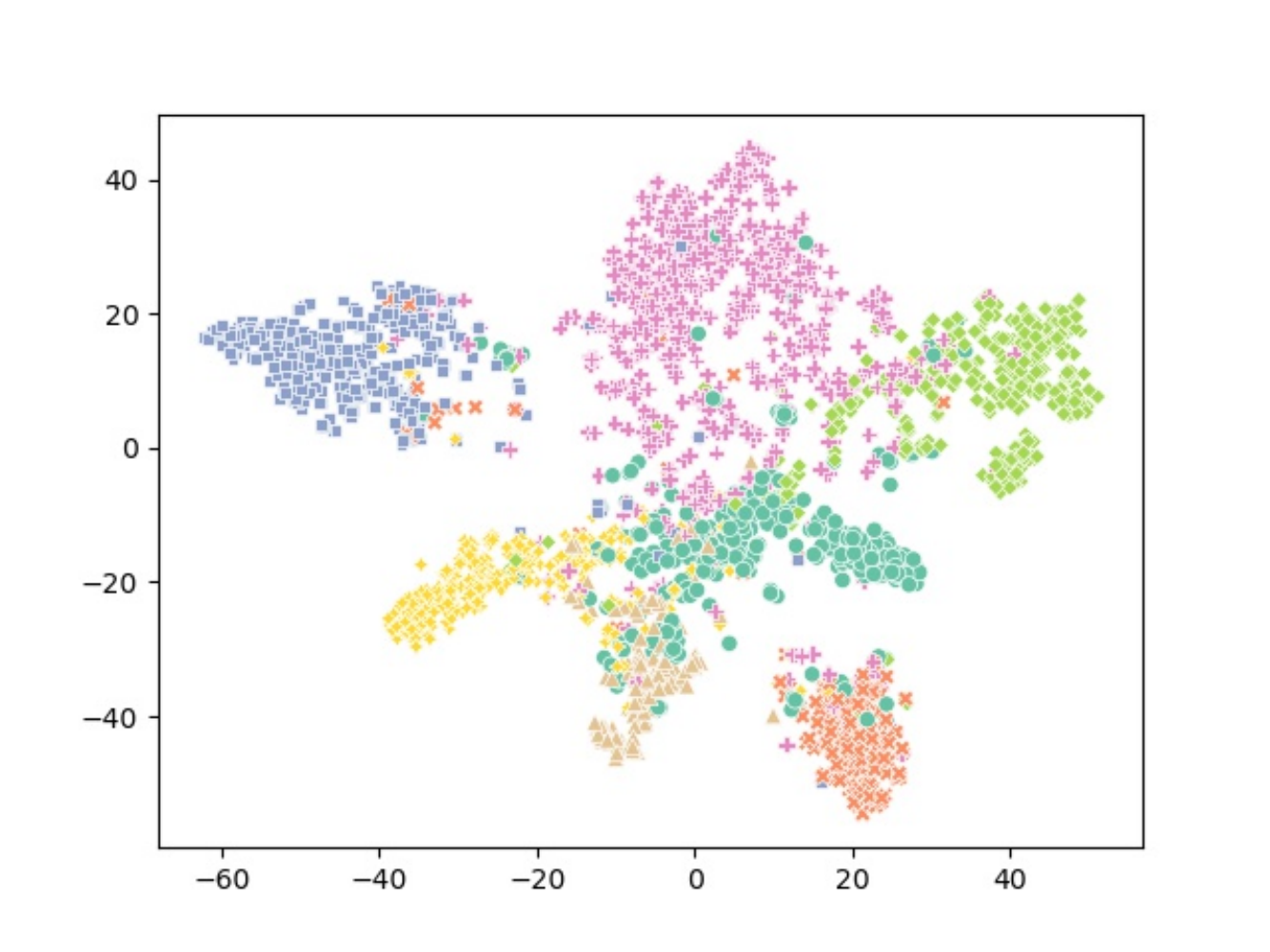}
			\includegraphics[width=0.33\linewidth]{figs/3.6TSC_GCNlayer32}
			\label{fg:node-representation-TSC}	
		}%
	\caption{Visualization of node representations in different layers. The 1st, 2nd, and 3rd columns depict the visualizations of node representations at the 1st, 8th, and 32nd layers, respectively. Row (a) shows the node distribution for vanilla GCN, (b) for DropMessage, (c) for ContraNorm, and (d) for our proposed TSC applied to GCN. Node colors indicate the categories. All sub-figures are visualized using t-SNE.  }
	\label{fg:node-representation}	
\end{figure}

\begin{figure}[t]
	\subfigure[The 1-order neighborhood]{
		\centering
		\includegraphics[width=0.5\columnwidth]{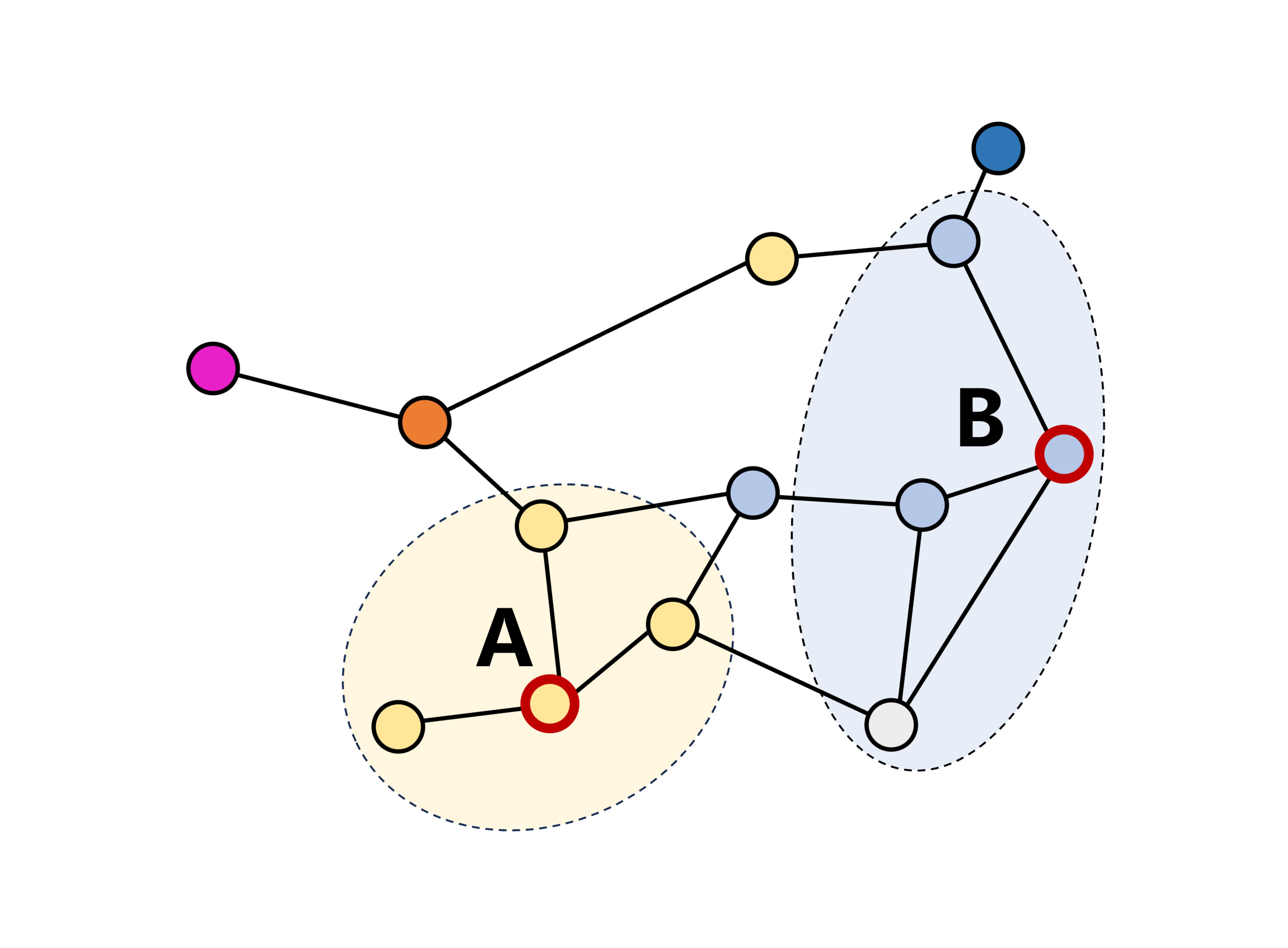}
		
	}%
	\subfigure[The 3-order neighborhood]{
		\centering
		\includegraphics[width=0.5\columnwidth]{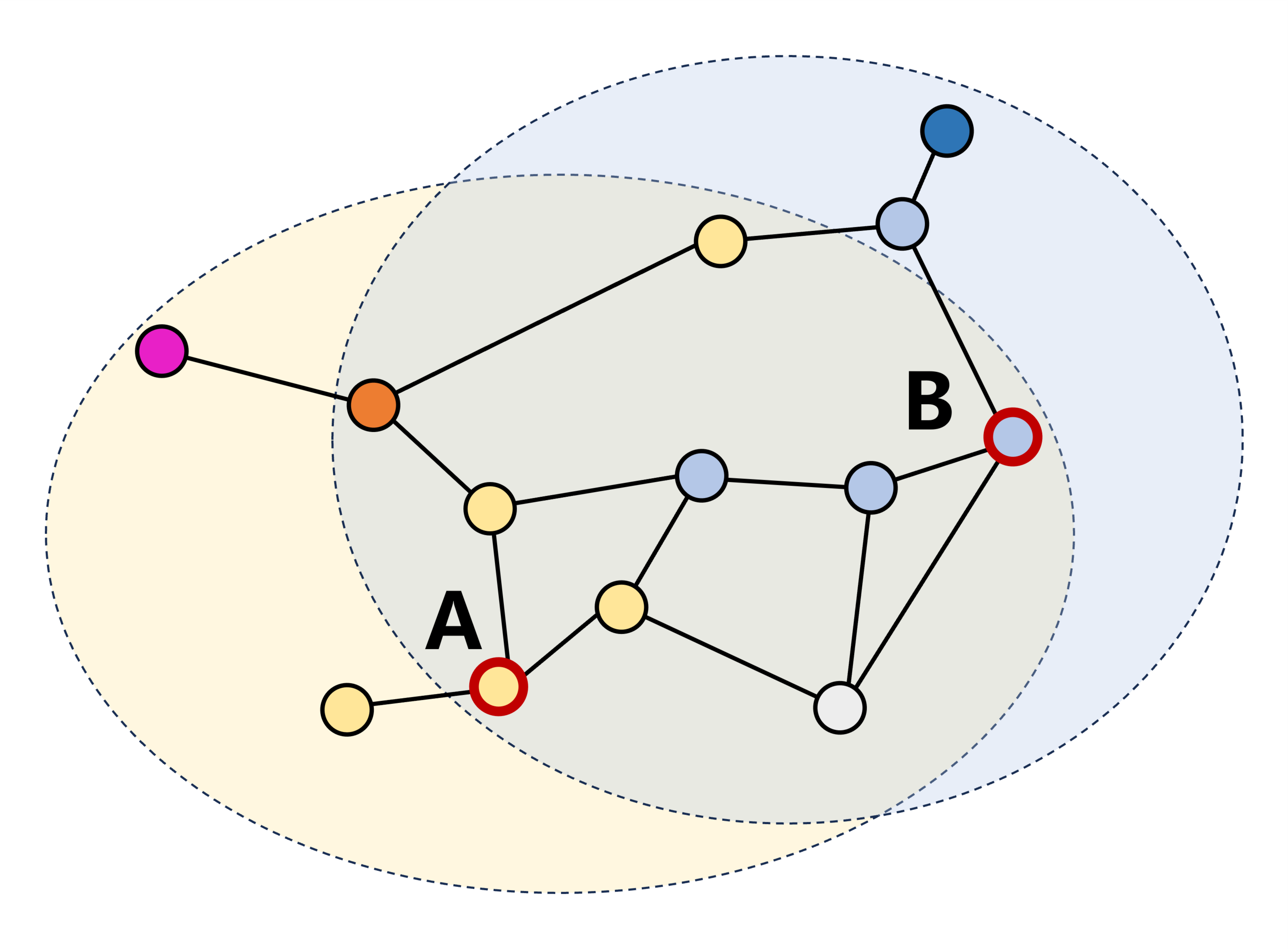}
		
	}%
	\caption{The neighbors changes across different orders. In the 3-order neighborhood, the node A and B have overlapped neighbors and have large number of neighbors.}
	\label{fg:node-neighbors}	
\end{figure}

The phenomenon of over-smoothing, initially analyzed by \cite{Li2018AAAI,gasteiger2018predict}, arises as the depth of GCN layers increases, deteriorating the ability to differentiate node representations and significantly impairing model performance. As shown in the 1st row of Figure \ref{fg:node-representation}, the node distribution of the GCN is quite scattered in the 1st layer. Yet, with layers increasing, node representations converge (notably at the 8th and 32nd layers), leading to node categories nearly indistinguishable in visualizations. The underlying reason lies in the high-order neighbor aggregation operation, which leads to \textit{neighbor overlapping} and \textit{individuality overwhelmed}. Figure \ref{fg:node-neighbors} shows that nodes \textit{A} and \textit{B} have unique neighbors in their 1-order neighborhood, but in the 3-order neighborhood, they share overlapping neighbors and accumulate many non-homogeneous nodes. High-order neighbors result in different nodes aggregating information from the same neighbors (by low-quality neighbors) and the loss of node individuality due to the large volume of neighbor nodes (by large quantity of neighbors). For example as mentioned by \cite{oono2019graph}. Detailed measurement can be found in Appendix F.

Several models have been proposed to address over-smoothing, and can be broadly categorized into neighbor filtering and individuality enhancement methods.

\textit{Neighbor Filtering Models:} These models selectively aggregate neighbors' information to prevent nodes from becoming indistinguishable, either through explicit or implicit filtering. The explicit methods include DropEdge \cite{Rong2020ICLR}, DropMessage \cite{Fang2023AAAI}, and DropNode \cite{Do2021ESWA}, which selectively drop edges, messages, or nodes. While these methods alleviate over-smoothing by filtering neighbors information, their accuracy diminishes in deep layers, leading to category confusion, as illustrated Figure \ref{fg:node-representation-dropmessage}. For instance, DropMessage exhibits a clear clustering structure in 8th layer but loses this advantage in the deeper 32th layer due to information overload from neighbors. Meanwhile, the implicit methods, including PairNorm \cite{Zhao2020ICLR}, DGN \cite{Zhou2020NIPS}, NodeNorm \cite{Zhou2021CKIM}, and ContraNorm \cite{Guo2023ICLR} adjust the received neighborhood information by adding normalization to the GCNs at each layer. These normalization methods implicitly re-weight the aggregated information, yet they remain ineffective in addressing over-smoothing caused by the overwhelming of neighbors, as illustrated in Figure \ref{fg:node-representation-contranorm}.  ContraNorm's node representation also becomes indistinguishable in the 32nd layer. From the above analysis, we can see that neighbor filtering can alleviate the over-smoothing caused by the low quality of  neighbors (neighbor overlapping issue) but neighbor quantity.

\textit{Individuality Enhancement Models:} These models add a node's self-representation in the last or first layer to the new layer's representation. Examples include JKNet \cite{Xu2018ICML}, GCNII \cite{Chen2020ICML}, and AIR \cite{zhang2022model_KDD}, which incorporate the first or shallow layer's representation to new layers alongside aggregated information. However, they add the shallow representation directly to the new layer according to a certain weighting instead of adding as a part of it. It may lead some nodes to overemphasize the impact of the first or shallow layer, and therefore unable to exploiting high-order neighbors.

In this paper, we focus on both neighbor quality and quantity issues, and introduce the \textbf{Two-Sided Constraint} with two techniques to combat over-smoothing: Firstly, \textbf{Random Masking} selectively drops columns of aggregated information, replacing them with the node's representation from the previous layer. This technique, acting as a neighbor filtering mechanism, effectively prevents representation convergence while preserving node individuality, as shown in Figure \ref{fg:node-representation-TSC}. Secondly, the \textbf{Contrastive Constraint} enhances node individuality by minimizing changes in the same node's representation across layers and enlarging the representation differences between distinct nodes, both within and across layers. This technique, which is applied to the row of node features, serves as a mechanism for enhancing individuality to mitigate the overwhelming effect of homogenization. These two constants are applied to the two sides of representation matrix, column and row respectively, and therefore are called two-sided constraint (TSC).

In addition to the over-smoothing issue, GCN also faces problems of gradient-vanishing and overfitting. To minimize the impact of these latter issues on the model, we opt to analyze the performance of$\;$TSC using Simplifying Graph Convolution (SGC). SGC, by eliminating linear transformations and nonlinear activation functions, can avoid these two problems. Additionally, to explore the versatility of TSC, we also implement it in GCN, providing relevant technical details and experimental results.

The main contributions of this work can be summarized as follows.

\textbf{(1)} We provide a new perspective of graph over-smoothing in view of \textit{neighbor overlapping} and \textit{individuality overwhelmed}. Solving the neighbor overlapping can alleviate the representation convergence. Taking care of individuality overwhelmed can maintain the overall performance from degradation. Base on this perspective, we proposed a simple but potent two-sided constraint to the column and row of the representation matrix.

\textbf{(2)} On the column of representation matrix, we propose random masking against representation convergence and keep nodes' individuality. This strategy is more effective than some recent SOTAs. 

\textbf{(3)} On the row of representation matrix, we introduce contrastive constraints to increase differences and discriminative power among nodes.

\section{Related Work}
As the number of Graph Convolutional Network (GCN) layers increases, over-smoothing occurs. This happens because higher-order neighbor aggregation operations cause node features to become indistinguishable and overwhelm node individuality due to variations in the quality and quantity of neighbors in high order. Many methods attempt to address this issue by reducing the number of neighbors through random dropping (e.g., DropMessage, DropNode) or normalization (e.g., ContraNorm), which can be categorized as neighbor filtering methods. The presence of high-order neighbors also leads to different nodes aggregating similar information caused by neighbor sharing, resulting in the loss of the node's individuality. Several methods attempt to mitigate this issue by reducing the influence of high-order neighbors and increasing the impact of information from shallow layers, as seen in JKNet and GCNII. These methods can be categorized as individuality enhancement techniques. We will discuss them in detail.

\subsection{Neighbor Filtering}
\textbf{Explicit Filtering Methods:} Models like DropEdge \cite{Rong2020ICLR} and DropNode \cite{Do2021ESWA} explicitly filter neighbors by randomly deleting edges or nodes. DropMessage \cite{Fang2023AAAI} directly drops information from the message passing matrix. While effective in mild deep layers, these explicit filtering methods suffer from over-smoothing in deeper layers, as illustrated in Figure \ref{fg:node-representation-dropmessage}.

\textbf{Implicit Filtering Methods:} These methods manipulate neighbor impact through normalization. PairNorm \cite{Zhao2020ICLR} adds a constant to the sum of pairwise node distances to reduce similarity between distant nodes. DGN \cite{Zhou2020NIPS} introduces differentiable group normalization, NodeNorm \cite{Zhou2021CKIM} normalizes each feature vector by node-specific statistical properties, and ContraNorm \cite{Guo2023ICLR} employs contrastive learning techniques for a more uniform distribution. While these methods scatter node representations in deep layers, normalization biases may cause the obtained distribution to deviate from node categories. In contrast, our approach introduces random masking to the columns of the representation matrix, preserving the distribution of nodes, as illustrated in Figure \ref{fg:node-representation-contranorm}.

\subsection{Individuality Enhancement}

The Individuality Enhancement technique involves adding a node's original feature (often from the first or last layer) to a new layer. JKNet \cite{Xu2018ICML} adds each layer's feature to the final representation, GCNII \cite{Chen2020ICML} adds the first layer's feature to each subsequent layer, and DeepGCNs \cite{li2019deepgcns} adopts a ResNet-like approach by adding features from the first and last layers simultaneously. However, excessively incorporating shallow node information into new layers may hinder obtaining information from high-order neighbors. This trade-off requires careful tuning for each scenario.

Recent models address this challenge by adaptively integrating shallow layer information. DRC \cite{yang2022difference_ACMMM} uses the difference between the previous layer's input and output as the input for the next layer, allowing adaptive selection of previous information. Zhang et al. \cite{zhang2022model_KDD} propose Adaptive Initial Residual (AIR) for propagation and transformation steps, preventing over-smoothing and performance degradation in early layers. While adding information from shallow layers prevents degradation, it may impede aggregation of information from high-order neighbors. In our work, we propose smoothing the node's transitions between layers through a contrastive constraint. This constraint not only brings the node's representation close to its last layer but also separates it from other nodes' representations.

\section{Preliminaries}

\subsection{Problem Definition}
In this paper, we focus on the unweighted undirected graph $G =\{V,E\}$, where $V =\{v_1,...,v_n\}$  denotes a $n$ node set and $E$ is the edge set defined on $G$. Let $\textbf{A}\in{\{0,1\}}^{n\times n}$ denotes the symmetric adjacency matrix with $\textbf{A}_{i,j}=\textbf{A}_{j,i}$. Then, the degrees of the nodes are defined by $D =\{d_1,...,d_n\}$, where $D_i={\textstyle\sum_{j\neq i}}\textbf{A}_{i,j}$. The degree matrix is further defined as the diagonal matrix $\textbf{D}$ with $\textbf{D}_{i,i}=D_i(i=1,...,n)$. $\textbf{L}={\textbf{D}}^{-\frac{1}{2}}(\textbf{A}+\textbf{I}){\textbf{D}}^{-\frac{1}{2}} $ is the Laplace matrix of $G$. $ \textbf{X}\in\mathbb{R}^{n\times f}$ denotes the feature matrix of the nodes, where $f$ is the dimension of feature. $\textbf{H}\in\mathbb{R}^{n\times d}$ is node representation matrix, that is used for node classification and message passing. Only parts of nodes are labeled, our task is to predict the labels of the rest nodes. This task is called semi-supervised node classification \cite{Zhao2020ICLR}.
\subsection{GCN}
Graph Convolutional Network (GCN), as a representative method, propagates node features through the adjacency matrix to aggregate the information of neighbors. The aggregation operation for layer $l$ can be defined by \cite{kipf2017semisupervisedICLR}:
\begin{equation}
	{\textbf{H}^{(l+1)}}=\sigma (\textbf{L}{\textbf{H}^{(l)}}{\textbf{W}^{(l)}}),
\end{equation}
where $\sigma (\cdot )$ is the activation function, $l$ represents the number of layers. $\textbf{H}^{(l)}$ and ${\textbf{W}^{(l)}}$ are the node representation and weight matrices of the $l$-th layer, respectively. $\textbf{L}$ is the Laplace matrix of the input Graph. In deep GCN models, the number of parameters grows rapidly, endowing them with strong expressive power. However, they often suffer from the issue of over-smoothing. Moreover, the vast volume of parameters also makes it challenging to handle large-scale graph data effectively.
\subsection{SGC}
Simplifying Graph Convolution (SGC)\cite{Wu2019SGCNICML}, a simplified version of GCN, removes the nonlinear activation and linear projection operation from GCN. The representation matrix in $l$-layer is calculated by a very simple linear projection.

\begin{figure*}[t]	
	\centering	
		\subfigure[SGC+TSC]{
		\centering
		\includegraphics[width=0.46\linewidth]{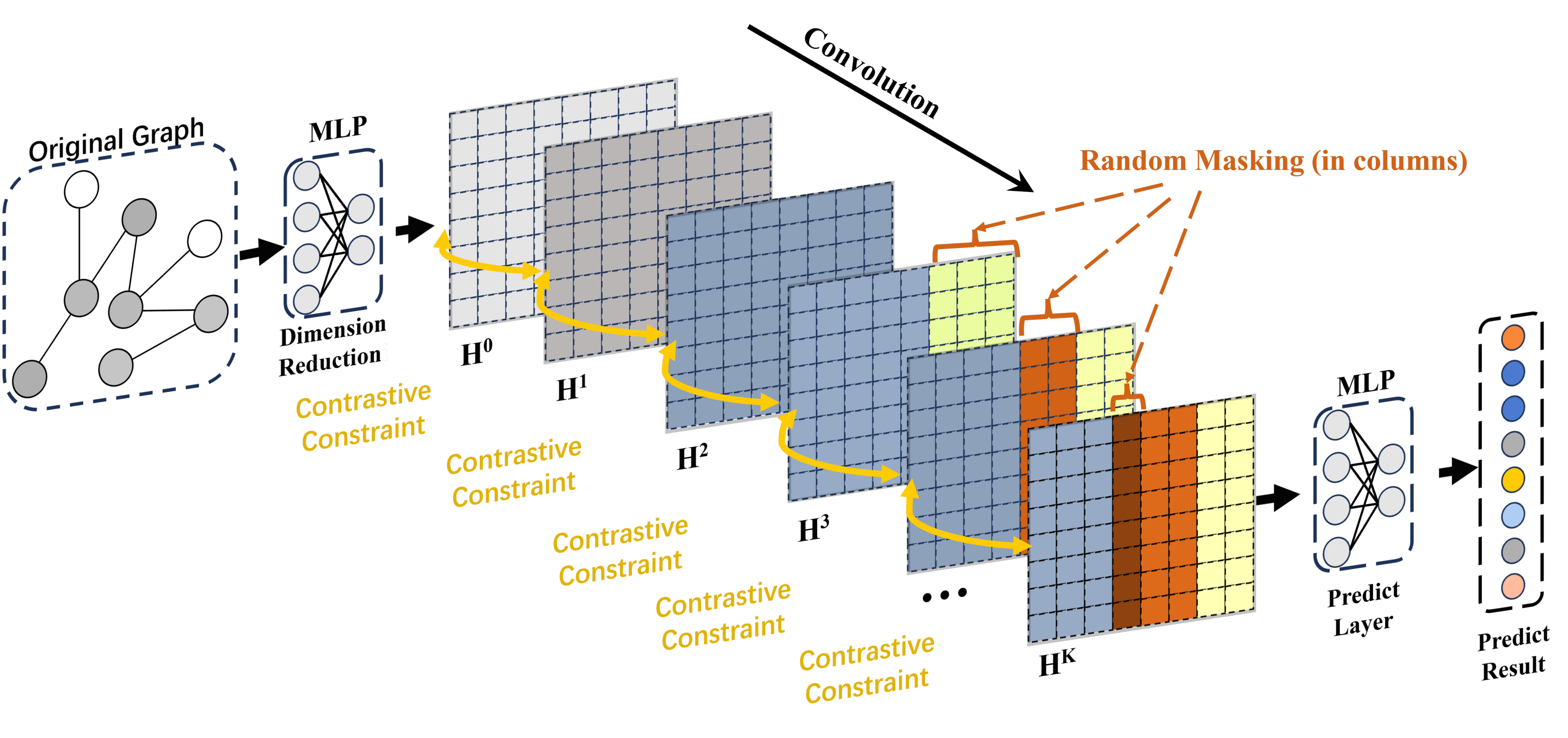}
		
	}%
	\subfigure[GCN+TSC]{
		\centering
		\includegraphics[width=0.46\linewidth]{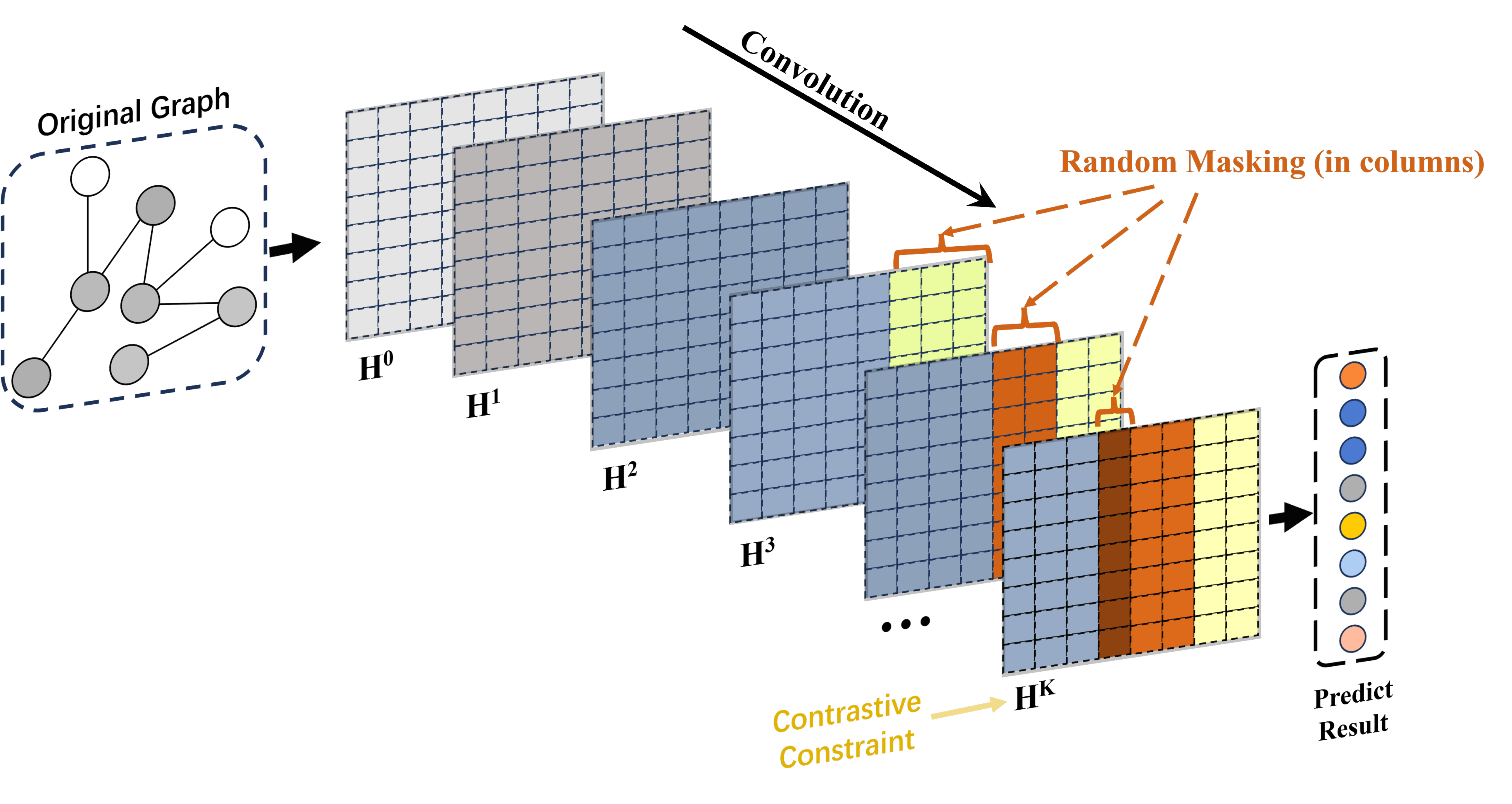}
		
	}%
	\caption{The overview of TSC applied to SGC and GCN. (a) illustrates the structure of TSC on SGC, while (b) depicts its application to GCN. They both add \textit{ random masking} to the columns of the representation matrix to mitigate representation convergence, and add \textit{contrastive constraints} to the rows of representation matrix to enhance node's individuality. }
	\label{fg:model}
\end{figure*}

\begin{equation}	
	{\textbf{H}^{(l)}}={\textbf{L}^{l}}\textbf{X},
\end{equation}
where ${\textbf{L}^{l}}$ represents the $l$ power of the Laplacian matrix. SGC, a single linear transformation across the graph structure, significantly reduces computational complexity and training time. This efficiency makes SGC particularly suitable for large-scale graph data. Benefiting from simplifying the model, SGC reduces the risk of overfitting and the gradient vanish problems. Compared with GCN, SGC preserves more of the original node features, preventing them from becoming indistinguishable, and consequently alleviating over-smoothing.
 

\section{Methodology}
The proposed two-sided constant includes random masking and contrastive constraint, that will implemented on SGC and GCN  as shown in Figure \ref{fg:model}. The random masking is conducted on the columns of the representation matrix, while the contrastive constraint is on the rows.

\subsection{Random Mask}
In the SGC model, using node features directly as their initial representations may lead to issues related to the sparse nature of features. This operation makes node representations to be affected by the sparsity of features, potentially resulting in overly sparse representations. Unlike SGC, GCN does not encounter this issue, owing to its inherent liner projection operation. To address this in SGC, we first perform a dimension reduction on the feature matrix $\textbf{X}$ with a linear projectorr, such as a Multilayer Perceptron (MLP),  to obtain the initial value of node representation. The projection operation can defined as
\begin{equation}
	\begin{aligned}
		\textbf{H}^{(0)}=MLP(\textbf{X}),\textbf{X}\in {\mathbb{R}^{n\times f}},\textbf{H}^{(0)}\in {\mathbb{R}^{n\times d}}.
	\end{aligned}
\end{equation}

\begin{figure}[t]	
	\centering
	\includegraphics[width=\columnwidth]{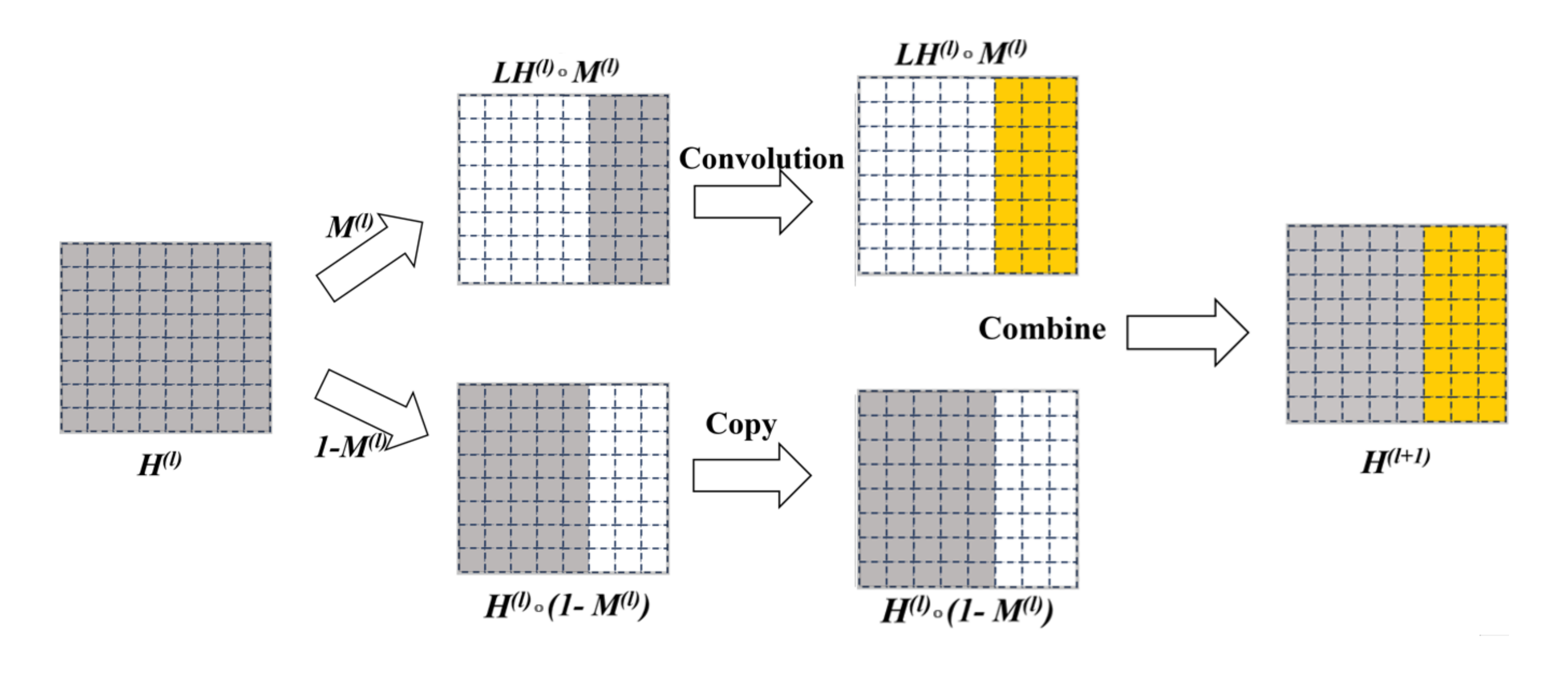}
	\caption{The Random Masking.}    
	\label{fg:masking}
\end{figure}

The random masking strategy on SGC, as illustrated by Figure \ref{fg:masking}, randomly masks the columns of the representation matrix with a masking rate, then performs graph convolution on the masked columns. The unmasked columns are directly copy to the next layer without convolutions. The masking strategy can be defined as 
\begin{equation}
	\textbf{H}^{(l+1)}=\textbf{L}{\textbf{H}^{(l)}} \circ {\textbf{M}^{(l)}}\ +{\textbf{H}^{(l)}} \circ (\textbf{1}-\textbf{M}^{(l)}),
\end{equation}
where $\textbf{M}^{(l)} \in \{0,1\}^{d\times d}$ is the mask matrix, $\textbf{M}_{i,:}^{(l)}$ is a $d$-dimensional row vector with all $0$ or $1$ in a masking rate, $\textbf{1}$ is a all $1$ matrix with the size of $d\times d$, $\circ$ is the element-wise matrix multiplication. If ${\textbf{M}_{i,:}}^{(l)}=1$, the $i$-th column of the representation matrix will  participate in the convolution, otherwise not.

Compared with DropNode \cite{Rong2020ICLR}, random masking filters the aggregated information in columns while DropNode drops messages before aggregation. Masking in columns can maintain all node's representation against convergence in layer-wise. Dropping nodes only discard few information form neighbors and update whole representation of each node. The excessive absorption of high-order neighbor information will lead to over-smoothing. However, random masking only update representations partially.

The neighbors in the shallow layer is very discriminative, as shown in Figure \ref{fg:node-neighbors}, and is useful for enhancing node representations, In this work, the masking strategy is only applied after layer \textit{3} ($l\geq3$). In deeper layers, nodes' neighbors become more similar, making over-smoothing more likely to occur. Therefore, we set masking rate gradually to decrease as the network layers increase. The  masking rate is as follows.
\begin{equation}
	\begin{aligned}
Masking Rate\left(\textbf{M}_{i,:}^{(l)}=0\right)=\left\{\begin{array}{l}\quad \quad 0 \quad \quad\quad \quad ,l\leq2\\1-\log(\lambda/l+1),l\geq3\end{array}\right.
	\end{aligned}
	\label{eq:random-mask}
\end{equation}
where $l$ is the number of layers and $\lambda>0$ is a hyper-parameter to tune the decreasing speed. A smaller value of $\lambda$ indicate faster decreasing. Meanwhile, to prevent the node representations from stopping updates as the structure becomes deeper. we enforce that each masking operation must mask at least one column, ensuring that the representations will always update.

\subsection{Contrastive Constraint }

Random masking effectively slows down feature updates to mitigate representation convergence, but it fails to enhance node individuality. This phenomenon has been discussed by Zhang \cite{zhang2022model_KDD}, that anti-over-smoothing technique can not always prevent GNN from performance degradation. This is because even after multiple rounds of iteration in high-order neighbor aggregation, nodes still lose their individuality and consequently become indistinguishable. To address this issue, we introduce contrastive constraints, aiming to optimize both the similarity between node features across layers and the dissimilarity between different nodes.

Specifically, in SGC, we consider identical node representations between two consecutive layers as positive pairs and add constraints to make them similar. Meanwhile, we consider the representations of any two different nodes within or between layers as negative pairs and push the nodes away from each other. The specific formula is as follows \cite{ICML2020SimCLR}:
\begin{equation}\small
	\label{eq:l-sgc}
	L_{SGC}=-\sum_{l=1}^{L-1}\sum_i^nlog\frac{e^{\left({s(h_i^{(l+1)},h_i^{(l)})/\tau}\right)}}{\displaystyle \sum_{j\neq i}^n e^{\left(s(h_i^{(l+1)},h_j^{(l+1)})/\tau\right)}+e^{\left(s(h_i^{(l+1)},h_j^{(l)})/\tau\right)}},
\end{equation}
where $n$ denotes the number of nodes, $l(l\geq0)$ is the number of layers, $\tau$ is the temperature coefficient, and $s(\cdot)$ is the cosine similarity calculation function. The numerator and denominator represent positive and negative pairs constraint respectively. 

In GCN, the computation complexity is larger than that of SGC. Adding contrastive constraints to each layer will further increase the computation complexity to make it impracticable in large scale graph. We generate the two subgraphs of the last layer by \textbf{\textit{dropout}} and by considering the same nodes of the two subgraphs as positive pairs and different nodes as negative pairs. This negative pair can keep the diversity between nodes and increase the difference of node information.

In GCN, its computational complexity is far exceeds that of SGC. Adding contrastive constraints to each layer would further make it impracticable in large dataset. Another difference between GCN and SGC is that in GCN, each layer's embedding is derived from the previous layer, whereas in SGC, it is directly computed from the initial representation. Therefore, we propose adding contrastive constraints only to the last layer, leveraging the transitivity of node representations across layers to constrain the representation of nodes in shallow layers. Specifically, we perform \textit{dropout} twice on the final layer $\textbf{H}^{(L)}$ to obtain two representations, $\widehat{\textbf{H}}^{(L)}$ and $\widetilde{\textbf{H}}^{(L)}$, then utilize contrastive constraints to enhance node individuality, similar to SGC. The calculation is as follows
\begin{equation}
	\boldsymbol {\widehat  H^{(L)}}=dropout(\boldsymbol H^{( L)})
\end{equation}
\begin{equation}
	\boldsymbol {\widetilde H^{(L)}}=dropout(\boldsymbol H^{(L)})
\end{equation}
\begin{equation}
	\label{eq:l-gcn}
	L_{GCN}=-\sum_i^nlog\frac{e^{\left(s(\widehat h_i^{(L)},\widetilde h_i^{(L)})/\tau\right)}}{\displaystyle\sum_{j\neq i}^ne^{\left(s(\widehat h_i^{(L)},\widehat h_j^{(L)})/\tau\right)}+e^{\left(s(\widehat h_i^{(L)},\widetilde h_j^{(L)})/\tau\right)}}
\end{equation}
where $L$ is the number of layers.

\section{Discussions}
In this section, we provide a theoretical analysis of our method to show that it can mitigate over-smoothing. And we also compare our method with other popular methods to illustrate the differences of our method. For detailed information, please refer to the Appendix B.

\subsection{Theoretical Analysis}

In the proposed method, we claim that the over-smoothing problem can be mitigated by using a Random Masking Strategy and a Contrastive Constraint. In this section, we analysis the claim.

\textbf{(a) Random masking strategy can retain the nodes' individual  information and mitigates representation convergence.}

Firstly, let us take SGC as an example to explain the over-smoothing problem. We know that the $l$-th layer of the SGC is

\begin{equation}
	\begin{aligned}
		\textbf{H}^{(l)}={{\boldsymbol L}^{l}} \boldsymbol X
	\end{aligned}
\end{equation}

 Since graph convolution is a special form of Laplacian smoothing, excessive message passing makes nodes with the same degree have identical representations. According to the derivation in \cite{mao2021UltraGCN}, the limit of message passing is as follows:
\begin{equation}
	\begin{aligned}
		\underset{l\to \infty }{\mathop{\lim }}\,(\boldsymbol L)_{ij}^{l}=\frac{\sqrt{({{d}_{i}}+1)({{d}_{j}}+1)}}{2m+n}
	\end{aligned}
\end{equation}
where $n$ and $m$ are the total number of nodes and edges in the graph, respectively. After GCN performs an infinite number of message passing steps, the features of node $i$ in the $l$-th layer will converge to the following form
\begin{equation}
	\begin{aligned}
		& \underset{l\to \infty }{\mathop{\lim }}\,\boldsymbol H_{i}^{(l)}=\underset{l\to \infty }{\mathop{\lim }}\,{({{\boldsymbol L}^{l}}\boldsymbol X)}_{i} \\ &=\frac{\sqrt{d_i+1}}{2m+n}\times\sum_{j\in V}^{}(\sqrt{d_j+1}\boldsymbol X_j)
	\end{aligned}
\end{equation}
As $l$ approaches infinity, the neighbors of node $i$ become all the nodes in the connected graph.

However, message passing with the proposed random mask strategy does not have such a issue. The message passing is as follows
\begin{equation}
	\begin{aligned}
		\boldsymbol H^{(l)}=\boldsymbol L\boldsymbol H^{(l-1)}\boldsymbol \circ \boldsymbol M^{(l)}\;+\;(\textbf{1}-\boldsymbol M^{(l)})\circ\boldsymbol H^{(l-1)}
	\end{aligned}
\end{equation}
When $l$ approaches infinity, we have
\begin{equation}
	\begin{aligned}
		& \underset{l\to \infty }{\mathop{\lim }}\,{{\boldsymbol H}^{(l)}}=\underset{l\to \infty }{\mathop{\lim }}\,\boldsymbol L \boldsymbol  X{{ \circ \boldsymbol M}^{(1)}}\ +\ (\textbf{1}-{{\boldsymbol M}^{(1)}}) \circ \boldsymbol X+{{\boldsymbol L}^{2}} \boldsymbol X\circ{{ \boldsymbol M}^{(2)}}\ +\\ & \ (\textbf{1}-{{\boldsymbol M}^{(2)}})\circ \boldsymbol L \boldsymbol X+...+{{\boldsymbol L}^{l}} \boldsymbol X{{\circ\boldsymbol M}^{(l)}}\ +\ (\textbf{1}-{{\boldsymbol M}^{(l)}})\circ{{\boldsymbol L}^{l-1}} X \\ 
		& =\underset{l\to \infty }{\mathop{\lim }}\,\boldsymbol X\sum\limits_{k=1}^{l-1}{({{ \boldsymbol M}^{(k)}}+(}1-{{\boldsymbol M}^{(k+1)}}))\circ{{\boldsymbol L}^{k}}+{{ \boldsymbol L}^{l}} \boldsymbol X\circ{{ \boldsymbol M}^{(l)}}+\\ &(\textbf{1}-{{\boldsymbol M}^{(1)}})\circ \boldsymbol X)
	\end{aligned}
\end{equation}
If we set ${{\widehat{ \boldsymbol M}}^{(k)}}={{ \boldsymbol M}^{(k)}}+(\textbf{1}-{{ \boldsymbol M}^{(k+1)}})$, $\overrightarrow{ \boldsymbol M}=\textbf{1}-{{ \boldsymbol M}^{(1)}}$, where both ${{\widehat{ \boldsymbol M}}^{(k)}}$ and $\overrightarrow{ \boldsymbol M}$ are mask matrices, the formula can be rewritten as
\begin{equation}
	\begin{aligned}
		& \underset{l\to \infty }{\mathop{\lim }}\,\sum\limits_{k=1}^{l-1}{{{\widehat{ \boldsymbol M}}^{(k)}}\circ{{ \boldsymbol L}^{(k)}} \boldsymbol X}+ {{ \boldsymbol M}^{(l)}}\circ{{ \boldsymbol H}^{(l)}}+\overrightarrow{ \boldsymbol M}\circ \boldsymbol X 
	\end{aligned}
\end{equation}

It is evident that the final representation of nodes encompasses information from all previous layers (${ \boldsymbol H}^{(k)}$) as well as from the first layer ($\boldsymbol X$). Information from the first layer typically possesses greater diversity and is effective in preserving the individual information. The masking rate can control the proportion of first layer information and information from all layers. Therefore, random masking can significantly mitigate the issue of over-smoothing.

We also apply our method to GCNs with nonlinear functions and linear transformations, and perform a theoretical analysis to mitigate over-smoothing. In GCNs, $f(\cdot)$ is non-linear function and $W^{(l)}$ is a linear transformation, we define the aggregrated information $\widetilde H^{(l)}$ at layer $l$ as
\begin{equation}
	\begin{aligned}
\boldsymbol {\widetilde  H}^{(l)}=f(\boldsymbol L \boldsymbol H^{(l-1)} \boldsymbol W^{(l)})
	\end{aligned}
\end{equation}
The column masking strategy in TSC will select columns from the current layer $\widetilde H^{(l)}$ and the last layer $H^{(l-1)}$
\begin{equation}
	\begin{aligned}
\boldsymbol H^{(l)}=\boldsymbol {\widetilde H}^{(l)}\circ \boldsymbol M^{(l)}+ \boldsymbol H^{(l-1)}\circ(\boldsymbol 1- \boldsymbol M^{(l)}), l>1
	\end{aligned}
\end{equation}
where $\boldsymbol H^{(0)}=\boldsymbol X$, $\boldsymbol H^{(1)}=\boldsymbol {\widetilde H}^{(1)}$. It should be noticed that the non-linear activation function only participates the information aggregation $\widetilde H^{(l)}$. That is when $H^{(l)}$ choose the columns from that last layer $H^{(l-1)}$, these columns are not affected by the non-linear function in the current layer. We can rewrite the column masking strategy in probability form. 
\begin{equation}
	\begin{aligned}
\boldsymbol H_{\cdot,j}^{(l)}=\left\{\begin{array}{lc} \boldsymbol {\widetilde H}_{\cdot,j}^{(l)}&if\;a>\alpha^{(l)} \\ \boldsymbol H_{\cdot,j}^{(l-1)}&else\end{array}\right.
	\end{aligned}
\end{equation}
where $\alpha^{(l)}=1-\log(\frac\lambda{l}+1)$ (Defined in Eq.(5) ), $a$ is an uninform random variable sampled form $a\sim U\lbrack0,1\rbrack,$ $l$ is layer number. Then ,we have the following probability
\begin{equation}
	\begin{aligned}
&P\left( \boldsymbol H_{\cdot,j}^{(l)}=\boldsymbol {\widetilde H}_{\cdot,j}^{(l)}\right)=1-\alpha^{(l)}\\
&P\left( \boldsymbol H_{\cdot,j}^{(l)}=\boldsymbol H_{\cdot,j}^{(l-1)}\right)=\alpha^{(l)}
	\end{aligned}
\end{equation}

At the layer  $l+1$, we have 
\begin{equation}
	\begin{aligned}
&\boldsymbol H_{\cdot,j}^{(l+1)}=\left\{\begin{array}{lc} \boldsymbol {\widetilde H}_{\cdot,j}^{(l+1)}&if\;a>\alpha^{(l+1)}\\ \boldsymbol H_{\cdot,j}^{(l)}&else\end{array}\right. \\
&=\left\{\begin{array}{lc} \boldsymbol {\widetilde H}_{\cdot,j}^{(l+1)}&if\;a>\alpha^{(l+1)}\\ \boldsymbol {\widetilde H}_{\cdot,j}^{(l)}&if \; a\leqslant\alpha^{(l+1)} \wedge \tilde{a} > \alpha^{(l)} \\ \boldsymbol H_{\cdot,j}^{(l-1)}&  else \end{array}\right.
	\end{aligned}
\end{equation}
where both $a$ and $\tilde{a}$ are random variables sampled from $U\lbrack0,1\rbrack$. We then have
\begin{equation}
	\begin{aligned}
&P\left( \boldsymbol H_{\cdot,j}^{(l+1)}=\boldsymbol {\widetilde H}_{\cdot,j}^{(l+1)}\right)=1-\alpha^{(l+1)}\\
&P\left( \boldsymbol H_{\cdot,j}^{(l+1)}=\boldsymbol {\widetilde H}_{\cdot,j}^{(l)}\right)=(1-\alpha^{(l)})\alpha^{(l+1)}\\
&P\left( \boldsymbol H_{\cdot,j}^{(l+1)}= \boldsymbol H_{\cdot,j}^{(l-1)}\right)=\alpha^{(l)}\alpha^{(l+1)}
	\end{aligned}
\end{equation}

By generalizing the above equations, we have
\begin{equation}
	\begin{aligned}
&P\left(  \boldsymbol H_{\cdot,j}^{(L)}= \boldsymbol H_{\cdot,j}^{(1)}\right)=\prod_{l=1}^L\alpha^{(l)}
=\prod_{l=1}^L\left(1-\log(\frac\lambda{l}+1)\right)\\
&\geq\prod_{l=1}^L\left(1-\frac\lambda{l}\right)
\geq \exp(-\sum_{l=1}^L\frac\lambda{l}-\gamma\sum_{l=1}^L(\frac\lambda{l})^2)\\
&> exp(-\lambda\log{L}-\gamma\frac{\pi^2}{6}\lambda^2)
	\end{aligned}
\end{equation}
where the first inequality is obtained from $\log (x+1)\leq x$ for $x>0$, the second inequality is obtained by  $1-x \geq exp(-x-\gamma x^2)$ for $x\geq 0$ with $\gamma>0.5$ being an arbitrary constant, the third inequality comes from two inequalities $\sum_{l=1}^L\frac{1}{l}\leq \log L$ and partial sum of the Basel Series $\sum_{l=1}^{L}(\frac{1}{l})^2<\frac{\pi^2}{6}$.

We can see that $P\left(\boldsymbol H_{\cdot,j}^{(L)}= \boldsymbol H_{\cdot,j}^{(1)}\right)$ is close to $1$ when the hyperparameter $\lambda=0$.  We can control the columns in the last layer comes form the initial layer by $\lambda$. This demonstrate TSC can effectively control the over-smoothing. We also provde the parameter analysis in Fig a1 to demonstrate its effectiveness. When  $\lambda \in\{0.1,\;0.5\}$   the ACC outperform than when $\lambda \in \{1,\; 1.5\}$. This revelas that more information form shallow layers can obtain better performance. However, when $\lambda=0.1$ its performance is lower than when $\lambda=0.5$. This is that overwhelm information form ininital layers will degrdata perforamcne, because remote information is not well aggegrated.

We observe that $P\left(\boldsymbol H_{\cdot,j}^{(L)} = \boldsymbol H_{\cdot,j}^{(1)}\right)$ approaches $1$ when the hyperparameter $\lambda = 0$. This allows us to control the flow of information from the initial layer to the last layer using $\lambda$. This demonstrates that column masking can effectively mitigate the issue of over-smoothing.

We provide a parameter analysis in figure.\ref{fg:param-lambda} to illustrate the effectiveness of different $\lambda$ values. As shown in Figure 1a, when $\lambda$ is set within $\{0.1, 0.5\}$, the accuracy (ACC) outperforms the model when $\lambda \in \{1,\; 1.5\}$. This suggests that incorporating  information from the shallower layers can yield better performance.

\textbf{(b) Contrastive constraint enhance node individuality and difference among nodes. }

The core idea of contrastive learning is to maximize agreement between augmented views of the positive pairs and disagreement of views from  negative pairs. Taking the contrastive loss on SGC as an example, we can decompose the $L_{SGC}$ at $(l+1)$-th layer $i$-th node into two parts, called \textbf{alignment loss} and \textbf{heterogeneity loss} respectively. 
\begin{equation}\label{eq:lalign}
	\begin{aligned}
		l^{(l+1)}_{align}(i)=s(h_i^{(l+1)},h_i^{(l)})/\tau\;\;
	\end{aligned}
\end{equation}

\begin{equation}\label{eq:lheter}
	\begin{aligned}
		l^{(l+1)}_{heter}(i)=\log(\sum_{j\neq i}^ne^{s(h_i^{(l+1)},h_j^{(l+1)})/\tau}+e^{s(h_i^{(l+1)},h_j^{(l)})/\tau})
	\end{aligned}
\end{equation}
The alignment loss promotes similar feature representations for the same nodes in different layers, thus maintaining invariance to unwanted noise and increasing consistency across layers. The heterogeneity loss effectively prevents nodes from becoming indistinguishable. It maximizes the average distance between all samples, resulting in representations that are roughly evenly distributed in the latent space, thus retaining more information. Consequently, it can mitigate the effects of over-smoothing in GNNs. Similar analysis can be easily obtained in GCNs by Equation (\ref{eq:lalign}) and (\ref{eq:lheter}).

%

\section{Experments}
In this section, we conduct experiments on semi-supervised node classification to evaluate the effectiveness of TSC-GCN and TSC-SGC. Five recent developed models for over-smoothing are chosen for evaluations. We use accuracy (ACC) to assess the performance of models and Mean Average Distance (MAD) \cite{Chen2020AAAI} to measure the degree of smoothing of models. Our code is available at: \url{https://github.com/Recgroup/TSC}.

\subsection{Experimental Setup}

\hspace{1em}\textbf{Datasets}. To evaluate the performance of our proposed method across different  scenarios, we conduct experiments on 5 datasets with various sizes and densities: \textit{Coauthor CS } \cite{shchur2019pitfalls}, \textit{Amazon Photo }\cite{shchur2019pitfalls}, \textit{Cora} \cite{McCallum2000InformationRetrieval,Sen2008AI}, \textit{CiteSeer} \cite{Giles1998CiteSeer,Sen2008AI}, and \textit{Pubmed }\cite{namata2012query,Sen2008AI}. The summary of datasets is provided in appendix A.

\textbf{Baselines}. To assess the advantages of the proposed TSC-GCN against the SOTAs, we compare our method with other GNNs. 

\textbf{GCN} \cite{berg2017graph} is the first work to use GCN for node classification.

\textbf{SGC} \cite{Wu2019SGCNICML} is a simplified graph convolutional network that removes the nonlinear activation function and many linear transformations.

\textbf{DropMessage} \cite{Fang2023AAAI} alleviates over-smoothing implicitly by adding dropout operation to  out messages of each node. 

\textbf{ContraNorm} \cite{Guo2023ICLR} mitigates over-smoothing by implicitly shattering node representations into a more uniform distribution.

\textbf{GCNII} \cite{Chen2020ICML} enhance nodes' individual information by adding the first layer's features to deep layers. 

\textbf{AIR} \cite{zhang2022model_KDD} adds the node representation from previous layers to next layer for individuality enhancement.

\textbf{NDLS} \cite{zhang2021NIPS} alleviates over-smoothing by controlling node-dependent local smoothness. 

\textbf{DropEdge} \cite{rong2019dropedge} mitigates over-fitting and over-smoothing by randomly censoring out edges in the original graph during model training.

\textbf{SJLR} \cite{giraldo2023trade} proposes Rewiring algorithm to mitigate over-smoothing and over-squeezing by adding and removing edges in a feasible way. 

\textbf{DeepGCN} \cite{li2019deepgcns} applys residual/dense concatenation and dilated convolution to the GCN architecture to achieve significant performance gains in point cloud semantic segmentation tasks.

\subsection{Accuracy Evaluation}
We first compare the overall classification performance of TSC against all baseline methods. We vary the number of layers in all models from 1 to 32 and list their optimal accuracy (ACC) in Table \ref{tb:over-performance}. Notably, GCN doesn't use and anti-over-smoothing technique, represents the best performance of a standard GCN. SGC, through model simplification, possesses some capability to prevent over-smoothing, but it sacrifices model expressiveness.
 
\begin{table}[hbt!]
	\centering
	\caption{Overall performance comparison. The classification performance is measured by ACC. The best and second best are marked with bold and underline respectively.}
	\resizebox{\linewidth}{!}{
	
		\begin{tabular}{lcccccc}
			\toprule
			& Cora & Citeseer & Pubmed & {\tiny CoauthorCS} & {\tiny AmazonPhoto}\\
			\toprule
			GCN [ICLR'17]     & 0.836   & 0.719  & 0.798  & 0.910  &  0.914       \\
			SGC [ICML'19]        & 0.828  &  0.740  & 0.79  &  0.914   & 0.905    \\
			DropEdge(GCN) [ICLR'20]        & 0.828  &  0.723  & 0.796  &  --   & --    \\
			SJLR [CIKM'23]        & 0.819  &  0.695  & 0.786  &  --   & --    \\
			DeepGCN [ICCV'19]        & 0.734  &  0.627  & 0.757  &  --   & --    \\
			ContraNorm(GCN) [ICLR'23]     & 0.837   & 0.728  &  0.798  & 0.920  &  0.912       \\
			{\small DropMessage(GCN)} [ICLR'23]     & 0.835   & 0.721  &  0.795  & 0.911  &  0.915      \\
			GCNII [ICML'20]     & \underline{0.852}      & 0.744  & 0.803  &  0.921   & 0.929   \\
			AIR(SGC) [KDD'22]     & 0.828    &  0.746  &  0.790  & \textbf{0.936}  &  0.93      \\		
			NDLS [NIPS'21]     & 0.841    &  0.736  &  \underline{0.807}  &  0.920  & 0.920      \\
			\toprule
			GCN+TSC(ours)  & 0.851  & \textbf{0.748}  & \underline{0.807}  & \underline{0.926}  & \textbf{ 0.934}\\
			SGC+TSC(ours)     & \textbf{0.854}   & \underline{0.747}  &  \textbf{0.825}  & 0.916  &  \underline{0.933}     \\
			\bottomrule
		\end{tabular}
	}
	
	\label{tb:over-performance}
\end{table}
From Table 1, the following conclusions can be drawn:
\textbf{(1)} Although GCN is prone to over-smoothing, it possesses strong model expressiveness and can achieve good performance at shallow layers. For instance, it outperforms SGC on datasets such as Cora, Pubmed, and AmazonPhoto.
\textbf{(2)} SGC can alleviate over-smoothing and overfitting, which is why it can surpass GCN on the Citeseer and CoauthorCS datasets. Enhancing SGC's capability to address over-smoothing can further obtain better performance. For example, ContraNorm achieves better results than SGC on Cora, Pubmed, CoauthorCS, and AmazonPhoto.
\textbf{(3)} Node filtering methods are primarily aimed at mitigating over-smoothing, whereas individuality enhancement methods aim to improve node uniqueness. As a result, the latter can achieve relatively better performance. For instance, GCNII and AIR obtain better results than ContraNorm and DropMessage on the Citeseer, CoauthorCS, and AmazonPhoto datasets. GCNII shows superior performance to ContraNorm and DropMessage on Cora and Pubmed datasets.
\textbf{(4)} By further optimizing the balance between node's individual information and local smoothness, NDLS achieves superior results to AIR on Cora and Pubmed.
\textbf{(5)} Considering both neighbor filtering and node individuality enhancement, model performance can be further improved. The proposed TSC, combined with SGC and GCN, achieves optimal and sub-optimal performance on Citeseer, Pubmed, and AmazonPhoto datasets. It obtains optimal and sub-optimal results on Cora and CoauthorCS, respectively.

\subsection{Over-Smoothing Analysis}
In order to verify the effectiveness of our approach on the over-smoothing problem, we report model's ACC when depth is continuously increased from 4 to 32. The results on Cora, Citeseer, Pubmed, CoauthorCS, and AmazonPhoto are reported in Table \ref{tb:acc at cora} to \ref{tb:acc at AmazonPhoto}. In these tables, OOM indicates that out of memory, and the best performance in each table is denoted by $^*$. 

Observations from Tables  \ref{tb:acc at cora} to \ref{tb:acc at AmazonPhoto} can be summarized as follows:
\textbf{(1)} GCN is severely impacted by over-smoothing, whereas SGC demonstrates a certain ability to alleviate this issue, showing a more gradual performance decline as network depth increases.
\textbf{(2)} Among neighbor filtering methods, the explicit filtering approach \textit{DropMessage} experiences significant over-smoothing at the 32nd layer, while the implicit filtering method \textit{ContraNorm} maintains stable performance at this depth.
\textbf{(3)} Individuality enhancement methods consistently exhibit strong resistance to over-smoothing. For instance, \textit{GCNII} achieves its best performance on Cora and Citeseer at the 32nd layer, fulfilling the expectation that deeper networks yield better results.
\textbf{(4)} Our proposed TSC, when combined with GCN and SGC, shows effective mitigation of over-smoothing. For example, GCN+TSC outperforms baselines and SGC+TSC in deeper layers on the Citeseer, CoauthorCS, and AmazonPhoto datasets, indicating that this method not only effectively reduces over-smoothing but also leverages the expressive power of GCN.
\textbf{(5)} All networks generally reach optimal or near-optimal results by the 4th or 8th layer. Therefore, in practical applications, to reduce computational complexity, it is advisable to limit network depth to 10 layers or fewer. This approach aligns with the small-world phenomenon \cite{milgram1967small} observed in complex networks.

\begin{table}[hbt!]
	\centering
	\caption{ACC comparison in different depth on Cora}
	\resizebox{0.9\linewidth}{!}{
		\begin{tabular}{ccccc}
			\toprule
			& Layer 4 & Layer 8 & Layer 16 & layer 32 \\
			\toprule
			GCN     & 0.818   & 0.303  & 0.311  & 0.319      \\
			SGC     & 0.828   & 0.815  &  0.790   &  0.725   \\
			ContraNorm(GCN)  & 0.824   &  0.795  & 0.520   & 0.289  \\		
			DropMessage(GCN)     & 0.836   & 0.801  &  0.434  & 0.304       \\
			GCNII & 0.840 & 0.829 & 0.838 & \textbf{0.852} \\
			AIR(SGC)    & 0.795  & 0.824  & 0.828  & 0.796  \\
			NDLS   &  0.814  &  0.811 &   0.841 &  0.834  \\
			\toprule
			GCN+TSC    & \textbf{0.845}  & \underline{ 0.847 } & \underline{0.851}   &  0.848    \\
			SGC+TSC    &  \underline{0.843 }  & \textbf{ 0.849 }  & \textbf{ 0.854 }$^{*}$  &  \underline{0.851 }  \\ 
			\bottomrule
	\end{tabular}
}
	\label{tb:acc at cora}
\end{table}
\begin{table}[hbt!]
	\centering
	\caption{ACC comparison in different depth on Citeseer}
	\resizebox{0.9\linewidth}{!}{
		\begin{tabular}{ccccc}
			\toprule
			& Layer 4 & Layer 8 & Layer 16 & layer 32 \\
			\toprule
			GCN     & 0.707   & 0.260  & 0.254  & 0.235      \\
			SGC    & 0.739   &  0.740  &  0.733  &  0.724   \\
			ContraNorm(GCN)  & 0.718  &  0.620  &  0.491  &  0.287  \\
			DropMessage(GCN)     & 0.711   & 0.701  &  0.550  & 0.321      \\
			GCNII  &  0.720 & 0.734 & 0.738 & \textbf{0.744} \\
			AIR(SGC)   &  0.731  &  0.743  & \textbf{ 0.746} &    0.739  \\			
			NDLS   &  0.713  &  0.730 &  0.736 & 0.723  \\			
			\toprule
			GCN+TSC     & \textbf{0.742}  &  \textbf{0.748$^*$}  & \underline{0.743 }   &  \textbf{0.744}   \\
			SGC+TSC    &  \underline{0.740}   &  \underline{0.747 }  & \underline{ 0.743 }  &  \underline{ 0.742 }  \\ 
			\bottomrule
	\end{tabular}
}	
	\label{tb:acc at citeseer}
\end{table}
\begin{table}[hbt!]
	\centering
	\caption{ACC comparison in different depth on Pubmed}	
	\resizebox{0.9\linewidth}{!}{
			\begin{tabular}{ccccc}
					\toprule
					& Layer 4 & Layer 8 & Layer 16 & layer 32 \\
					\toprule
					GCN     & 0.769   & 0.635  & 0.413  & 0.419      \\
					SGC    &  0.761  &  0.737  &   0.719   &  0.702  \\
					ContraNorm(GCN)   &   0.782  & OOM & OOM & OOM  \\
					DropMessage(GCN)     & 0.776   & 0.783  &  0.604  & 0.621      \\
					GCNII  &  0.792  & 0.801 & 0.801 & 0.803  \\
					AIR(SGC)  &  0.790  & 0.785  &  0.765  &  0.738  \\
					NDLS   &  0.790  & \underline{ 0.802 }&  \underline{  0.807} &  \underline{ 0.804}  \\
					\toprule
					GCN+TSC     & \underline{0.806}  & 0.798  & \underline{0.807}    &  0.800    \\
					SGC+TSC    & \textbf{ 0.817$^*$ }  & \textbf{ 0.812 }  & \textbf{ 0.813 }  &  \textbf{ 0.809 }  \\ 
					\bottomrule
			\end{tabular}
		}
	\label{tb:acc at pubmed}
\end{table}
\begin{table}[hbt!]
	\centering
	\caption{ACC comparison in different depth on CoauthorCS}	
	\resizebox{0.9\linewidth}{!}{
			\begin{tabular}{ccccc}
					\toprule
					& Layer 4 & Layer 8 & Layer 16 & layer 32 \\
					\toprule
					GCN    & 0.891  & 0.257  & 0.184  & 0.139      \\
					SGC    &  0.889  &  0.875  &   0.843   &  0.737  \\
					ContraNorm(GCN)   & 0.910   &  OOM  &   OOM  &  OOM    \\
					DropMessage(GCN)     & 0.902   & 0.653  &  0.424  & 0.424      \\
					GCNII & 0.910  &  \underline{0.921}  & \underline{0.920} & \underline{0.913} \\
					AIR(SGC)  &  \textbf{0.936$^*$} &   OOM  &  OOM   &   OOM  \\
					NDLS  &  0.917  &  0.920 &  0.837 &  0.711  \\
					\toprule
					GCN+TSC & \underline{ 0.925 } & \textbf{ 0.926 } & \textbf{0.926}   &  \textbf{0.922}    \\
					SGC+TSC &   0.916   &  0.912   &  0.913   &  0.910   \\ 
					\bottomrule
			\end{tabular}
		}
		\label{tb:acc at CoauthorCS}
\end{table}
\begin{table}[hbt!]
	\centering
	\caption{ACC comparison in different depth on {\small AmazonPhoto}}
	\resizebox{0.9\linewidth}{!}{
	\begin{tabular}{ccccc}
	\toprule
	& Layer 4 & Layer 8 & Layer 16 & layer 32 \\
	\toprule
	GCN     & 0.906   & 0.874  & 0.829  & 0.580      \\
	SGC     & 0.907   & 0.900   & 0.899  &  0.887    \\
	ContraNorm(GCN)     & 0.912   & 0.893  &  0.868 & 0.578       \\	
	DropMessage(GCN)     & 0.915   & 0.894  &  0.864  & 0.866      \\
	GCNII  &  0.929  & 0.922  & \underline{0.925} & \underline{0.926} \\
	AIR(SGC)  &  0.930  & \underline{ 0.929 } &  0.918  & 0.918 \\
	NDLS   &  0.917  &  0.920 &  0.891 &  0.890  \\
	\toprule
	GCN+TSC    & \textbf{0.934$^*$}  &   \textbf{0.934$^*$}  & \textbf{0.931}    & \textbf{ 0.929 }   \\
	SGC+TSC    &  \underline{0.933 }  & \underline{ 0.929 }  &  0.923   &   0.922   \\ 
	\bottomrule
	\end{tabular}
}	
	\label{tb:acc at AmazonPhoto}
\end{table}

\begin{figure}[hbt!]	
	\includegraphics[width=0.9\linewidth, draft=false]{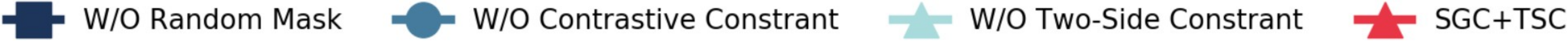}
	\vspace*{-0.1cm}	
	\subfigure[ACC on Cora]{
		\label{fg:ablation-acc-cora}	
			\centering
			\includegraphics[width=0.45\linewidth,draft=false]{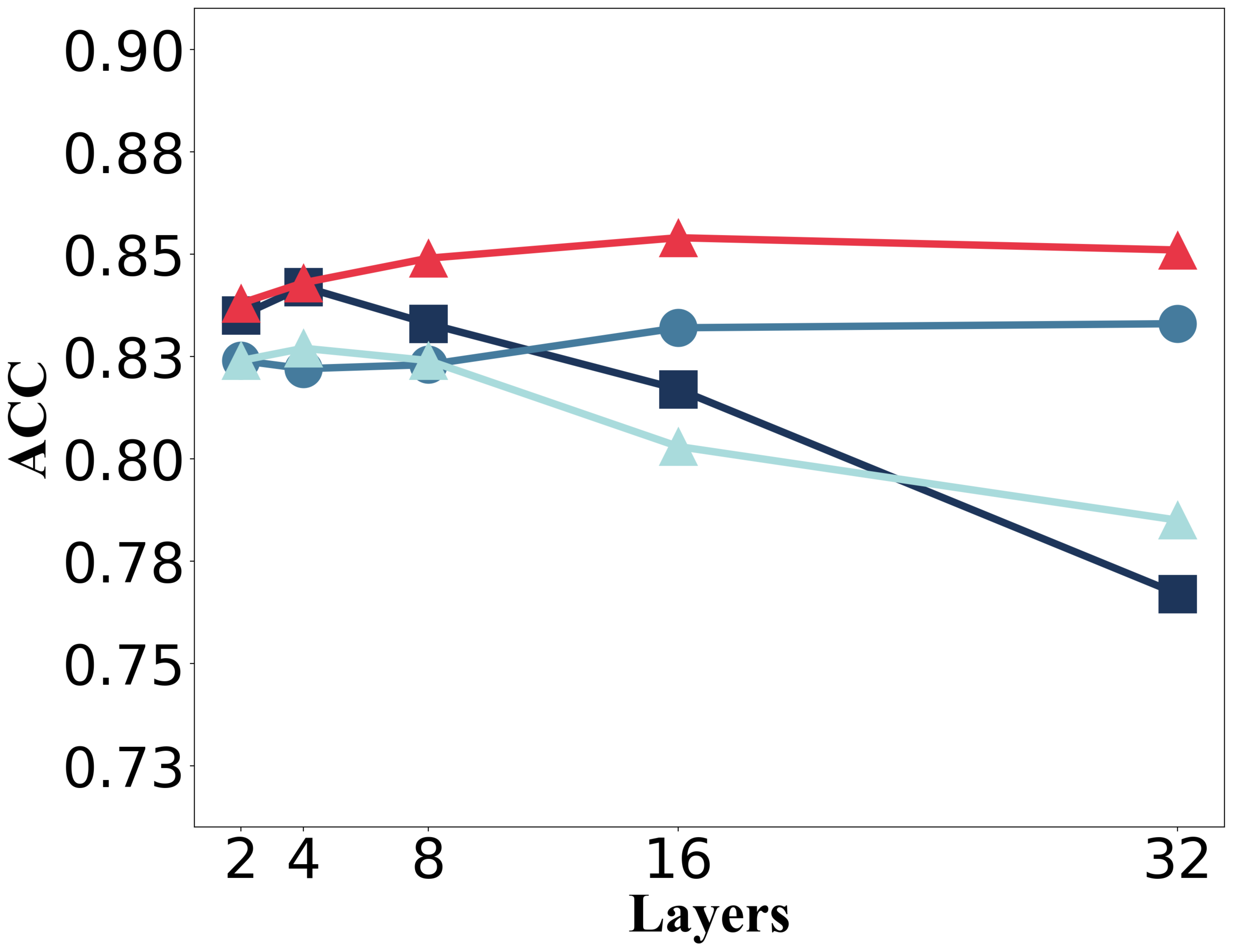}
		}
	\subfigure[ACC on Citeseer]{
		\label{fg:ablation-acc-citeseer}	
			\centering
			\includegraphics[width=0.45\linewidth]{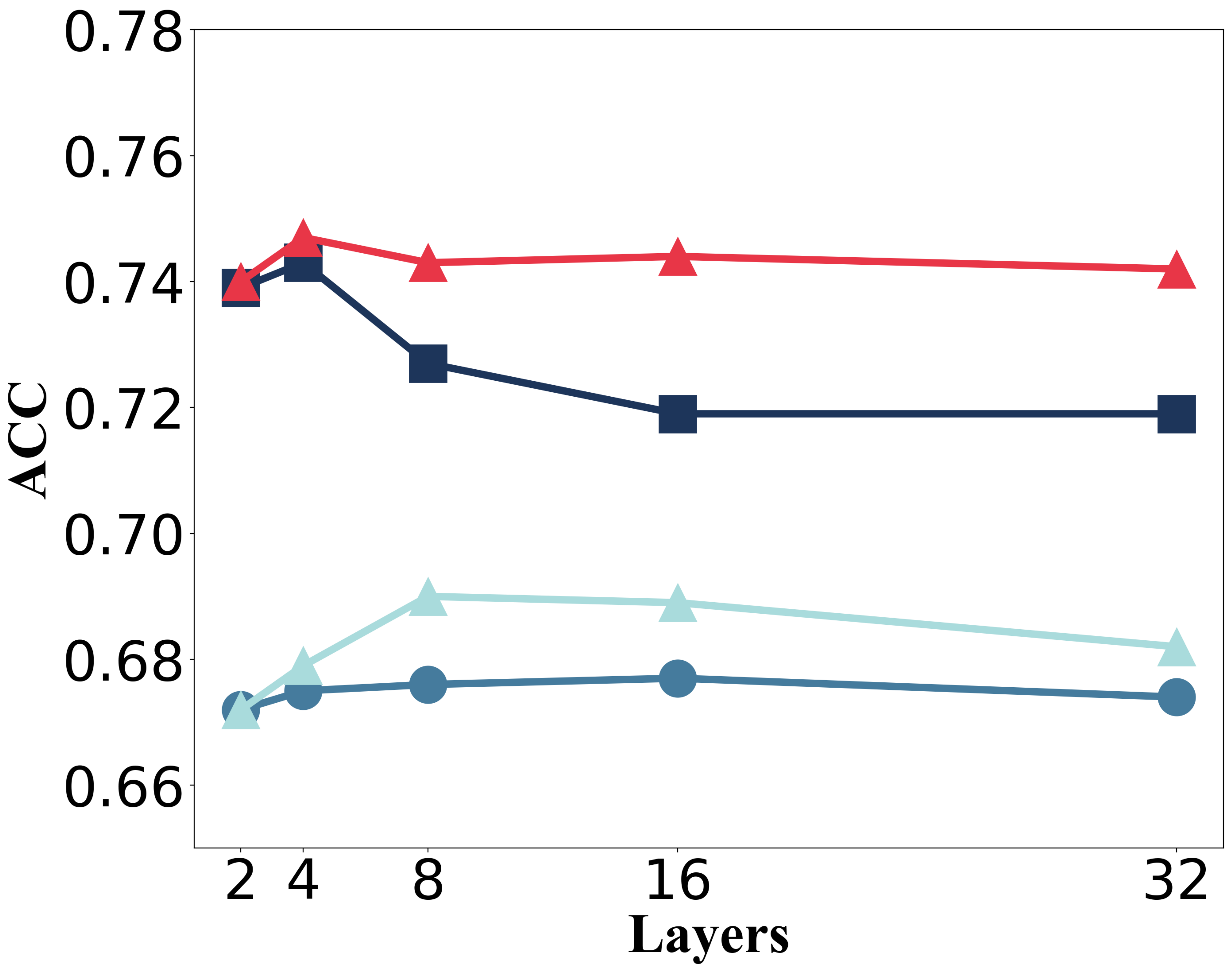}
		}	
	\vspace*{-0.5cm}
	\subfigure[MAD on Cora]{
		\label{fg:ablation-mad-cora}	
			\centering
			\includegraphics[width=0.46\linewidth]{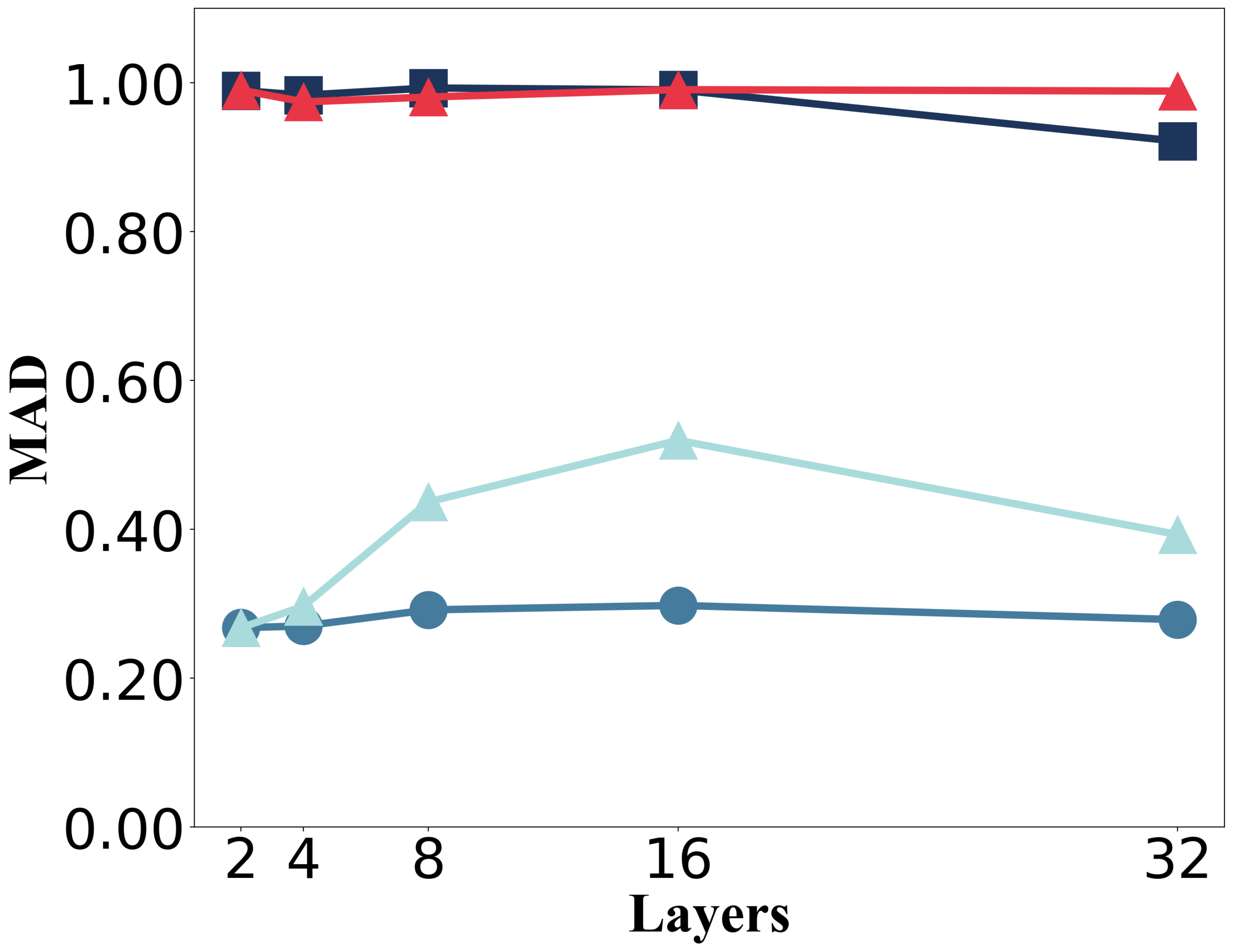}
		}
	\subfigure[MAD on Citeseer]{
		\label{fg:ablation-mad-citeseer}
			\centering
			\includegraphics[width=0.45\linewidth]{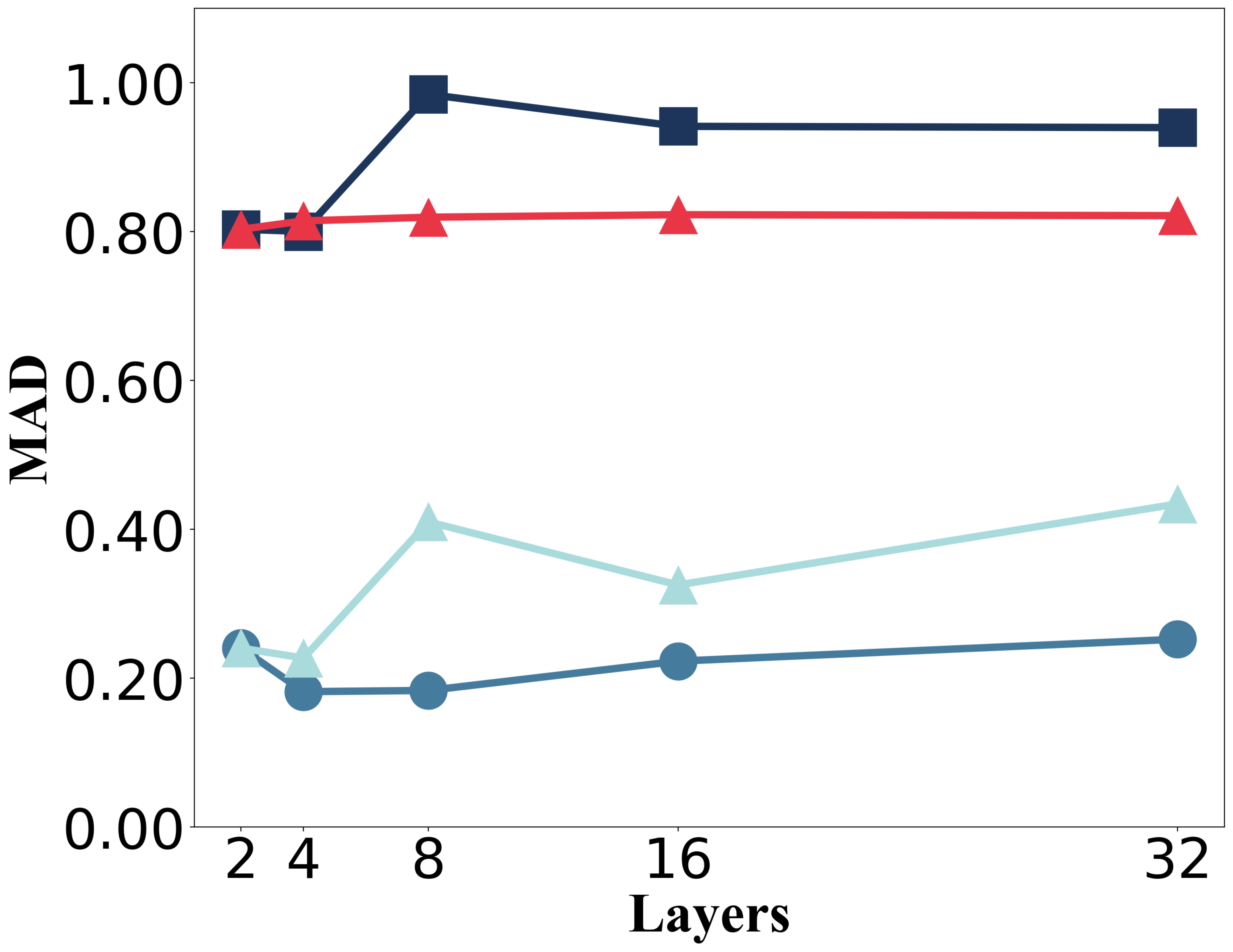}
		}
	\caption{The ablation study. The 1st row depicts the ablation results on ACC metric and the 2nd row is the result on MAD. }
	\label{fg:ablation}	
\end{figure}

\subsection{Ablation Study}
We take SGC+TSC as an example to evaluate the impact of random mask and contrastive constraint on Cora and Citeseer datasets by ablation study. Along with ACC, we also adopt MAD \cite{Chen2020AAAI}  to measure  the degree of smoothing of node representation. The ablation studies are reported in Figure \ref{fg:ablation}. The ablation operations is as  

\textit{W/O Random Mask}: removing random mask from SGC+TSC,

\textit{W/O Contrastive Constraint}: removing Contrastive Constraint strategy from SGC+TSC,

\textit{W/O Two-Side Constraint}: removing random mask and contrastive constraint from SGC+TSC, which is just the SGC model.

Comparing Figure \ref{fg:ablation-acc-cora} and \ref{fg:ablation-acc-citeseer}, we can seen that removing random masking ACC will slowly decrease with the increase of layers. This indicates that random mask can alleviate  representation convergence but not for performance degradation.  Removing contrastive contrast, the acc is gradated but keep stable with the increase of layers. This indicates that contrastive can significantly maintain node's difference, and therefore improve the overall performance.

Comparing Figure \ref{fg:ablation-mad-cora} and \ref{fg:ablation-mad-citeseer}, it can be found that both random mask and contrastive constraint can alleviate the representation convergence ($MAD=0$ indicates over-smoothing happen), although they have different MAD values. It is notable that removing both random mask and contrastive constraint, GCN+TSC can still keep MAD off zeros, because SGC removing mapping weights that can alleviate over-smoothing.

\subsection{Visualization}
\begin{figure}[hbt!]	
	\centering	
	\subfigure[GCN]{
		\centering
		\includegraphics[width=0.3\linewidth]{figs/1.3GCNlayer32.pt}
	}
	\subfigure[GCN+DropMess.]{
		\centering
		\includegraphics[width=0.3\linewidth]{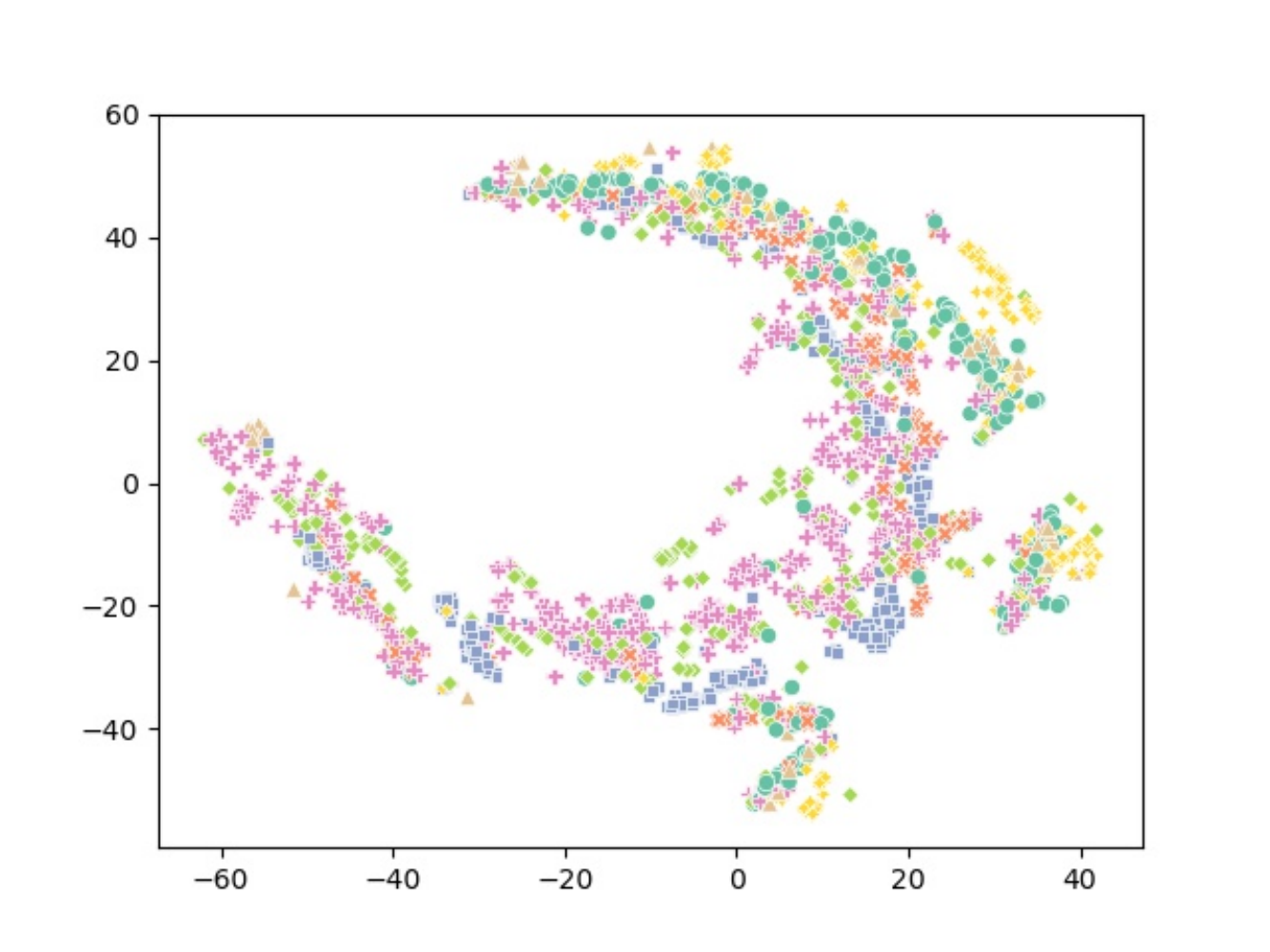}
	}	
	\subfigure[GCN+TSC(ours)]{
		\centering
		\includegraphics[width=0.3\linewidth]{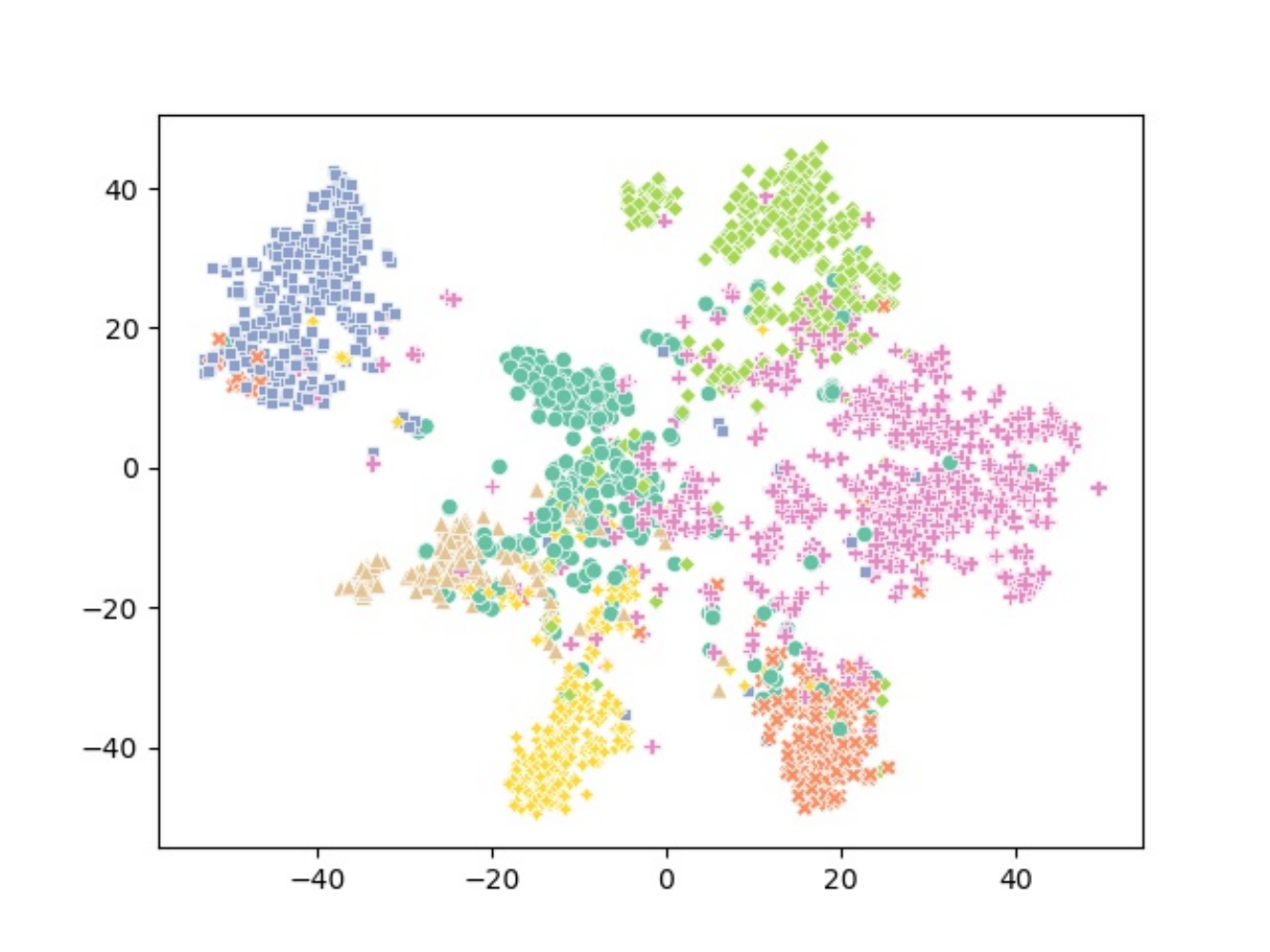}
	}	
	\subfigure[SGC]{
		\centering
		\includegraphics[width=0.3\linewidth]{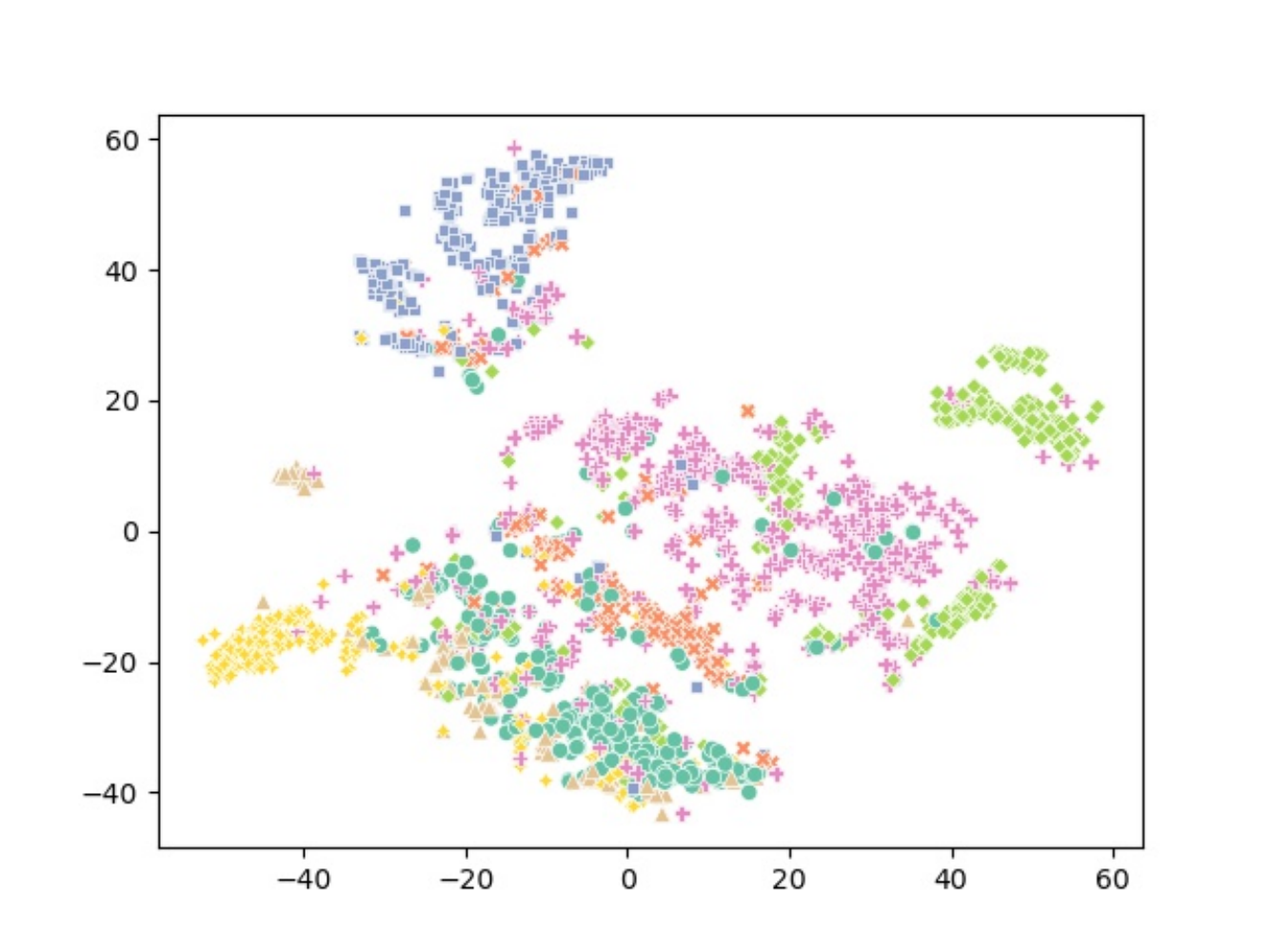}
	}
	\subfigure[SGC+AIR]{
		\centering
		\includegraphics[width=0.3\linewidth]{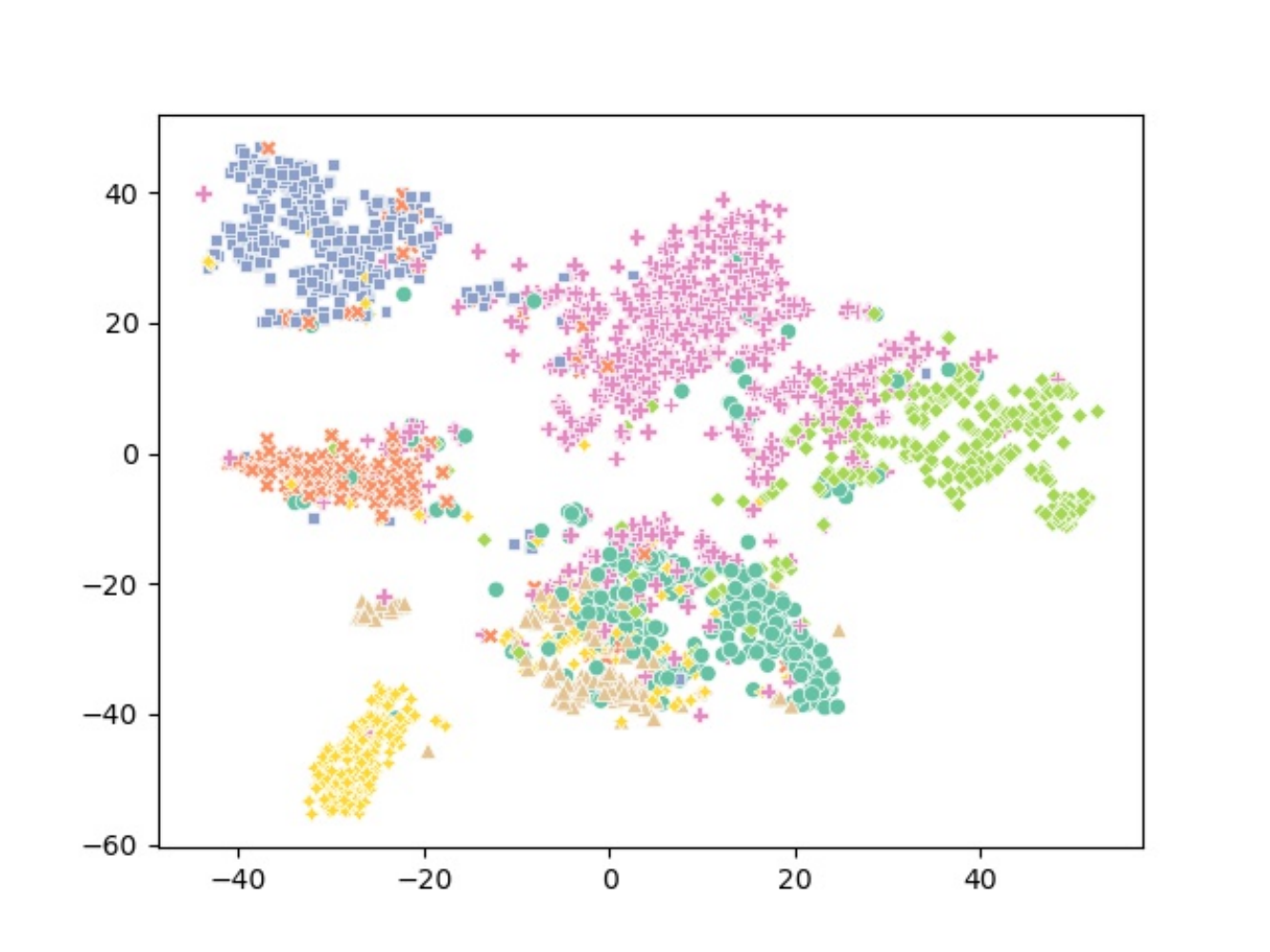}
	}
	\subfigure[SGC+TSC(ours)]{
		\centering
		\includegraphics[width=0.3\linewidth]{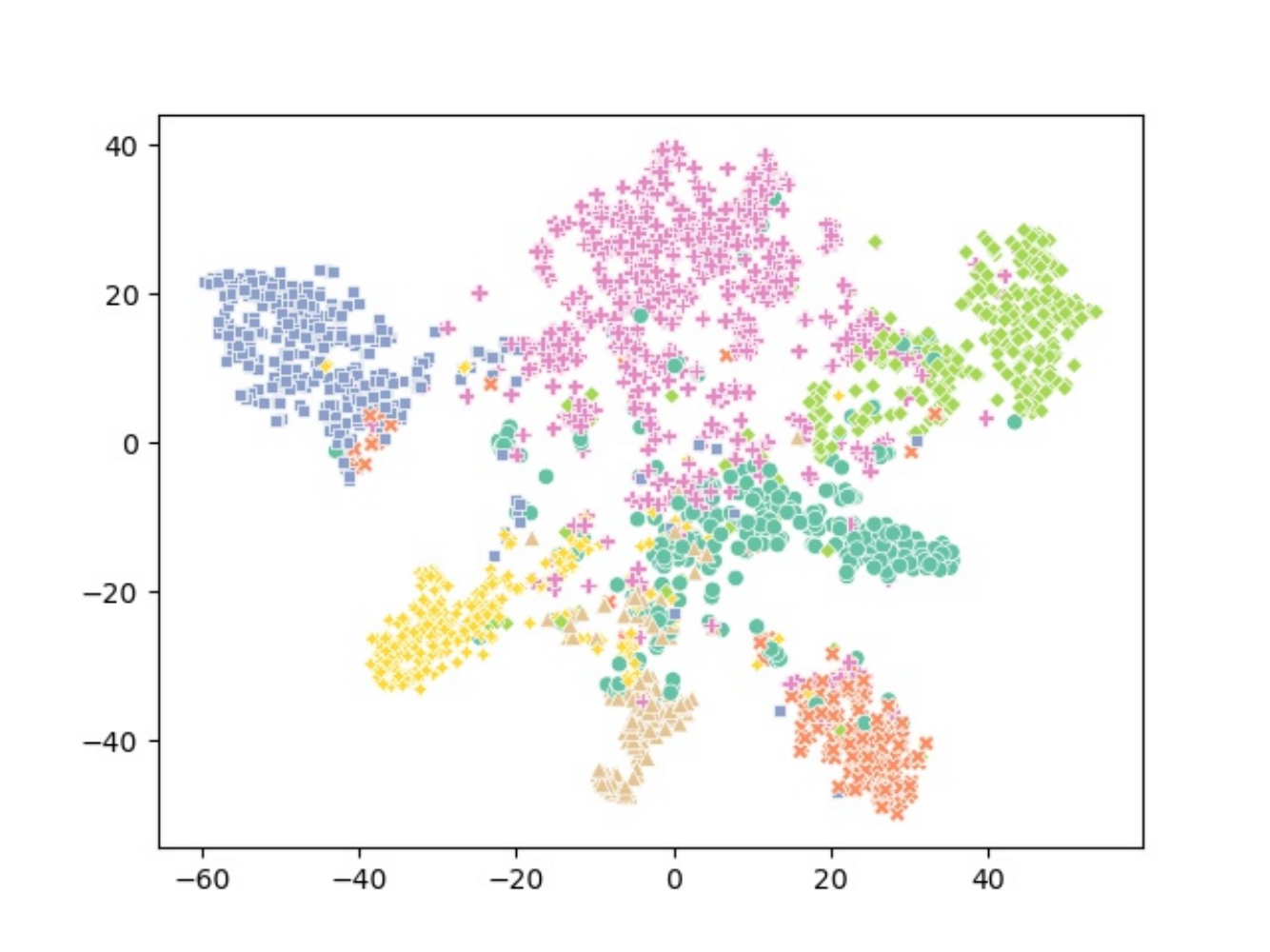}
	}
	\caption{The 2D visualization of node representations in layer 32. Nodes are colored to represent different classes. }
	\label{fg:vis}
\end{figure}

We employ t-SNE \cite{Laurens2008tSNE} to visualize the node representations in layer 32 on Cora. The results for GCN,  GCN+DropMessage, GCN+TSC, SGC, SGC+AIR, and SGC+TSC are presented in Figure \ref{fg:vis}.

In the first row of Figure \ref{fg:vis}, all GCN-based models except GCN+TSC exhibit over-smoothing, resulting in indistinguishable node representations in the two-dimensional embedding space. This highlights GCN's susceptibility to over-smoothing due to excessive parameters, whereas TSC can effectively address this issue.

In the second row of Figure \ref{fg:vis}, all points display relatively dispersed distributions, avoiding the collapsing phenomenon observed in Figures \ref{fg:vis}(d-f). This suggests that SGC-based models have the ability to mitigate over-smoothing. With the addition of AIR optimization, the differentiation between node categories becomes clearer. In comparison, TSC aggregates points of each class into smaller clusters, exhibit better discriminative power.

\section{Conclusion}
In this paper, we analyzed the over-smoothing in view of neighbor quality and quantity, and reviewed that existing methods increase the differences between nodes by neighbor filtering or maintain the individual information of nodes by individuality enhancement. The node difference can alleviate representation convergence, whereas the node individuality can maintain the discriminative power. We combine these two advantages and design a two-sided constraint (TSC), comprising random mask and contrastive constraint, to the column and row of representation matrix. As a plugin model, TSC is implemented in GCN and SGC. Experiments on 5 datasets demonstrate that TSC can significant alleviate the over-smoothing of GCN, improve the discriminative power of SGC, and achieve SOTA by ACC and visualization. Theoretically, we also discuss that the keys of TSC for mitigating representation convergence and maintain discriminative power are random mask and contrastive constraints respectively.

\section{Acknowledgments}
This work was supported by the Science and Technology Innovation 2030-"New Generation of Artificial Intelligence" Major Program  (No.2021ZD0112400), the National Natural Science Foundation of China (Nos. 62276162, 62106132, 62136005, 62272286), the Fundamental Research Program of Shanxi Province (No. 202203021222016), the Science and Technology Major Project of Shanxi (No. 2022010201-01006), and the Central guidance for Local scientific and technological development funds (No. YDZJSX20231B001).

\balance



\newpage

\appendix
\section{Dataset Summary}
\begin{table}[h]
	\caption{The summary  of datasets}
	\label{tb:dataset}
	\centering
	\resizebox{\linewidth}{!}{
			\begin{tabular}{lcccccl}
					\toprule
					Datasets  & Classes & Nodes & Edges & Features & Train/Test & Density \\
					\midrule
					Cora     & 7          & 2708  &  5429  & 1433  &  140/1000       & 0.0014\\
					Citeseer     & 6          & 3327  &  4732  & 2703  &  120/1000       &0.0008  \\
					Pubmed     & 3          & 19717  &  44338  & 500  &  60/1000       &0.0002 \\
					CoauthorCS     & 15          & 18333  &  81894  & 6805  &  300/1000       & 0.0005 \\
					AmazonPhoto     & 8          & 7650  &  119043  & 745  &  160/1000       &0.0042 \\
					\bottomrule
			\end{tabular}}
\end{table}

\section{Differences from Other Methods}
We use a simple illustration to show the difference between our method and other popular models such as GCN, SkipNode \cite{lu2024skipnode}, DropNode, DropEdge, DropMessage.

\begin{figure}[hbt!]	
	\centering	
	\subfigure{
		\centering
		\includegraphics[width=0.25\linewidth]{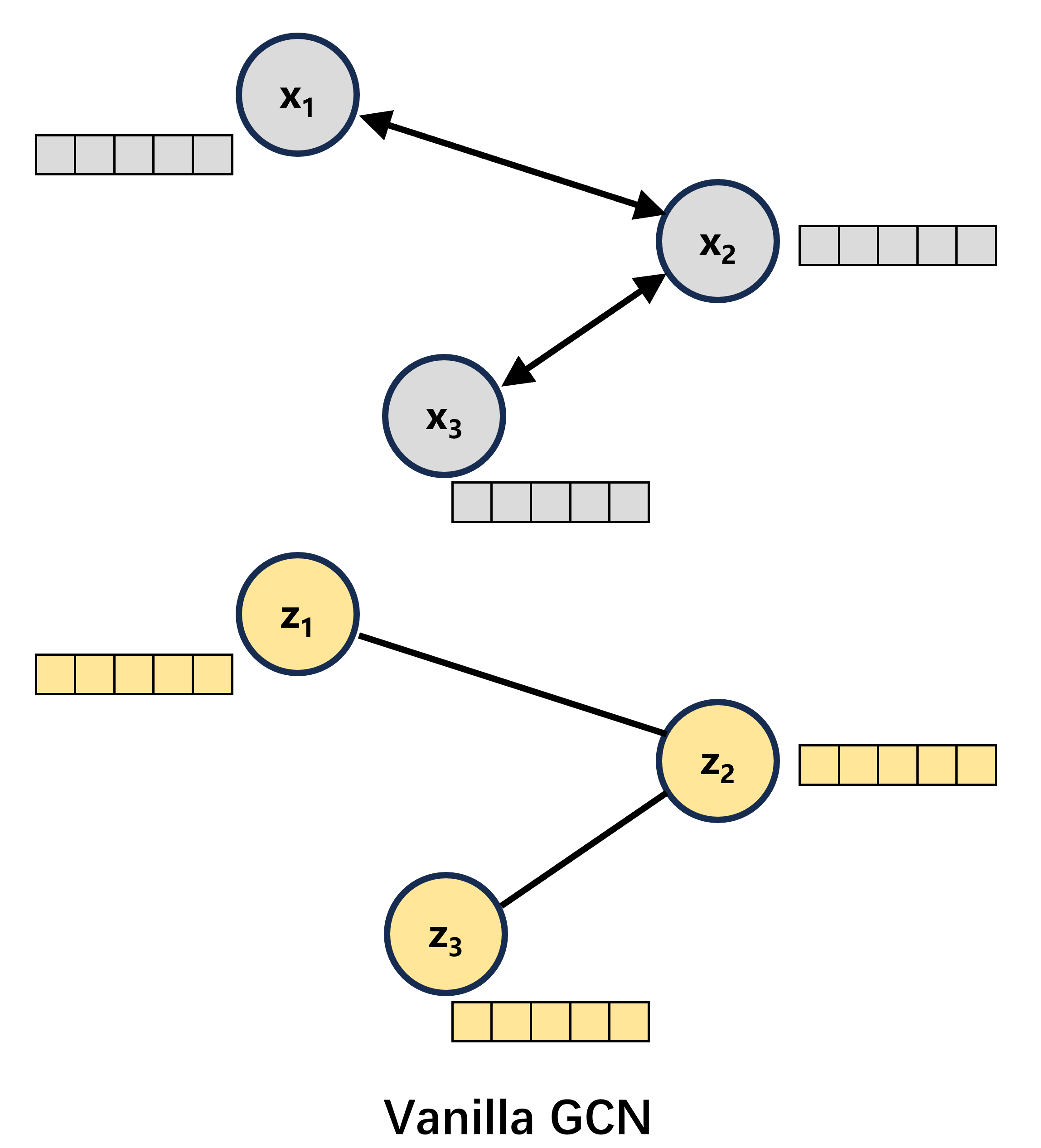}
	}
	\subfigure{
		\centering
		\includegraphics[width=0.25\linewidth]{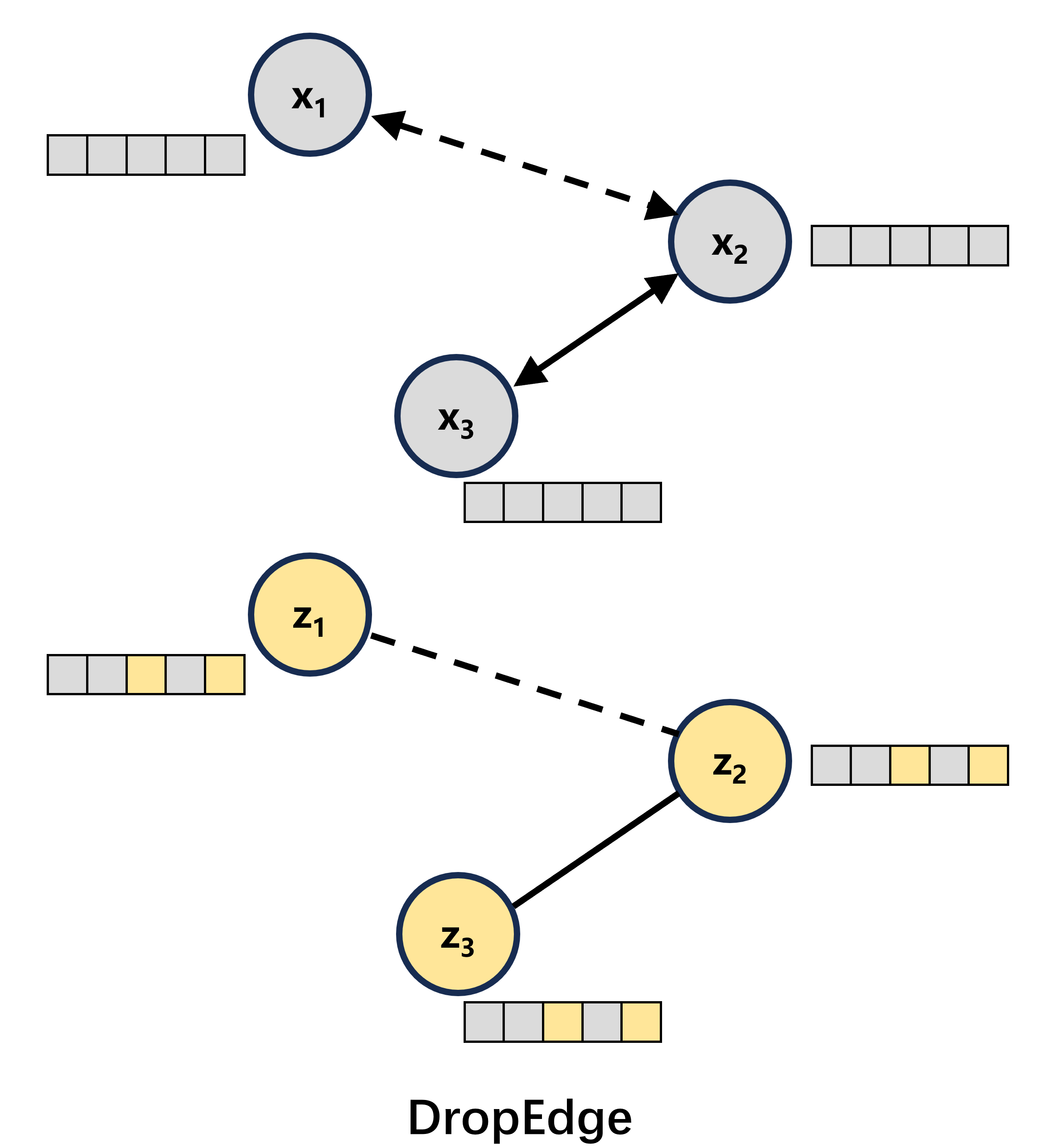}
	}
	\subfigure{
		\centering
		\includegraphics[width=0.25\linewidth]{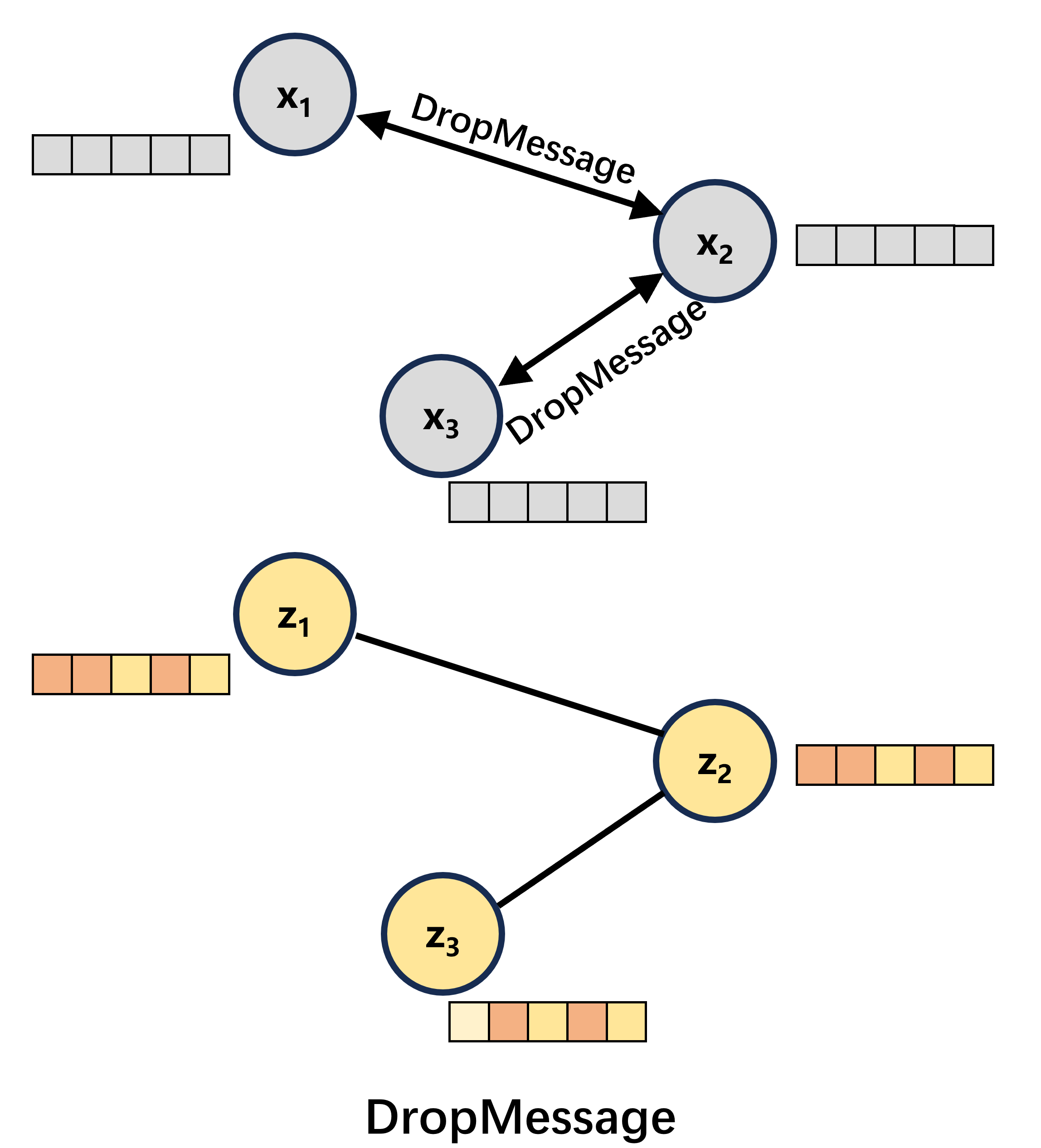}
	}
	\subfigure{
		\centering
		\includegraphics[width=0.25\linewidth]{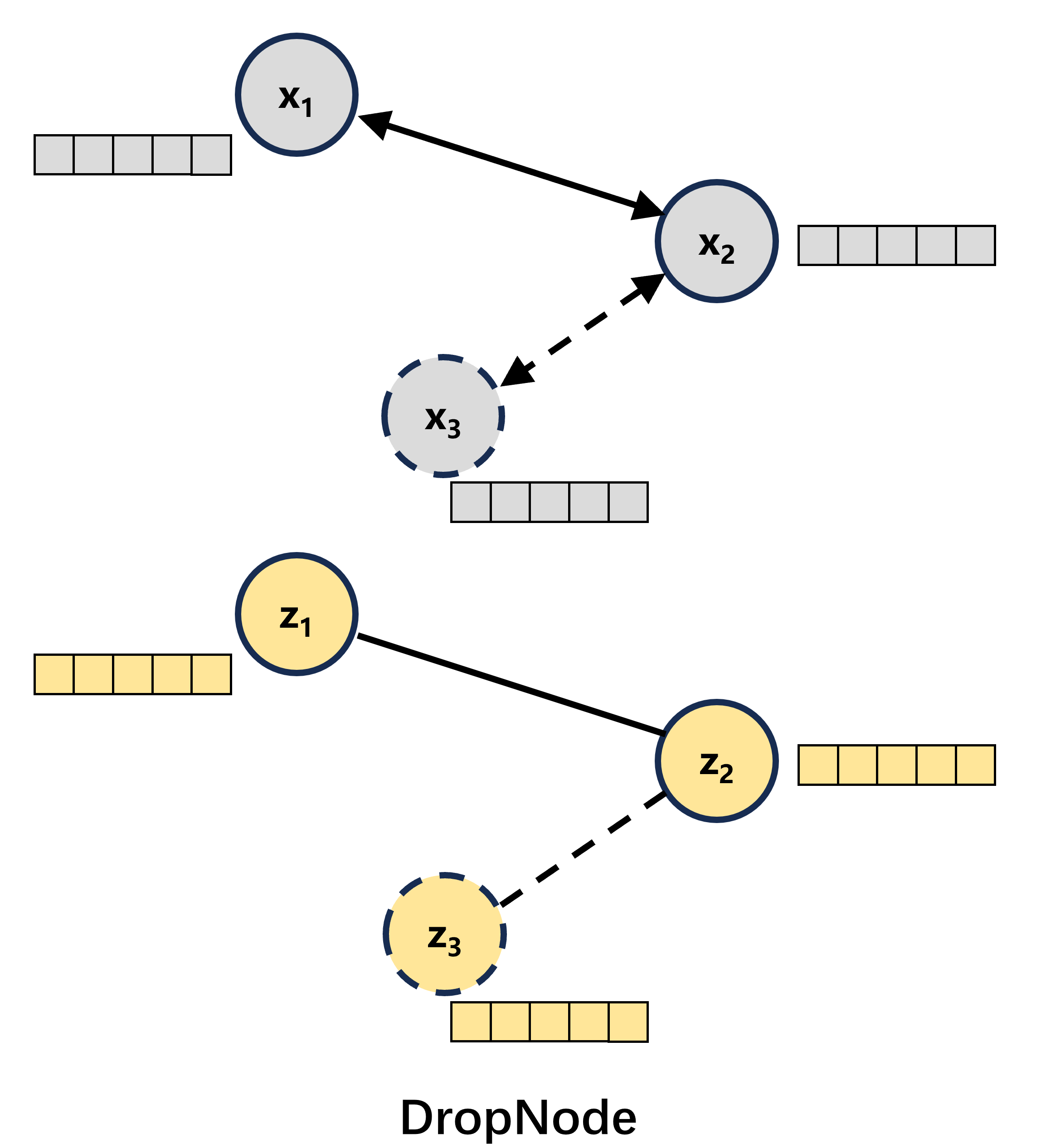}
	}
	\subfigure{
	\centering
	\includegraphics[width=0.25\linewidth]{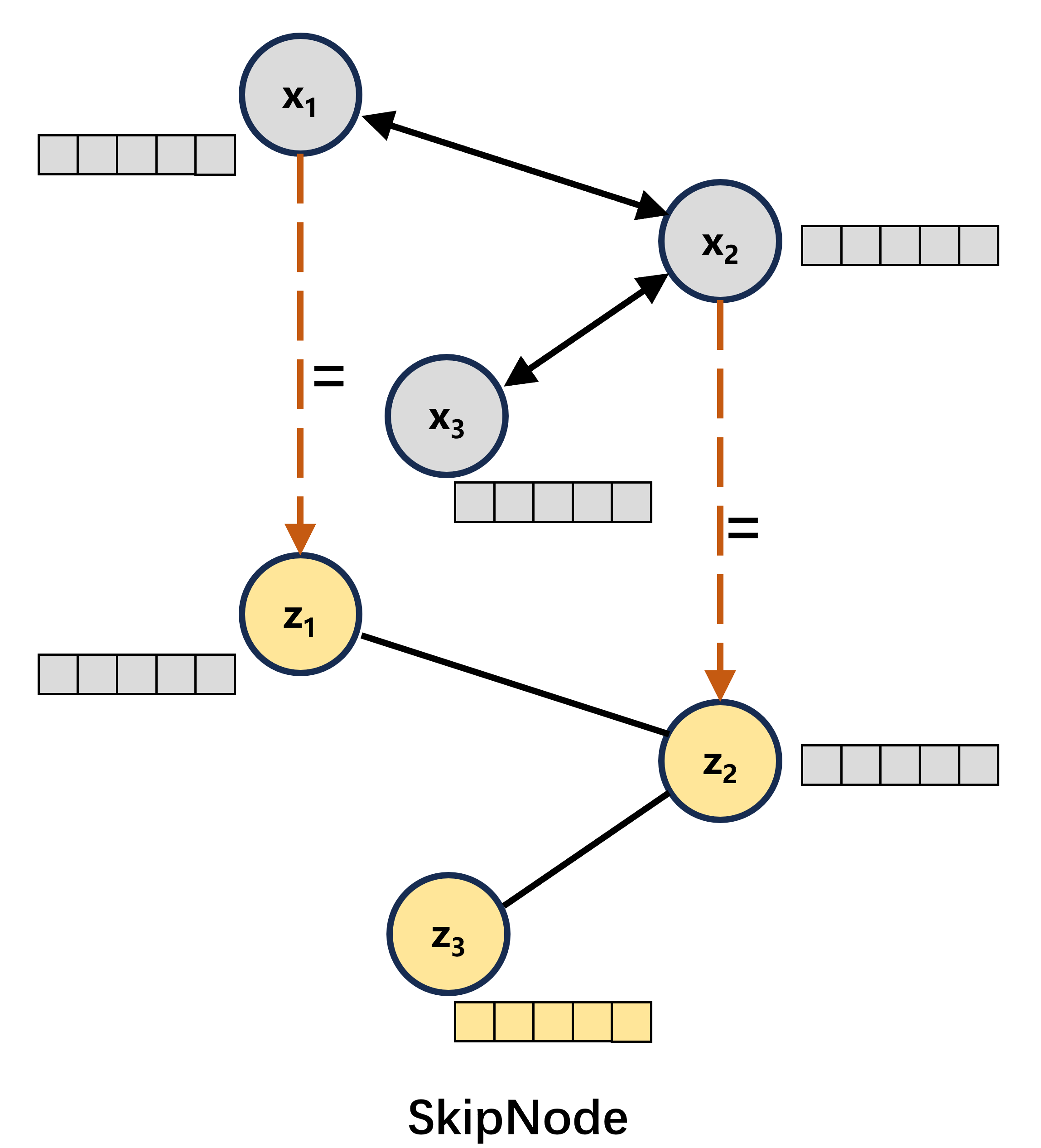}
}
\subfigure{
	\centering
	\includegraphics[width=0.25\linewidth]{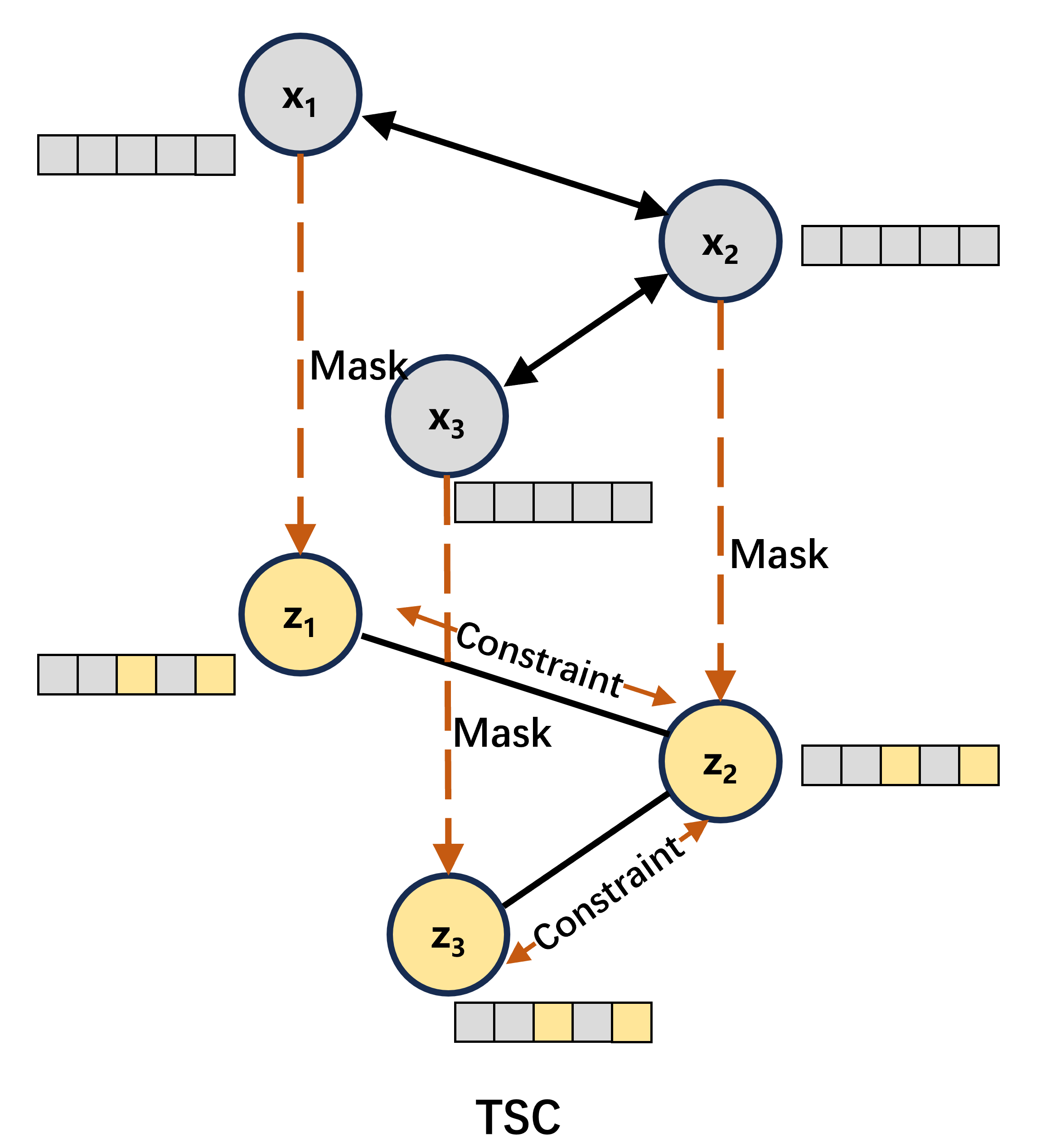}
}
\subfigure{
	\centering
	\includegraphics[width=0.9\linewidth]{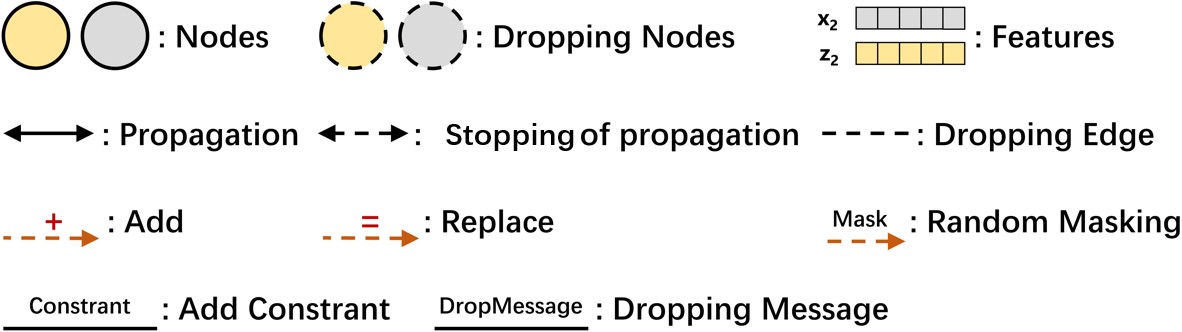}
}
	\caption{Comparing our method with popular models. We use Vanilla GCN as an example for illustration. Where the solid dots indicate the nodes, x1, x2, x3 are the features of the three nodes, and z1, z2, z3 are the features of the nodes obtained after one convolution operation for the corresponding nodes.
	}    
	\label{fg:drops}
\end{figure}

\section{Implementation Details}
For the node classification task, we apply the  fixed training/testing split on all datasets, with 20 nodes per class for training and 1,000 nodes for testing. The hidden dimensions of all baselines use $256$, which is the same as the hidden dimensions of our model, and we tuned these baselines for optimal performance. Our model has two hyperparameters, the temperature coefficient $\tau$ and $\lambda$ in the Random Masking strategy. The $\tau$  generally takes the values of $0.4$ and $0.5$. $\lambda$ controls the masking rate and it is generally between $0.2 $ and $ 0.8$. 

\section{Hyperparameter Analysis}
The hyperparameters of TSC mainly consist of the mask change rate $\lambda$ in the random mask as defined in Equation (\ref{eq:random-mask}), and the temperature coefficient $\tau$ in the contrastive constraint as defined in Equation (\ref{eq:l-sgc}) and (\ref{eq:l-gcn}). We take the Cora and Citeseer datasets as examples to analyze the impact of these parameters on the SGC+TSC model in terms of ACC and MAD metrics.

\subsection{Performance v.s. $\lambda$}
\begin{figure}[H]	
	\centering
	\subfigure[$\lambda$ v.s. ACC   on Cora]{
		\includegraphics[width=0.46\linewidth]{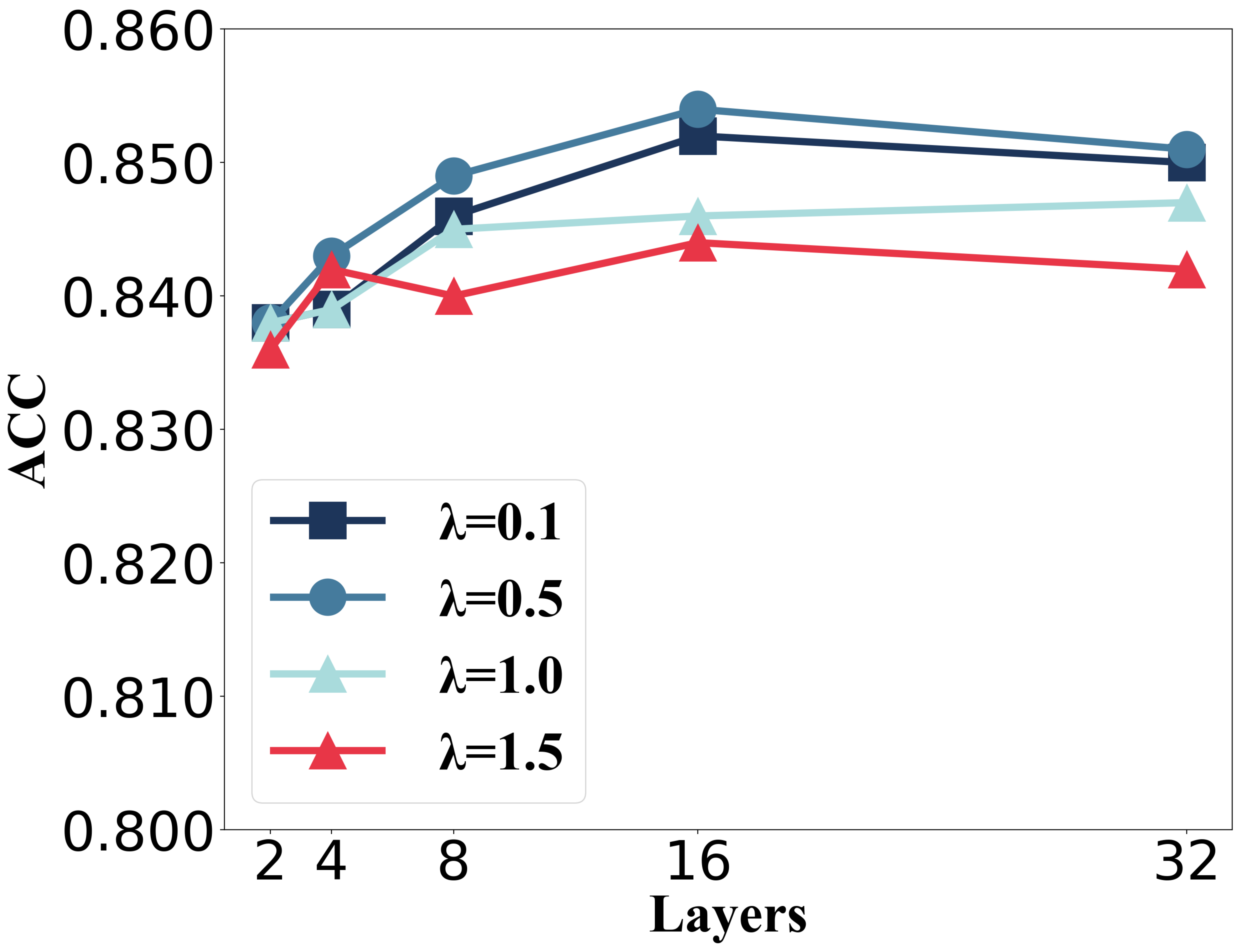}
	}	
	\subfigure[$\lambda$ v.s. ACC   on Citeseer]{
		\includegraphics[width=0.46\linewidth]{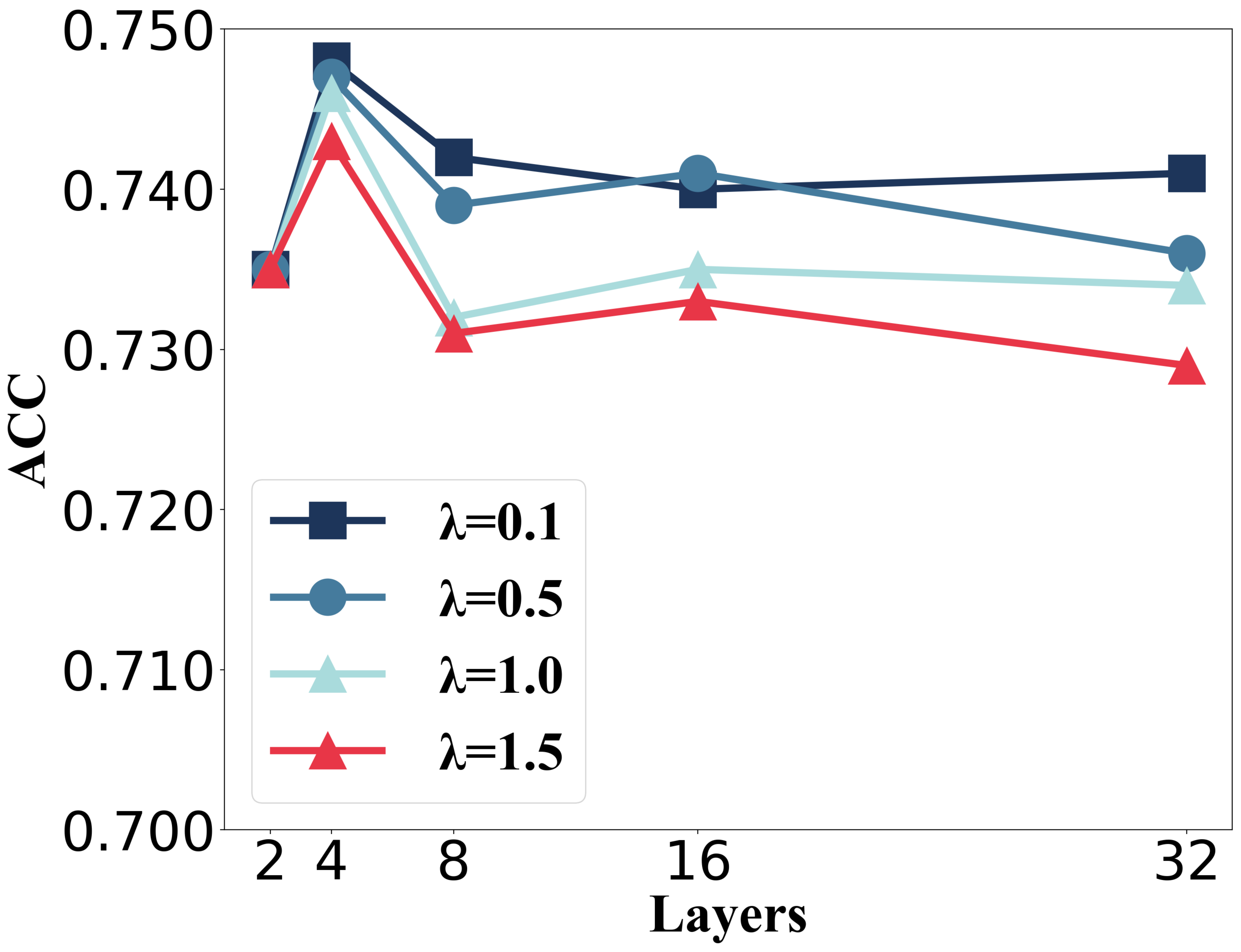}
	}
	\subfigure[$\lambda$ v.s. MAD  on Cora]{
		\includegraphics[width=0.46\linewidth]{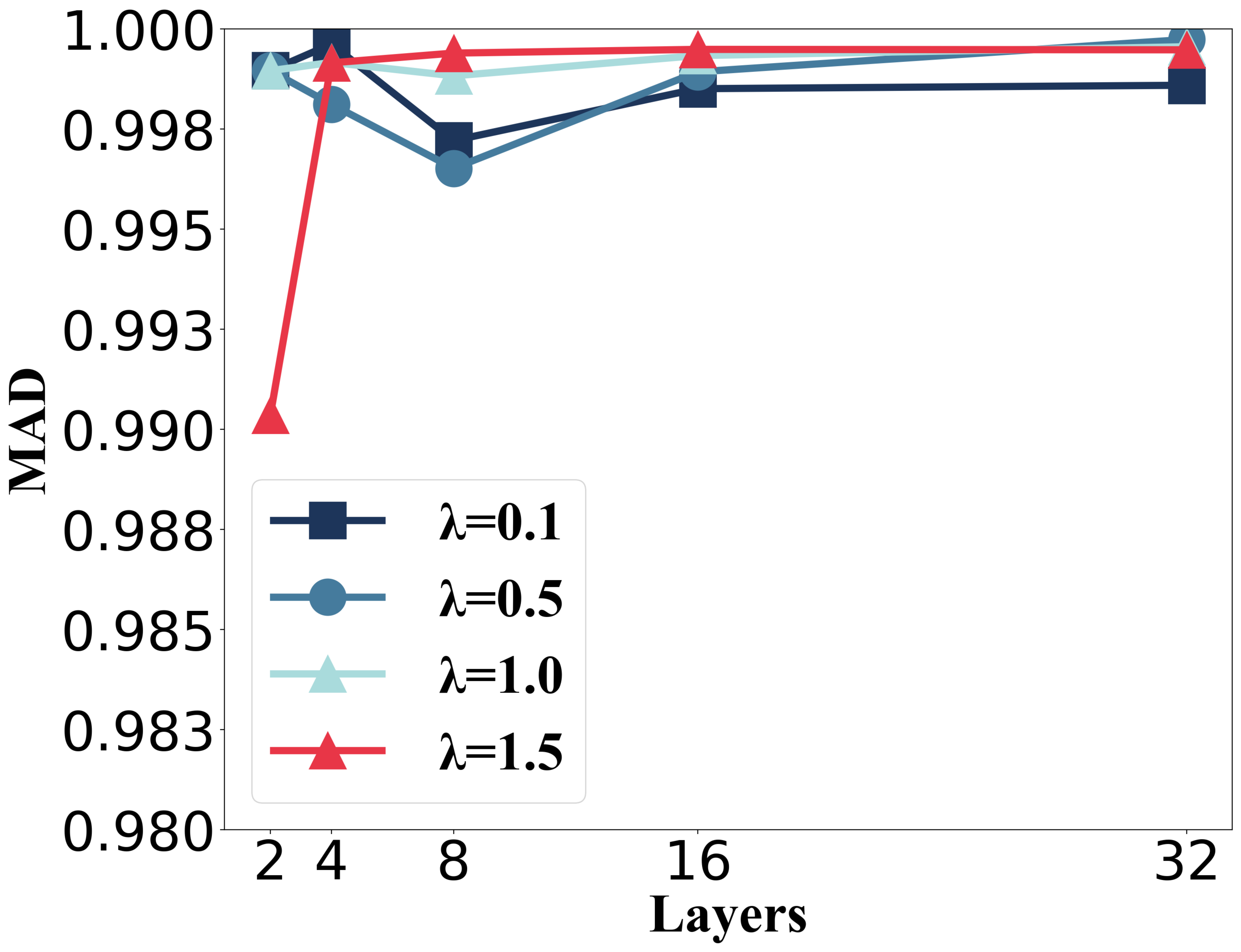}
	}
	\subfigure[$\lambda$ v.s. MAD on Citeseer]{
		\includegraphics[width=0.46\linewidth]{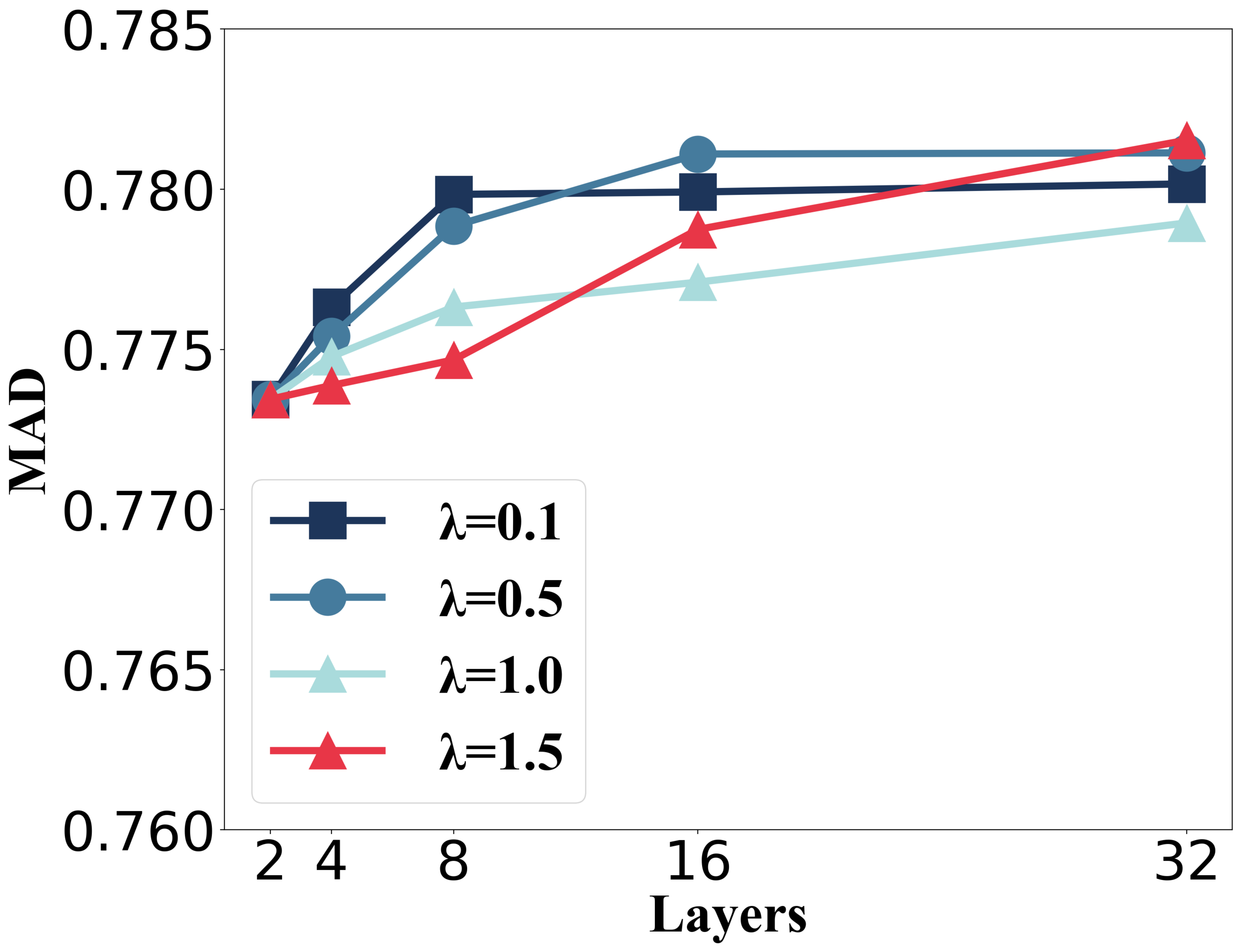}
	}
	\caption{Performance v.s. $\lambda$}
	\label{fg:param-lambda}
\end{figure}

Figure \ref{fg:param-lambda} shows the impact of $\lambda$ on the ACC and MAD metrics for SGC+TSC. From the first row of Figure \ref{fg:param-lambda}, it is observed that the accuracy of the model rapidly declines when the depth exceeds 16 layers. For instance, with $\lambda=1.5$, there is a noticeable performance drop in ACC on both the Cora and Citeseer datasets. The second row of Figure \ref{fg:param-lambda} indicates that, as $\lambda$ increases, the MAD metric does not significantly decrease but rather shows an increase. This suggests that updating by columns can effectively mitigate over-smoothing but does not address the decline in classification performance. Therefore, further consideration of individuality enhancement for nodes is suggested again.

\subsection{Performance v.s. $\tau$}
\begin{figure}[t]	
	\centering
	\subfigure[$\tau$ v.s. ACC  on Cora]{
		\includegraphics[width=0.45\linewidth]{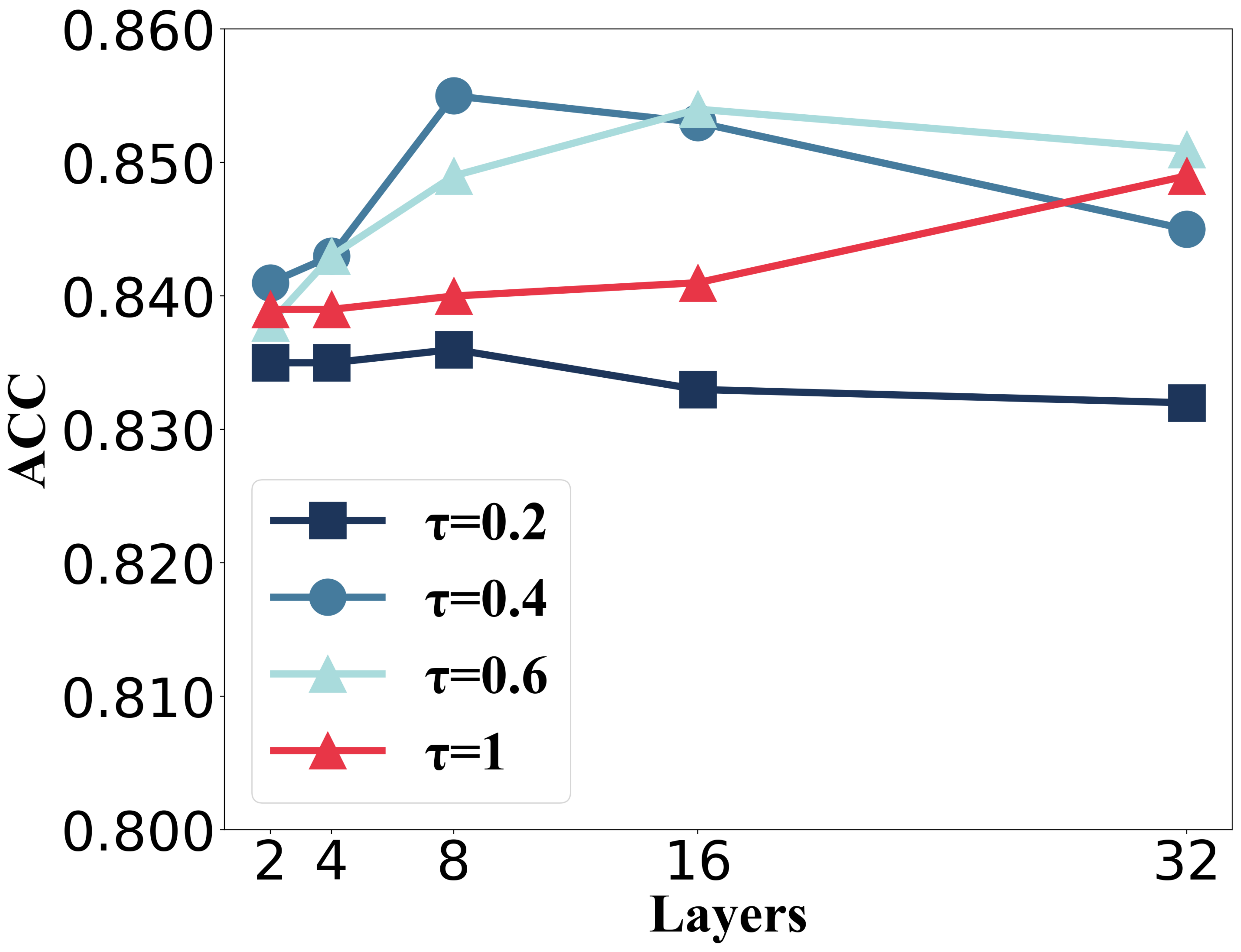}
	}	
	\subfigure[$\tau$ v.s. ACC   on Citeseer]{
		\includegraphics[width=0.45\linewidth]{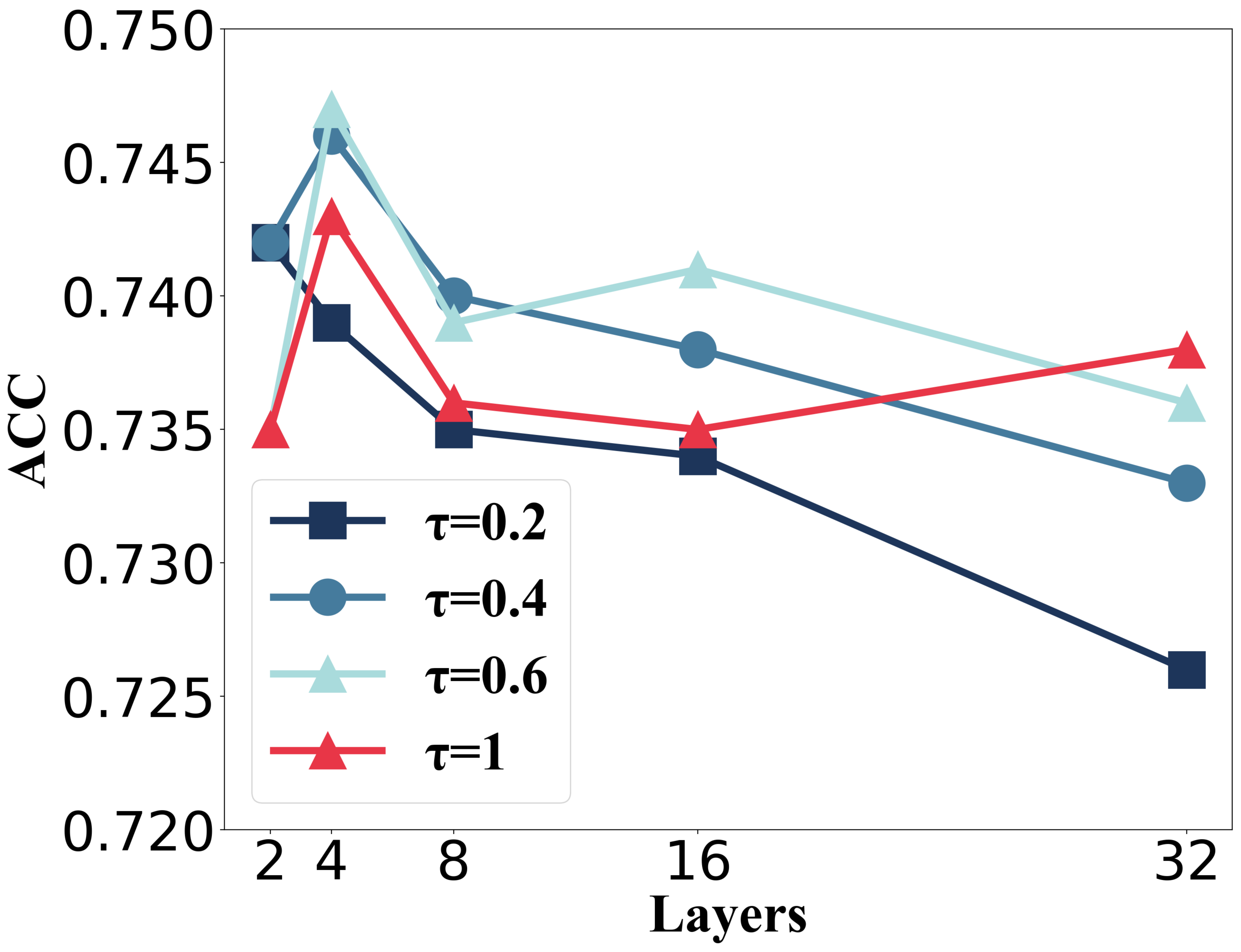}
	}
		\subfigure[$\tau$ v.s. MAD  on Cora]{
		\includegraphics[width=0.45\linewidth]{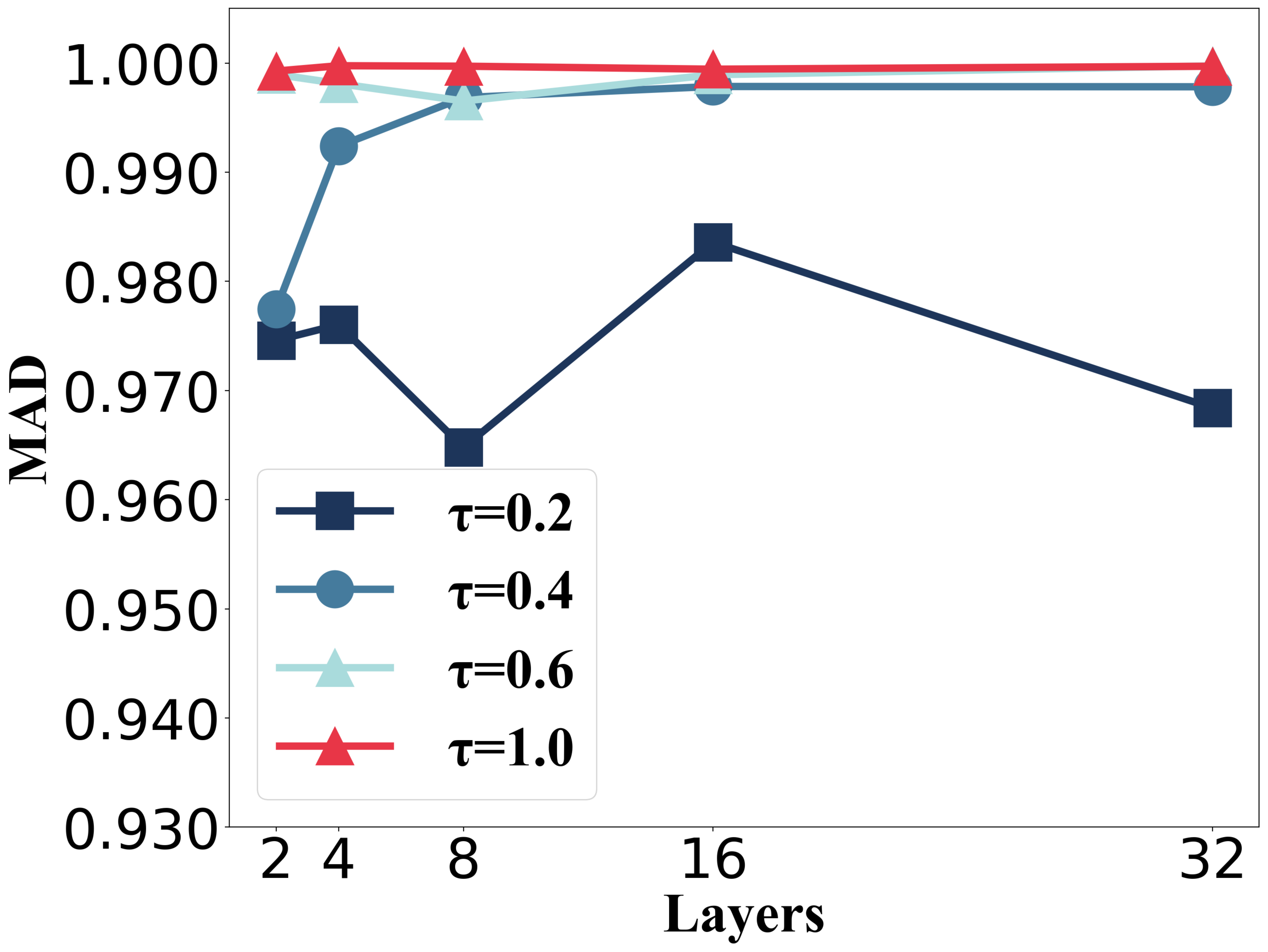}
	}
	\subfigure[$\tau$ v.s. MAD on Citeseer]{
		\includegraphics[width=0.45\linewidth]{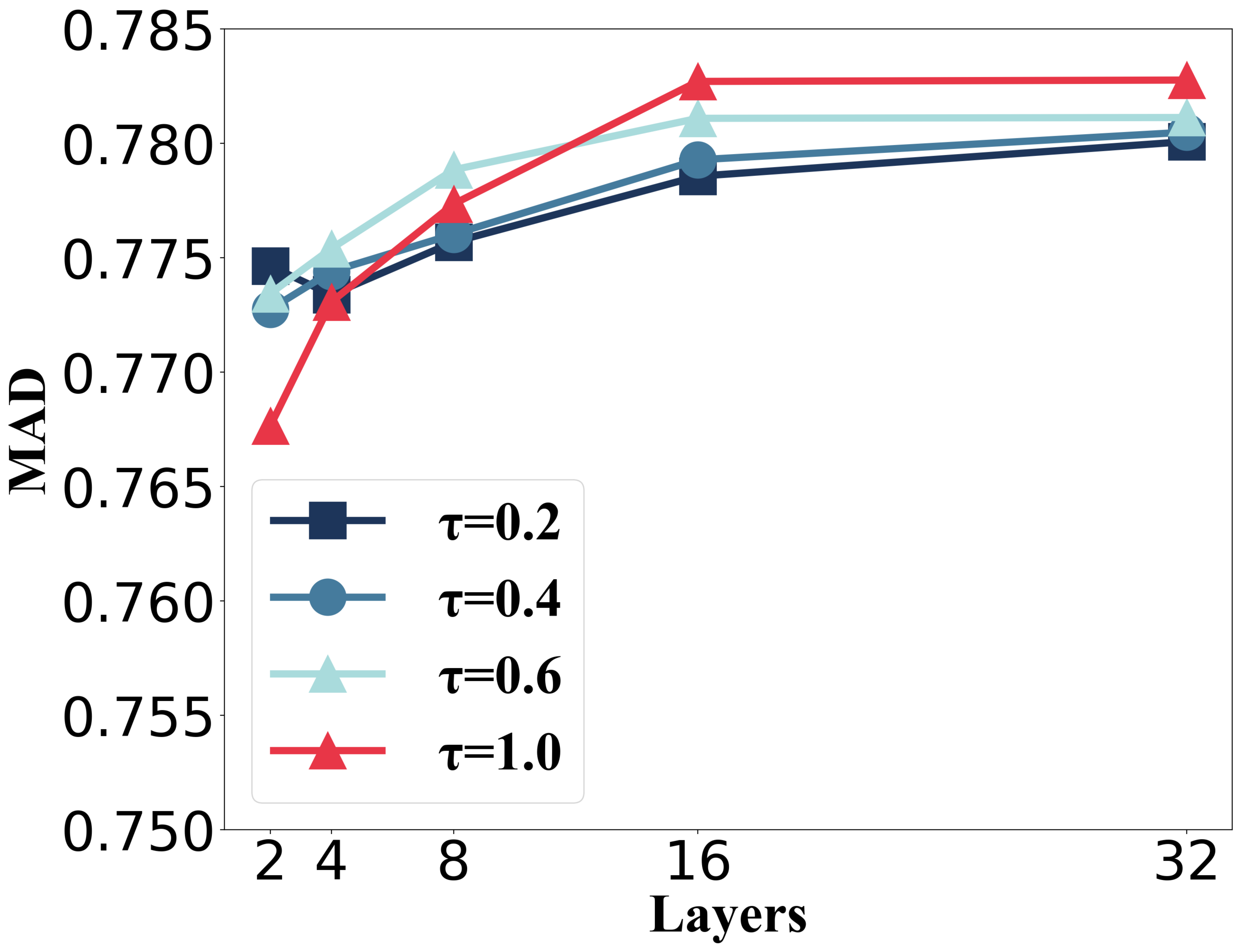}
	}
	\caption{Performance v.s. $\tau$}
	\label{fg:param-tau}
\end{figure}

Figure \ref{fg:param-tau} presents the effects of $\tau$ on ACC and MAD metrics for SGC+TSC. The first row of Figure \ref{fg:param-tau} reveals that a too large value of $\tau$ can lead to significant fluctuations in accuracy. For example, at $\tau=1$ in Figure \ref{fg:param-tau}(a), the model exhibits fluctuations on the Cora dataset that are distinct from other parameter settings, resulting in slow performance improvement. The second row of Figure \ref{fg:param-tau} shows that while $\tau=1$ achieves good results on the MAD metric, it does not translate into improved classification performance. Therefore, we recommend choosing a smaller value for $\tau$ (around 0.5) to ensure better classification outcomes.

\section{More Experiments}
We also added more experiments. For example, combining our method with the popular GAT \cite{velickovic2017graph} and GCNII to validate the pluginability of the TSC by Table \ref{tb:acc-at-Cora}, \ref{tb:acc-at-Citeseer}, \ref{tb:acc-at-Cora2}, and \ref{tb:acc-at-Citeseer2}. As well, we demonstrate the effect of our method on two heterogeneous datasets by Table \ref{tb:acc-hete-Cornell} and \ref{tb:acc-hete-Wisconsin}.
\begin{table}[h]
	\centering
	\caption{ACC comparison in different depth on Cora}
		\resizebox{0.8\linewidth}{!}{
		\begin{tabular}{ccccc}
			\toprule
			& Layer 2 & Layer 4 & Layer 8 & Layer 16  \\
			\toprule
			GAT     & \textbf{0.831}   & 0.603  & 0.13  & 0.13      \\
			\toprule
			GAT+TSC    & 0.827  &   \textbf{0.816}  & \textbf{0.811}    & \textbf{ 0.773 }   \\
			\bottomrule
		\end{tabular}
		}	
	\label{tb:acc-at-Cora}
\end{table}

\begin{table}[h]
	\centering
	\caption{ACC comparison in different depth on Citeseer}
		\resizebox{0.8\linewidth}{!}{
		\begin{tabular}{ccccc}
			\toprule
			& Layer 2 & Layer 4 & Layer 8 & Layer 16  \\
			\toprule
			GAT     & \textbf{0.729}   & 0.594  & 0.077  & 0.077      \\
			\toprule
			GAT+TSC    & 0.724  &   \textbf{0.717}  & \textbf{0.713}    & \textbf{ 0.716 }   \\
			\bottomrule
		\end{tabular}
		}	
	\label{tb:acc-at-Citeseer}
\end{table}

\begin{table}[h]
	\centering
	\caption{ACC comparison in different depth on Cora}
		\resizebox{0.8\linewidth}{!}{
		\begin{tabular}{cccccc}
			\toprule
			& Layer 2 & Layer 4 & Layer 8 & Layer 16 & Layer 32  \\
			\toprule
			GCNII   & \textbf{0.830}  & \textbf{0.840}   & \textbf{0.829}  & 0.838  & 0.852      \\
			\toprule
			GCNII+TSC   & 0.606 & 0.772  &   0.815  & \textbf{0.855}    & \textbf{ 0.860 }   \\
			\bottomrule
		\end{tabular}
		}	
	\label{tb:acc-at-Cora2}
\end{table}

\begin{table}[H]
	\centering
	\caption{ACC comparison in different depth on Citeseer}
		\resizebox{0.8\linewidth}{!}{
		\begin{tabular}{cccccc}
			\toprule
			& Layer 2 & Layer 4 & Layer 8 & Layer 16 & Layer 32  \\
			\toprule
			GCNII   & \textbf{0.706}  & 0.720   & 0.734  & 0.736  & \textbf{0.744}     \\
			\toprule
			GCNII+TSC   & 0.607 & \textbf{0.744}  &  \textbf{ 0.748}  & \textbf{0.737}    &  0.739    \\
			\bottomrule
		\end{tabular}
		}	
	\label{tb:acc-at-Citeseer2}
\end{table}

\begin{table}[H]
	\centering
	\caption{ACC comparison in different depth on heterogeneous dataset Cornell}
	\resizebox{0.8\linewidth}{!}{
		\begin{tabular}{cccccc}
			\toprule
			& Layer 2 & Layer 4 & Layer 8 & Layer 16 & Layer 32  \\
			\toprule
			GCN   & 0.40  & 0.46  & 0.26  & 0.48  & 0.22    \\
			SGC   & 0.50  & \textbf{0.56}   & \textbf{0.6}  & 0.54  & 0.58    \\
			\toprule
			GCN+TSC   & \textbf{0.56} & 0.50  &   0.56  & 0.56   &  0.52    \\
			SGC+TSC   & \textbf{0.56} & 0.56 &  \textbf{ 0.58}  & \textbf{0.62}    &  \textbf{0.66}    \\
			\bottomrule
		\end{tabular}
	}	
	\label{tb:acc-hete-Cornell}
\end{table}

\begin{table}[H]
	\centering
	\caption{ACC comparison in different depth on heterogeneous dataset Wisconsin}
	\resizebox{0.8\linewidth}{!}{
		\begin{tabular}{cccccc}
			\toprule
			& Layer 2 & Layer 4 & Layer 8 & Layer 16 & Layer 32  \\
			\toprule
			GCN   & 0.54  & 0.52  & 0.40  & 0.54  & 0.32    \\
			SGC   & 0.54  & \textbf{0.60}   & 0.56  & 0.56  & 0.54    \\
			\toprule
			GCN+TSC   & \textbf{0.56} & 0.50  &   0.56  & 0.56   &  0.52    \\
			SGC+TSC   & \textbf{0.56} & 0.58 &  \textbf{ 0.60}  & \textbf{0.58}    &  \textbf{0.60}    \\
			\bottomrule
		\end{tabular}
	}	
	\label{tb:acc-hete-Wisconsin}
\end{table}

\section{Understanding Neighbor quantity and Neighbor quality}

\textbf{Neighbor quality}: the term neighbor quality  refers to the variability of neighbor information of all nodes aggregated. The higher this variability indicates that the neighbor quality of the node is higher, otherwise it is lower. We can use an indicator \textbf{Average Mutual Overlapping} (AMO) to describe the \textbf{neighbor quality}. A smaller AMO indicates a better quality neighborhood, otherwise a poorer one. Poor neighbor quality is also more likely to cause node features to become indistinguishable. AMO is defined as follows:
$$\left.S_{i,j}=\left\{\begin{array}{l}1,A_{i,j}^l>0\wedge i\neq j\\0,A_{i,j}^l=0\\\end{array}\right.,AMO=\mathrm{Mean}(\mathrm{SS}^T))\right.$$
$l$ is the number of layers, $S$ is an indicator matrix, and $A$ is the adjacency matrix.
As its name suggests, this indicator averages the number of common neighbors of two nodes over all possible pairs of nodes. It is a sufficient indicator since two nodes have similar aggregation information when they have a large number of common neighbors. 

\textbf{Number of neighbor}: the information received from neighbors grows exponentially, causing nodes to lose their own individuality and accuracy decreases. The number of neighbors aggregated by one node increases exponentially as the number of convolutions increases, and the influx of a large number of nodes that are not of the same class causes the node to eventually lose its own individuality, leading to a decrease in the accuracy rate. We can measure the impact of the \textbf{number of neighbor} of one node on the node by \textbf{Average number of different classes of nodes in the neighbor}(ANDCNN). When the ANDCNN is larger, the more the neighbors of the node aggregation contain nodes that are not of the same class, the more the node suffers from the impact. Nodes are more likely to lose their individuality.
$$ANDCNN=\frac{1}{N}\sum_{i=0j=0}^{N}1_{[L_{i}!=L_{j} \wedge  S_{i,j}=1]} , $$

Where $L_i$ and $L_j$ represent the labels of node $i$ and node $j$. $N$ is the total number of nodes in the graph.

\end{document}